\documentclass[12pt,a4paper]{report}
\usepackage{thesis}
\usepackage{lmodern}
\usepackage[numbers,sort&compress]{natbib}
\usepackage[hidelinks]{hyperref}
\usepackage[left=2.5cm,top=2.5cm,right=2.5cm,bottom=2.5cm]{geometry}
\usepackage{nomencl}
\usepackage{booktabs}
\usepackage{tabularx}
\usepackage{latexsym}
\usepackage{mathrsfs}
\usepackage{amsmath,amsfonts,amssymb,amsthm, bm}
\usepackage{bbm}
\usepackage{tabularray}

\usepackage{graphicx}
\usepackage{algorithm}
\usepackage{algpseudocode}
\usepackage{multirow}
\usepackage{CJKutf8}
\usepackage{xurl}
\usepackage{booktabs}
\usepackage{lipsum}
\usepackage{enumitem}
\usepackage{tabto}
\usepackage{mathrsfs}
\usepackage{caption}
\usepackage{subcaption}
\usepackage{wrapfig}
\usepackage{amsmath}
\DeclareMathOperator*{\argmax}{argmax} 
\usepackage{pifont}

\usepackage{titlesec}
\titleformat{\paragraph}
{\normalfont\normalsize\bfseries}{\theparagraph}{1em}{}
\titlespacing*{\paragraph}
{0pt}{3.25ex plus 1ex minus .2ex}{1.5ex plus .2ex}

\newcommand{\xmark}{\ding{55}}%

\usepackage{url}
\usepackage{mathptmx} % times new roman font
\usepackage[T1]{fontenc}

\usepackage[export]{adjustbox}

\makenomenclature

\usepackage{tocloft}
\usepackage{parskip}

\usepackage{arydshln}

\usepackage{soul}
\usepackage{xcolor}

\colorlet{soulred}{red!20}
\DeclareRobustCommand{\hlred}[1]{{\sethlcolor{soulred}\hl{#1}}}

\colorlet{soulbleu}{cyan!40}

\colorlet{soulgreen}{green!20}
\DeclareRobustCommand{\hlgreen}[1]{{\sethlcolor{soulgreen}\hl{#1}}}

\colorlet{soulyellow}{yellow!40}

\colorlet{soulorange}{orange!50}
\DeclareRobustCommand{\hlorange}[1]{{\sethlcolor{soulorange}\hl{#1}}}

\colorlet{soulpurple}{blue!50}

\definecolor{cornflowerblue}{rgb}{0.39, 0.58, 0.93}
\colorlet{soulcornflowerblue}{cornflowerblue!50}

\definecolor{mypink1}{rgb}{0.858, 0.188, 0.478}
\definecolor{mypink2}{RGB}{219, 48, 122}
\definecolor{mypink3}{cmyk}{0, 0.7808, 0.4429, 0.1412}
\definecolor{mygray}{gray}{0.6}
\definecolor{mygreen1}{RGB}{46, 139, 87}
\definecolor{mygreen2}{rgb}{0,153,0}

\makeatletter

\def\uwave{\bgroup \markoverwith{\lower3.5\p@\hbox{\sixly \textcolor{red}{\char58}}}\ULon}

\newcommand{\thesistitle}{\textbf{Generative Long-form Question Answering: Relevance, Faithfulness and Succinctness}}
\newcommand{\thesisauthor}{\textbf{Dan Su}}
\newcommand{\programname}{Electronic and Computer Engineering}
\newcommand{\departmentname}{Department of Electronic and Computer Engineering}
\newcommand{\thesisdate}{Aug 2022}

\newcommand{\thesisdegree}{PhD}

\begin{document}

\pagenumbering{roman}
\pagestyle{plain}
\setcounter{page}{1}
\addcontentsline{toc}{chapter}{Title Page}
\thispagestyle{empty}
\null\vskip0.5in
\begin{center}
  \begin{LARGE}
    % \rmfamily
    \thesistitle
  \end{LARGE}
  \vfill
  \vspace{20mm}

  \begin{Large}
  by
  \end{Large}

  \vspace{4mm}
  \begin{Large}
  \thesisauthor 
  \end{Large}\\
  \vfill
  \vspace{20mm}
  \begin{large}
  A Thesis Submitted to\\
  The Hong Kong University of Science and Technology \\
  in Partial Fulfillment of the Requirements for\\
  the Degree of Doctor of Philosophy \\
  in the Department of \programname \\
  \vfill \vfill
  \thesisdate, Hong Kong
  \end{large}
  \vfill
\end{center}

\vfill

% \newpage
% \addcontentsline{toc}{chapter}{Authorization Page}
% \include{2_authorization}

% \newpage
% \addcontentsline{toc}{chapter}{Signature Page}
% \include{3_signature}

% \newpage
% \include{4_dedicate}

% \newpage
% \addcontentsline{toc}{chapter}{Acknowledgments}
% \include{5_acknowledgement}

\newpage
\addcontentsline{toc}{chapter}{Table of Contents}
\tableofcontents

\newpage
\addcontentsline{toc}{chapter}{List of Figures}
\listoffigures

\newpage
\addcontentsline{toc}{chapter}{List of Tables}
\listoftables

\newpage
\addcontentsline{toc}{chapter}{Abstract}
\begin{center}
{\Large \textbf{\thesistitle}}\\
\vspace{20mm}
by \thesisauthor\\
%\vspace{15mm}
\departmentname\\
%\vspace{10mm}
The Hong Kong University of Science and Technology
\end{center}
\vspace{8mm}
\begin{center}
\Large Abstract
\end{center}
\par
\noindent

\setlength{\emergencystretch}{10pt}

Question answering (QA) aims to build computer systems that can automatically answer questions posed by humans, and it has been a long-standing problem in natural language processing (NLP). This thesis investigates a particular problem of generative long-form question answering (\textbf{LFQA}), which aims to generate an in-depth, paragraph-length answer for a given question posed by a human. 

% why long-form, and why generative
Generative LFQA is an important task. A large ratio of the questions that humans deal with daily and ask on search engines are complicated \textit{why/how} types, which require multi-sentence explanations to answer. For example, \textit{'How do jellyfish function without a brain?'}, \textit{'What are the risking factors related to COVID-19?'}. Furthermore, the answers normally need to be generated by synthesizing information from multiple documents, since a short phrase span extracted from a single existing document can't answer those complicated questions.

% However, LFQA is quite challenging and under-explored. Few works have been investigated to build an effective LFQA system. Since it involves retrieving multiple relevant documents from a large external corpus which usually contains tens of millions of documents, it needs to generate the answer from the multiple documents containing tens of thousands of tokens. Moreover, it is even more challenging to generate a good-quality long-form answer relevant to the query and faithful to facts since a considerable amount of redundant, complementary, or contradictory information will be contained in the retrieved documents. 

On the other hand, LFQA is quite challenging and under-explored. Few works have been done to build an effective LFQA system. It is even more challenging to generate a good-quality long-form answer relevant to the query and faithful to facts, since a considerable amount of redundant, complementary, or contradictory information will be contained in the retrieved documents. Moreover, no prior work has been investigated to generate succinct answers.

% Little prior work has been done, and none has tried to leverage the informative answers.

% grammatically correct and semantically informative is even more challenging. 

In this thesis, we investigate the task of LFQA and tackle the challenges mentioned above. Specifically, we focus on 1) how to build a practical application for real-time open-domain LFQA, and generate more query-relevant answers, 2) how to generate more factual long-form answers, and 3) how to generate succinct answers from long-form answers.

% To elaborate, we first present a real-time LFQA system that can efficiently generate multiple-documents-based answers. The system demonstrates its effectiveness at generating fluent and somewhat relevant answers, winning one of the Kaggle competitions related to COVID-19. In addition, to further improve the query-relevance of the answer, we propose to incorporate the explicit relevance score of the source documents into the generation model.

To elaborate, we first present a coarse-to-fine method to extract the document-level and sentence-level query-relevant information, to help a traditional Seq2Seq model to handle long and multiple documents as input, and consider query relevance. We further introduce QFS-BART, a model that incorporates the explicit answer relevance attention of the source documents into the generation model's encoder-decoder attention module, to further enhance the query relevance. The CAiRE-COVID system, a real-time long-form question answering system for COVID-19, that we built has won one of the Kaggle competitions related to COVID-19, judged by medical experts.

Secondly, we present a new architectural method to tackle the answer faithfulness issue. We augment the generation process with global predicted salient information from multiple source documents, which can be viewed as an emphasis on answer-related facts. State-of-the-art results on two LFQA datasets demonstrate the effectiveness of our method in comparison to solid baselines on automatic and human evaluation metrics. The method also topped one public leaderboard on the LFQA task. 

Finally, we take a step further and propose to generate succinct answers from the long-form answers. Specifically, we extract short-phrase answers for \textit{closed-book} question answering (CBQA) task from the long-form answers. Experimental results on three QA benchmarks show that our method significantly outperforms previous \textit{closed-book} QA methods and is on par with traditional \textit{open-book} methods that exploit external knowledge sources.

\printnomenclature[1in]

\newpage
\pagenumbering{arabic}
\pagestyle{plain}
\setcounter{page}{1}

\chapter{Introduction}

\begin{figure}[!ht]
  \centering
  \includegraphics[width=0.9\linewidth]{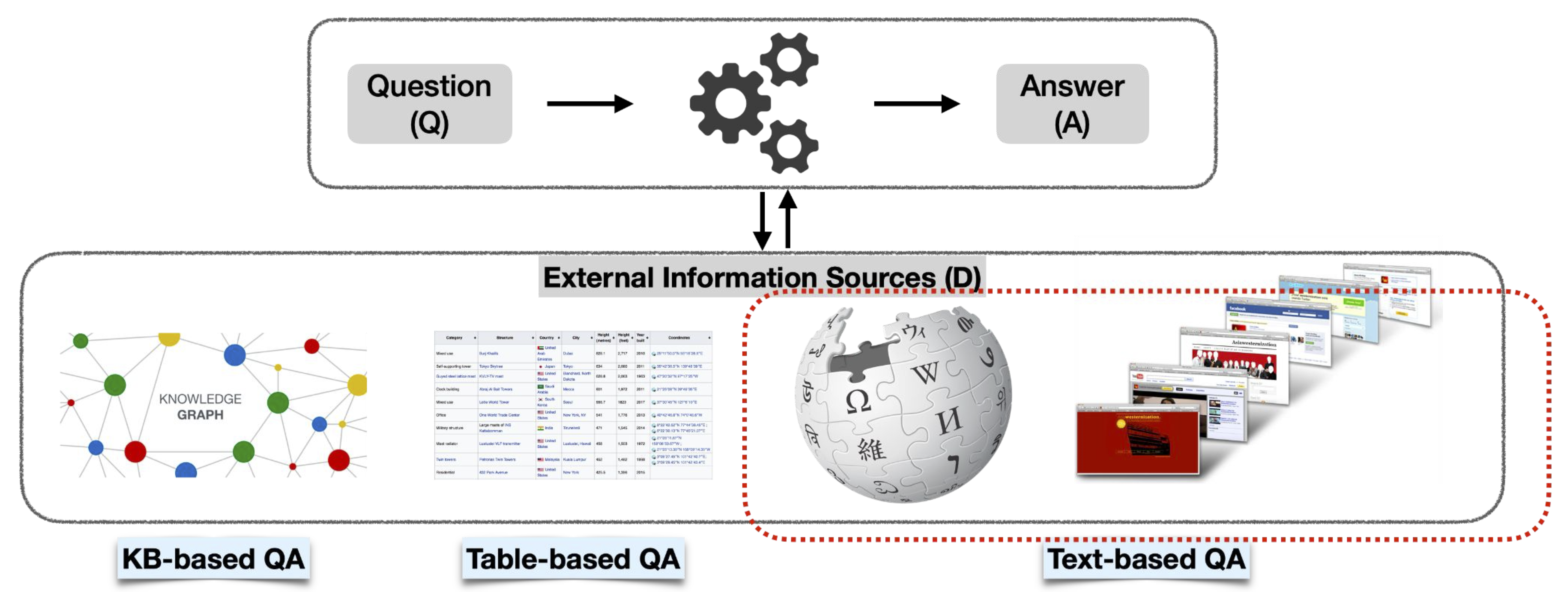}
  \caption{A Open-domain question answering system illustration. The text-based QA is the most extensively studied ODQA task.}
  \label{fig:intro_odqa}
\end{figure}

% \section{Open-domain Question Answering}
Open domain question answering (ODQA) has been a long-standing problem in the history of NLP~\cite{simmons1964indexing,kupiec1993murax,voorhees1999trec,moldovan2000structure, harabagiu2000experiments, brill2002analysis, ferrucci2010building, baudivs2015yodaqa, lin2003question,gliozzo2012natural, yih2016question, sachan2018standardized, chen2017reading}. The goal of open domain question answering is to build automated computer systems which are able to answer any sort of (factoid) questions that humans might ask, based on a large collection of unstructured natural language documents, structured data, semi-structured data or even other modalities such as images or videos (as illustrated in Figure~\ref{fig:intro_odqa}). 

% \textbf{Text-based QA} 
% The most extensively studied ODQA task is \textit{text-based} QA, which derives an answer from unstructured natural language documents (e.g., encyclopedias, dictionaries, news articles, and general Web documents). 
% Dating back to the 1960s~\cite{simmons1964indexing}, the text-based open-domain question answering(QA) has been broadly investigated by both the information retrieval (IR) and natural language processing (NLP) communities  ~\cite{simmons1964indexing,kupiec1993murax,voorhees1999trec,moldovan2000structure, harabagiu2000experiments, brill2002analysis, ferrucci2010building, baudivs2015yodaqa, lin2003question,gliozzo2012natural, yih2016question, sachan2018standardized, chen2017reading, su2020improving}. Notable text-based QA systems include Microsoft’s ASKMSR~\cite{brill2002analysis}, IBM’s DEEPQA~\cite{ferrucci2010building}, and DrQA~\cite{chen2017reading}. 

\begin{figure}[!ht]
  \centering
  \includegraphics[width=0.9\linewidth]{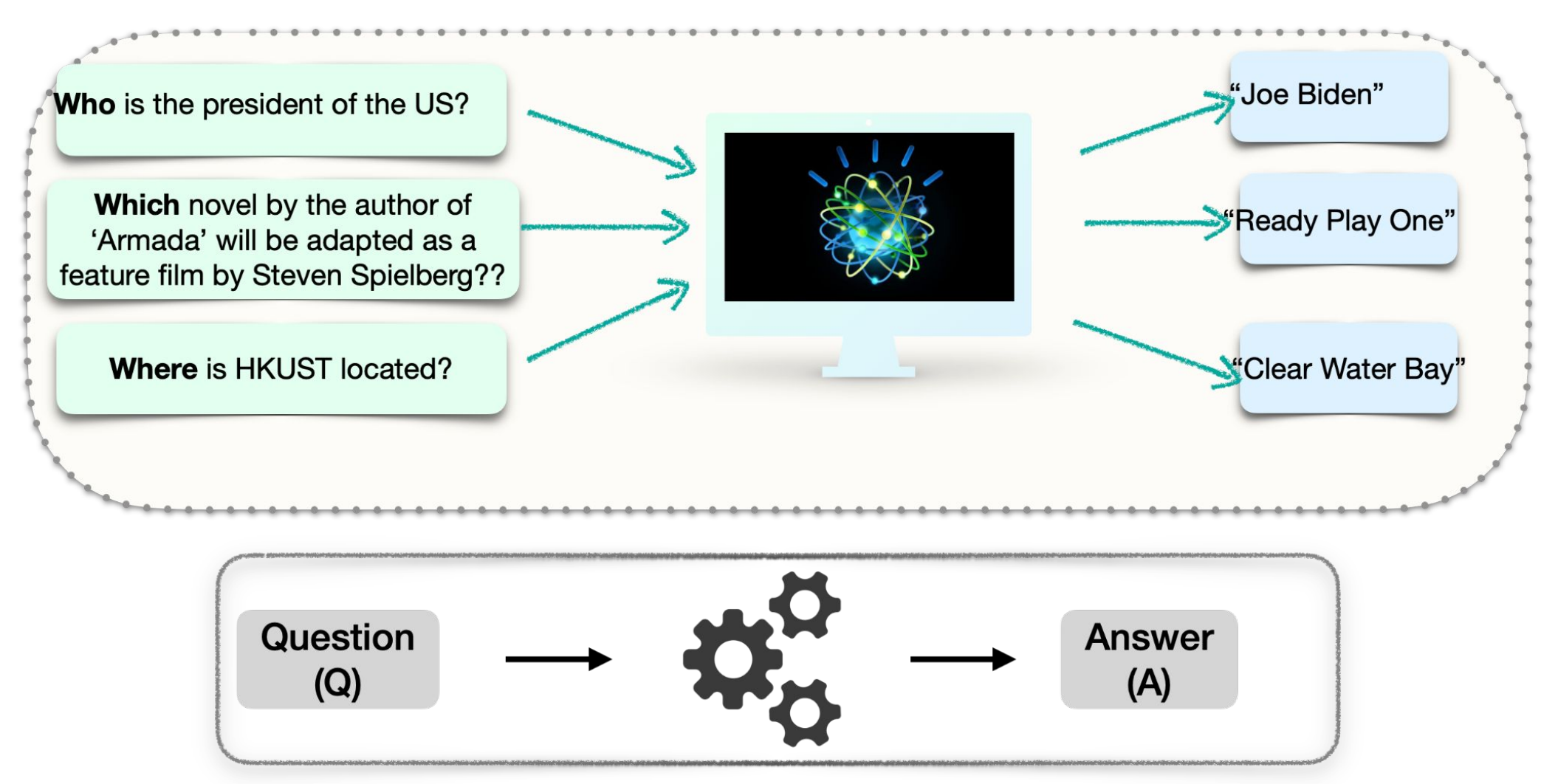}
  \caption{An extractive-based open-domain QA system illustration.}
  \label{fig:intro_extractive_qa}
\end{figure}

% \textbf{Extractive open-domain QA} 
% The most extensively studied ODQA task is \textit{text-based} QA, which derives an answer from unstructured natural language documents (e.g., encyclopedias, dictionaries, news articles, and general Web documents). 
Previous work on ODQA are mainly \textbf{extractive}-based~\cite{chen2017reading,izacard2021leveraging, lewis2021paq}. The system answers the question by extracting a short and concise span from a retrieved document as the answer, as the example in Figure~\ref{fig:intro_extractive_qa} illustrated. The extractive-based ODQA system can provide factoid information for simple straightforward, factual queries such as \textit{Which artist sings this song?} , \textit{Who is the president of the US?}, or \textit{Where is HKUST located?}.

% However, a large ratio of the questions that humans pose to search engines every day are more complicated, such as \textit{How do jellyfish function without a brain?}, \textit{What are the risking factors related to COVID-19?}. Those questions require multi-sentence, in-depth explanations to answer. A short phrases span extracted from existing document by the \textbf{extractive-based} system are incompetent at answering those questions. Therefore, 
Long-form question answering (LFQA), a more challenging task, was recently proposed~\cite{fan2019eli5}. LFQA aims to \textbf{generate} an in-depth, paragraph-length answer for a given question. The questions are more complicated such as the \textit{why/how} type questions, which cannot be directly answered by short-phrase span or sentences from existing documents. Instead, they usually require synthesizing information from multiple web documents comprising hundreds of thousands of words, identifying the relevant information in those documents, and using world knowledge to answer. We show an LFQA examples in Table~\ref{tab:intro_lfqa_example}.

% While the \textbf{extractive} open-domain QA task have been extensively studied in prior years, the \textbf{generative} open-domain QA that tries to generate an answer for a given question obtains little attention. 

% \section{Generative Long-form Question Answering}

% As we can see, it requires multiple sentences to explain the answer for the given question, and the long-form answers are generated from multiple documents and using world knowledge.

\begin{table}[htb!]
    \centering
    \begin{adjustbox}{width={0.9\textwidth},totalheight={\textheight},keepaspectratio}
\begin{tabular}{p{1\columnwidth}}
    \hline
    % \hline
% \textbf{Question:} Which  French ace pilot and adventurer fly L'Oiseau Blanc?  \\
% \textbf{Short answer}: Charles Eugène \\
% \textbf{Long answer}:  On May 8, 1927, \textbf{Charles Nungesser} and Francois Coli boarded L'Oiseau blanc, a 450-hp Lorraine-powered Levasseur biplane designed to compete for the Orteig Prize. They took off from Paris on 8 May 1927 and were last seen over Ireland. Less than two weeks later, Charles Lindbergh successfully made the New York-Paris journey and claimed the prize in the Spirit of St. Louis. \\
% \hline
\textbf{Question:} How do Jellyfish function without brains or nervous systems? \\
\textbf{Answer:} Jellyfish may not have a brain, but they have a rough
nervous system and innate behaviours. However, they are
very simple creatures. They’re invertebrate: creatures without
a backbone. Most jellyfish have really short life spans.
Sometimes just a couple of hours. [...] As their name implies,
they are largely composed of basically jelly inside a
thin membrane. They’re over 95\% water. (327 words) \\
    \hline
    % \hline
    \end{tabular}
    % }
\end{adjustbox}
\caption{Generative long-form question answering example. The example is from the “Explain Like I’m Five” (ELI5)~\cite{fan2019eli5}.}
\label{tab:intro_lfqa_example}
\end{table}

%  The first question is from the HotpotQA dataset~\cite{yang2018HotpotQA}, seeking information for a multi-hop question. As we can see, the extractive QA system tries to extract a short precise answer, the generative-based LFQA aims to generate a long answer. The multi-sentence long answer gives a detailed explanation, which could be used to understand and verify the short answer. 

\section{Motivation}
% Building automated computer systems to answer questions asked by humans in natural language is one of the most elusive and long-standing challenges in the history of artificial intelligence. While we are thrilled at the achievements the community has made~\cite{brill2002analysis, ferrucci2010building, chen2017reading}, the current textual-based ODQA systems have limitations. Almost all of the previous ODQA systems are \textbf{extractive}-based, they can only provide factoid information for simple queries, such as the \textit{who/when/where} type questions, and are incompetent at answering non-trivial questions such as \textit{How do jellyfish function without a brain?}, \textit{Why wouldn't life on another habitable planet look similar to Earth's?}, or \textit{What are the risking factors related to COVID-19?}. However, a large ratio of the questions that humans pose to search engines every day are those complicated types, and require multi-sentence, in-depth explanations. Therefore, it motivates us to work on the newly-proposed, more challenging LFQA task.

Building automated computer systems to answer questions humans ask in natural language is one of the most elusive and long-standing challenges in the history of artificial intelligence. The questions that human tends to ask everyday involves straightforward, factual questions such as “Which artist sings this song?”, a large ratio of the questions are those complicated types, such as \textit{How do jellyfish function without a brain?}, \textit{Why would not life on another habitable planet look similar to Earth's?}, or \textit{What are the risking factors related to COVID-19?}.

While we are thrilled at the achievements the community has made~\cite{brill2002analysis, ferrucci2010building, chen2017reading}, the current extractive-based ODQA systems can not answer non-trivial questions such as the \textit{why/how} type ones. The reasons can be two folds: firstly, there is no short answer for the \textit{why/how} type questions and they require multi-sentence, in-depth explanations; secondly, the answers usually need to be generated by synthesizing information from multiple documents, and a short phrase span extracted from a single existing document by the extractive-based ODQA model cannot answer those complicated questions. Therefore, it motivates us to work on the newly-proposed, more challenging LFQA task.

LFQA has many possible applications. It could be integrated to facilitate a more intelligent search engine ~\cite{metzler2021rethinking}. For example, if we search Google with a query, it could not only returns a list of search documents but also attempts to read these Web documents and even display some generated answers at the top of the search results. Additionally, LFQA could play an important role in digital personal assistants systems such as Amazon's ALEXA, Apple's SIRI, and GOOGLE ASSISTANT to enhance user experience. As it has been shown\footnote{\tt \small https://voicebot.ai/2019/03/12/smart-speaker-owners-agree-that-questions\\-music-and-weather-are-killer-apps-what-comes-next/smart-speaker-use-case-\\frequency-january-2019-01/}, \textit{'asking a question'} is the most frequently used functions from those smart speakers. Specifically, it might be used to generate explanations as a complement to short phrase answers for factoid information-seeking open-domain questions~\cite{kwiatkowski2019natural, yang2018hotpotqa}, or to answer open-ended questions like the ones from the Reddit forum "Explain Like I'm Five"~\cite{fan2019eli5}. 

\begin{table}[!ht]
\centering
\begin{adjustbox}{width={0.99\textwidth},totalheight={0.9\textheight},keepaspectratio}
\begin{tabular}{p{1\columnwidth}}
\hline
\textbf{Question:} what does abc stand for in liquor store? \\
\textbf{Generated Answer:} \hlgreen{\textbf{A liquor store} is a retail shop that predominantly sells prepackaged alcoholic beverages intended to be consumed off the store's premises. Depending on region and local idiom (social issue), they may also be called bottle store, off licence, (in United Kingdom, Ireland, and New Zealand) bottle shop, bottle-o package store, party store (in Michigan), ABC store, state store, or other similar terms.}\\
% \textbf{RBG:} ABC stands for Alcoholic Beverages Control (ABC Stores) and is a type of liquor store that is owned and operated exclusively by the state government, where liquor
% stores often sell only spirits or sometimes sell spirits and wine but not beer.\\
\textbf{Golden Answer:} Most of these states have an \textbf{Alcoholic Beverage Control (ABC)} board and run liquor stores called ABC stores. In all monopoly states a parallel license system is used to regulate the sale and distribution of lighter alcoholic beverages such as beer and wine.\\
\hline
\end{tabular}
\end{adjustbox}
\caption{Example from the MSMARCO dataset~\cite{nguyen2016ms}. The generated answer gives a detailed definition and introduction of 'liquor store', however, it doesn't answer the representation of 'abc'. Therefore, the answer lacks of \textbf{relevance} to the given question. We highlight in green the sentences \hlgreen{irrelevant} to the question.}
\label{tab:intro_challenge_examples_1}
\end{table}

\begin{table}[!ht]
\centering
\begin{adjustbox}{width={0.99\textwidth},totalheight={0.9\textheight},keepaspectratio}
\begin{tabular}{p{1\columnwidth}}
\hline
\textbf{Question:} definition of sadduction \\
\textbf{Generated Answer:} Sadducees. noun. \hlred{The definition of a sadducee is a person who acts in a deceitful or duplicitous manner.  An example of a Sadduceee is a politician who acts deceitfully in order to gain political power}. 1 a member of a Jewish sect that was active during the Second Temple. \\
% \textbf{RBG:} The Sadducees were a sect or group of Jews that were active in Judea during the Second Temple period, starting from the second century BCE through the destruction of the Temple in 70 CE. The sect was identified by Josephus with the upper social and economic ecoshon of Judean society. \\
\textbf{Golden Answer:} The Sadducees (Hebrew: \d{S}ĕdûqîm) were a sect or group of Jews that were active in Judea during the Second Temple period, starting from the second century BCE through the destruction of the Temple in 70 CE. The sect was identified by Josephus with the upper social and economic ecoshon of Judean society.\\
 \hline
\end{tabular}
\end{adjustbox}
\caption{Example from the MSMARCO dataset~\cite{nguyen2016ms}. The answer is hallucinated and lacks of \textbf{faithfulness}. We mark in red the \hlred{unfaithful} snippets.}
\label{tab:intro_challenge_examples_2}
\end{table}

\begin{table}[!ht]
    \centering
    \begin{adjustbox}{width={0.92\textwidth},totalheight={0.9\textheight},keepaspectratio}
\begin{tabular}{p{1\columnwidth}}
    \hline
\textbf{Question:} Which  French ace pilot and adventurer fly L'Oiseau Blanc?  \\
\textbf{Long answer}:  On May 8, 1927, \hlorange{\textbf{Charles Nungesser}} and Francois Coli boarded L'Oiseau blanc, a 450-hp Lorraine-powered Levasseur biplane designed to compete for the Orteig Prize. They took off from Paris on 8 May 1927 and were last seen over Ireland. Less than two weeks later, Charles Lindbergh successfully made the New York-Paris journey and claimed the prize in the Spirit of St. Louis. \\
\textbf{Short answer}: Charles Eugène \\
\hline
    \end{tabular}
    % }
\end{adjustbox}
\caption{For question types such as \textit{who/which/when/where}, the short answer that is more succinct, might be more preferred. The example is from the HotpotQA dataset~\cite{yang2018hotpotqa}}
\label{tab:intro_challenge_example_3}
\end{table}

\section{Problem Setup}

The task of LFQA can be formulated as: given a collection of training examples ${(q_i,a_i)}_{i=1}^n$, the goal is to learn a predictor $\mathcal{F}$, which takes the question $q$ as input and output the answer $a$.

\begin{equation}
    \mathcal{F}: q->a \\
\end{equation}

$a=\left\{a_{1},a_{2,}...,a_{l_{a}}\right\},\,\text{where}\;\,a_{i}\in\mathbb{V}\,\text{for}\;i=1,...,l_{a}, q=\left\{q_{1},q_{2},...,q_{l_{q}}\right\}\,\text{where}\;q_{i}\,\in\,\mathbb{V}\,\text{for}\;i=1,...l_{q}$

Usually we will also find some supporting information from large knowledge source, such as Wikipedia. But the support information is not provided in the training examples.
\begin{equation}
    \mathcal{F}: (q, \mathcal{D})->a \\
\end{equation}

In this thesis, we will mainly focus on answering the questions $q$ based on information from the external knowledge corpus $\mathcal{D}$, and we will not answer questions that is unanswerable or 'out-of-domain' of supporting document $\mathcal{D}$.

The evaluation metrics include (1) Automatic metrics, such as ROUGE, which calculate the overlap of ngrams between the generated and reference answers (2) Human evaluation, which normally evaluate the fluency, relevance, and factual correctness of the generated answer.

\section{Research Directions}

\begin{figure}[!ht]
  \centering
  \includegraphics[width=0.9\linewidth]{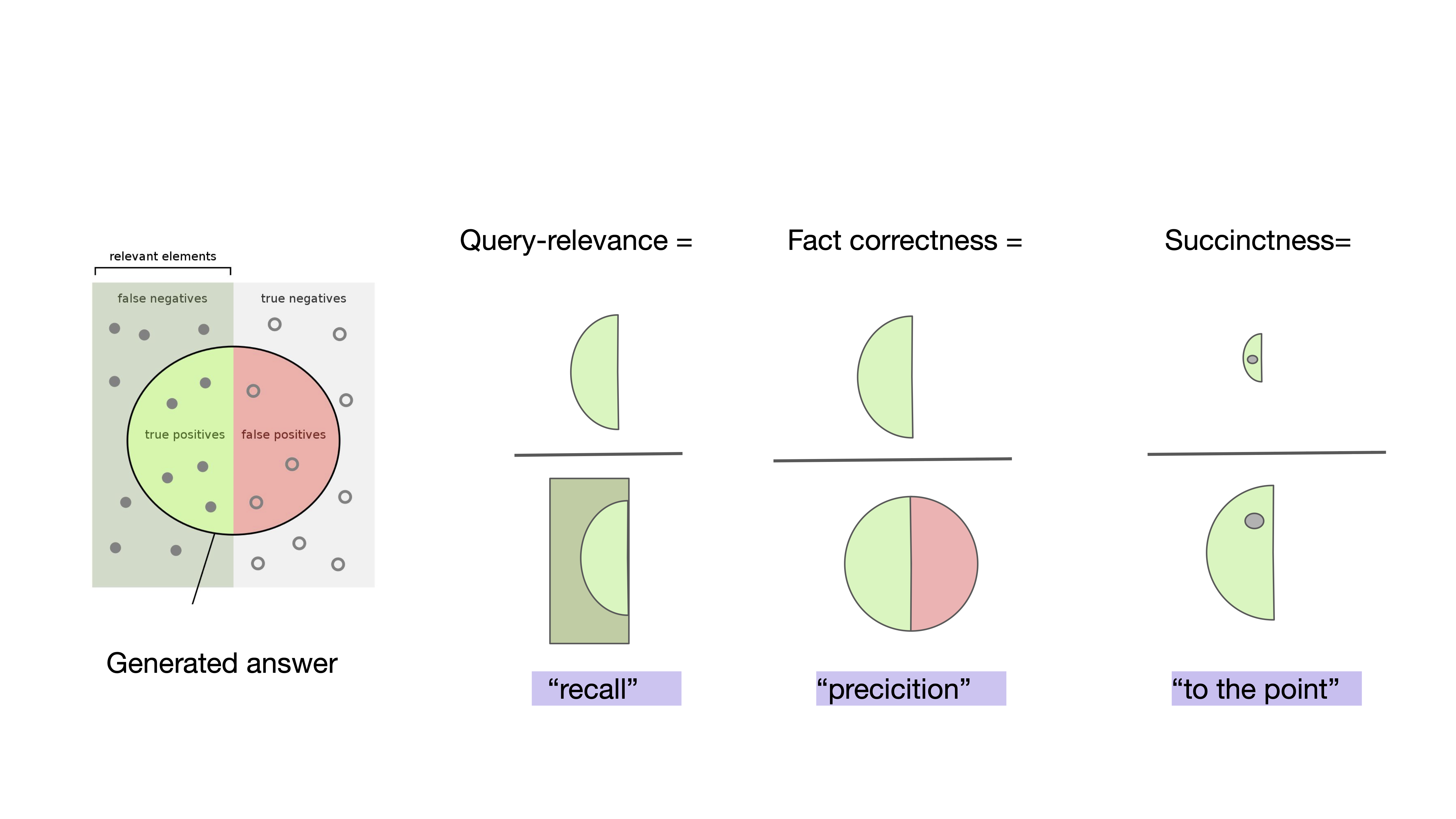}
  \caption{Illustration of 1) query-relevance, 2) fact-correctness and 3) succinctness, using similar information retrieval metrics.}
  \label{fig:intro_concept}
\end{figure}

LFQA is a newly proposed, more challenging task compared to the traditional extractive-based open-domain QA, and little work has yet been done on it. In this thesis, we focus on the three corresponding research directions: 1) generating query-relevant long-form answers 2) generating fact-aware long-form answers 3) generating succinct answers from long-form answers. We show an simple but concrete illustration of the three aspects in Figure~\ref{fig:intro_concept}.

In this thesis, we study methods to build and improve a LFQA system based on the three aspects. Specifically, we focus on:

\begin{itemize}
    \item \textbf{Generating query-relevant long-form answers} In recent years, we have witnessed rapid progress in dense retrieval techniques like REALM~\cite{guu2020realm}, DPR~\cite{karpukhin2020dense} and ORQA~\cite{lee2019latent}. Also, the large pre-trained language models~\cite{lewis2020bart,raffel2020exploring} have demonstrated its superior performances at many generation tasks such as summerization~\cite{su2021improve,yu2020dimsum}, question generation~\cite{su2022qa4qg,su2020multi}. However, little work has been investigated to build an effective real-time LFQA system. Since LFQA is quite a challenging task that requires more than just concatenating the two techniques. Firstly, it needs to handle the issue of applying a pre-trained language model with limited input sequence length, to process the multiple relevant documents containing more than a thousand tokens for generation. Secondly, those multiple retrieved documents may include irrelevant, complementary, or contradictory information; how to generate more query-relevant, long-form answers is another challenge (we show an example in Table~\ref{tab:intro_challenge_examples_1} to provide a better understanding). Therefore, we first propose CAiRE-COVID, a real-time long-form question answering system for COVID-19, which combines information retrieval with state-of-the-art QA and query-focused multi-document summarization techniques to answer high-priority COVID-19-related questions. Then, we introduce QFS-BART, a model that incorporates the explicit answer relevance of the source documents given a query into the answer generation model to generate more query-relevant answers.

    % Fluency is the first important evaluation element in many generation tasks, such as abstractive summarization~\cite{su2021improve}, text generation~\cite{su2020multi,su2022read}. It measures the linguistic quality of the generated text. If a long-form answer that is not fluent, such as containing repetitions or grammatical incorrect statements, it will harm the answer contents greatly. 
    
    % Furthermore, the generated answer should contain information that is relevant to the topic of the question. We show an example in Table~\ref{tab:intro_challenge_examples_1} to provide a better understanding. The generated answer gives a detailed definition and introduction of 'liquor store' in a fluent way, however, it lacks \textbf{query-relevance} since the information does not answer the question: \textit{'What does abc stand for?'}. 

    % Therefore, how to generate fluent, query-relevant long-form answers is an important task to build an effective LFQA system.

    \item \textbf{Generating fact-aware long-form answers} Improving the factual correctness of the generated answer is one of the other main challenges for LFQA task~\cite{krishna2021hurdles, lin2021truthfulqa, nakano2021webgpt}. Specifically, the generated answer should be faithful to facts in the source documents~\cite{ji2022survey}. In the example in Table~\ref{tab:intro_challenge_examples_2}, the answer is fluent and relevant to the query when trying to elaborate the \textit{'definition of sadduction'}. However, the content itself is hallucinated and lacks \textbf{faithfulness}. Such a fluent and relevant but unfaithful answer will mislead the user. Therefore, we propose an end-to-end framework named RBG to augment the generation model with fine-grained, answer-related salient information predicted by a machine reading comprehension(MRC) module, to improve and generate a fact-aware answer that is more faithful to the source documents.
    
    \item \textbf{Generating Succinct Answers from Long-form answers}
    % As illustrated in Figure~\ref{fig:intro_extractive_qa}, a traditional open-domain QA system normally retrieves multiple relevant contexts from an external knowledge corpus, then it extracts a short answer span from the retrieved contexts as the answer. 
    A LFQA system should be an ‘empathetic machine’: the long-form answer should not only provide relevant and factual information, but also be succinct. For questions like \textit{who/what/when/where/which} types, people might prefer a more concise answer, as shown in Figure~\ref{tab:intro_challenge_example_3}. Therefore, how to generate a succinct answer is also an important direction. 
    
    While no prior work has been investigated, we take the initial step by leveraging the generated long-form answers as an context to extract succinct short-phrase answers for \textit{closed-book} question answering (CBQA) task. Specifically, we generate a long-form answer leveraging the amount of parameterized knowledge stored in pre-trained language models, and then extract the short-span answer from the long-form answers without access to any external knowledge sources. Experimental results on three QA benchmarks show that our method significantly outperforms previous \textit{closed-book} QA methods, and is on par with traditional \textit{open-book} methods which extracts the answer from the retrieved documents.
% . 
\end{itemize}

\section{Thesis Outline}
The thesis is divided in four main chapters, plus a conclusion. In Chapter 2 first gives an overview of open-domain question answering. Then we will introduce the preliminaries needed throughout the thesis, followed by the related work to LFQA. In Chapter 3, we describes how to build a LFQA system and generate more query-relevant long-form answers. In Chapter 4, we describes how we generate more fact-aware long-form answers. In Chapter 5, we show how we further generate succinct answers from the long-form answers. Finally, in Chapter 6, we summarize the thesis and we discuss possible future research directions.

\section{Contributions}
The contributions of this thesis are summarized as follows:
\begin{itemize}
% \item We made the effort and we are the first to tackle the faithfulness issue of LFQA. We proposed a framework RBG, which do dynamic global salient information prediction from multiple source documents and fusion-in-decoder, to improve the factual correctness of the generated answer. State-of-the-art results have been obtained.  We also topped public leaderboard on LFQA task!

% \item We pioneered the research direction and we are the first to extract succinct answers from generated long-form answer. Our method outperforms prior closed-book QA methods, and on-par with state-of-the-art open-book QA method that explicit retrieve from external DBs, on three open-domain QA datasets.

% \item We set out to tackle and improve the query-relevance of the generated answer. We are the first to build an LFQA system for COVID-19. The system is Kaggle competition winner! We further improve query-relevance and obtained state-of-the-art results.

\item We were among the first to research the LFQA task, and we pioneered the research direction to improve the answer quality in terms of 1) query-relevance, 2) answer faithfulness, and 3) succinctness. 

\item We investigated the core challenges to high answer quality in the LFQA task, in terms of the three aspects. Specifically,
\begin{itemize}
    \item we propose a coarse-to-fine method to extract the document-level and sentence-level query-relevant information, to help a traditional Seq2Seq model to handle long and multiple documents, and consider query-relevance. We further introduce QFS-BART, a model that incorporates the explicit answer relevance attention of the source documents into Seq2Seq model's encoder-decoder attention, to further enhance the answer’s relevance.

    \item we proposed a framework RBG, which does dynamic global salient information prediction from multiple source documents and fusion-in-decoder, to improve the factual correctness of the generated answer.

    \item we propose a two-stage method, to leverage generated contexts to further extract a succinct answer. Specifically, we generate a long-form answer leveraging the amount of parameterized knowledge stored in pre-trained language models~\cite{raffel2020exploring, brown2020language, ye2020studying}, and extract a short-phrase span answer from the generated long-form answer without access to any external knowledge sources.
\end{itemize}
  
\item State-of-the-art results have been obtained on large-scale LFQA datasets, by automatic and human evaluation. We topped the only public leaderboard on LFQA task! And we are the first to build an LFQA system for COVID-19. The system is Kaggle competition winner! 
\end{itemize}

\chapter{Background and Preliminaries}

In this chapter, we first give a brief introduction to open-domain question answering task. Then, we will introduce the preliminary technologies involved in our work for generative-based LFQA task, followed by related work.

\section{Background: Open-domain Question Answering}

Open-domain question answering has been an important research topic in the history of NLP~\cite{harabagiu2000experiments,lin2003question,gliozzo2012natural, yih2016question, sachan2018standardized, chen2017reading}. The goal of open-domain question answering is to build computer systems to answer any sort of (factoid) questions that humans might ask automatically, based on a large collection of unstructured natural language documents, structured data, semi-structured data or even other modalities such as images or videos~\cite{chen2018neural}. 

\subsection{A Brief History of Open-domain QA}
Dating back to the 1960s~\cite{simmons1964indexing}, the task of open-domain question answering has been broadly investigated by both the Information retrieval (IR) and natural language processing (NLP) communities.

\begin{figure*}[!th]
    \centering
    \includegraphics[width=0.8\linewidth]{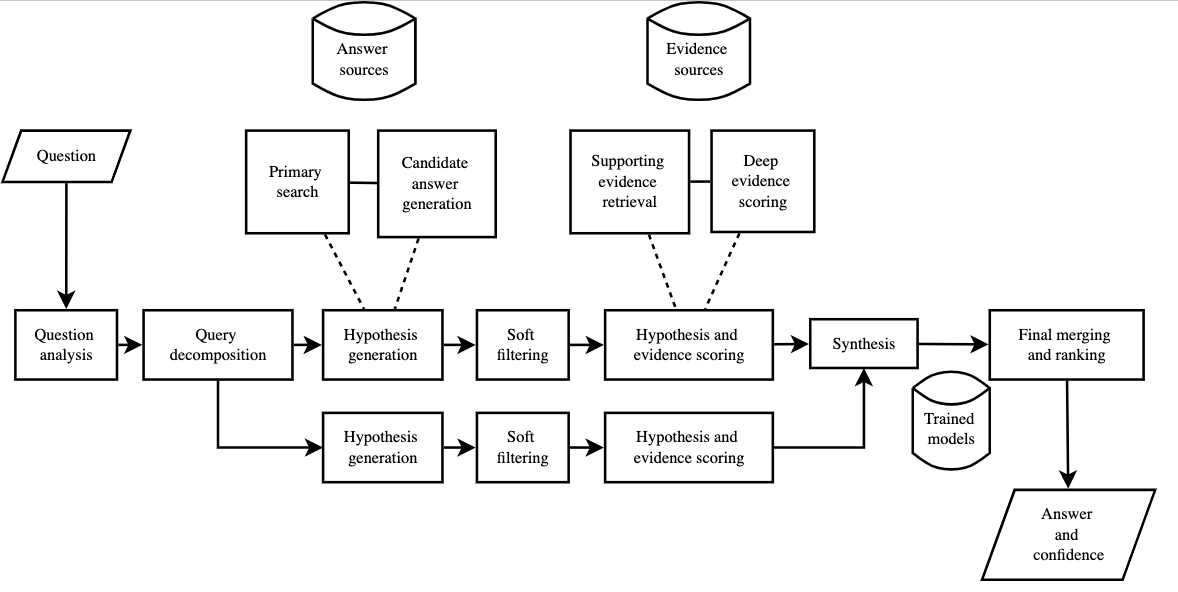}
    \caption{The high-level architecture of IBM’s DEEPQA~\cite{ferrucci2010building} used in WATSON. Image courtesy: \url{https://en.wikipedia.org/wiki/Watson (computer)}.}
    \label{fig:deepQA}
\end{figure*}

\begin{figure*}[!ht]
    \centering
    \includegraphics[width=0.75\linewidth]{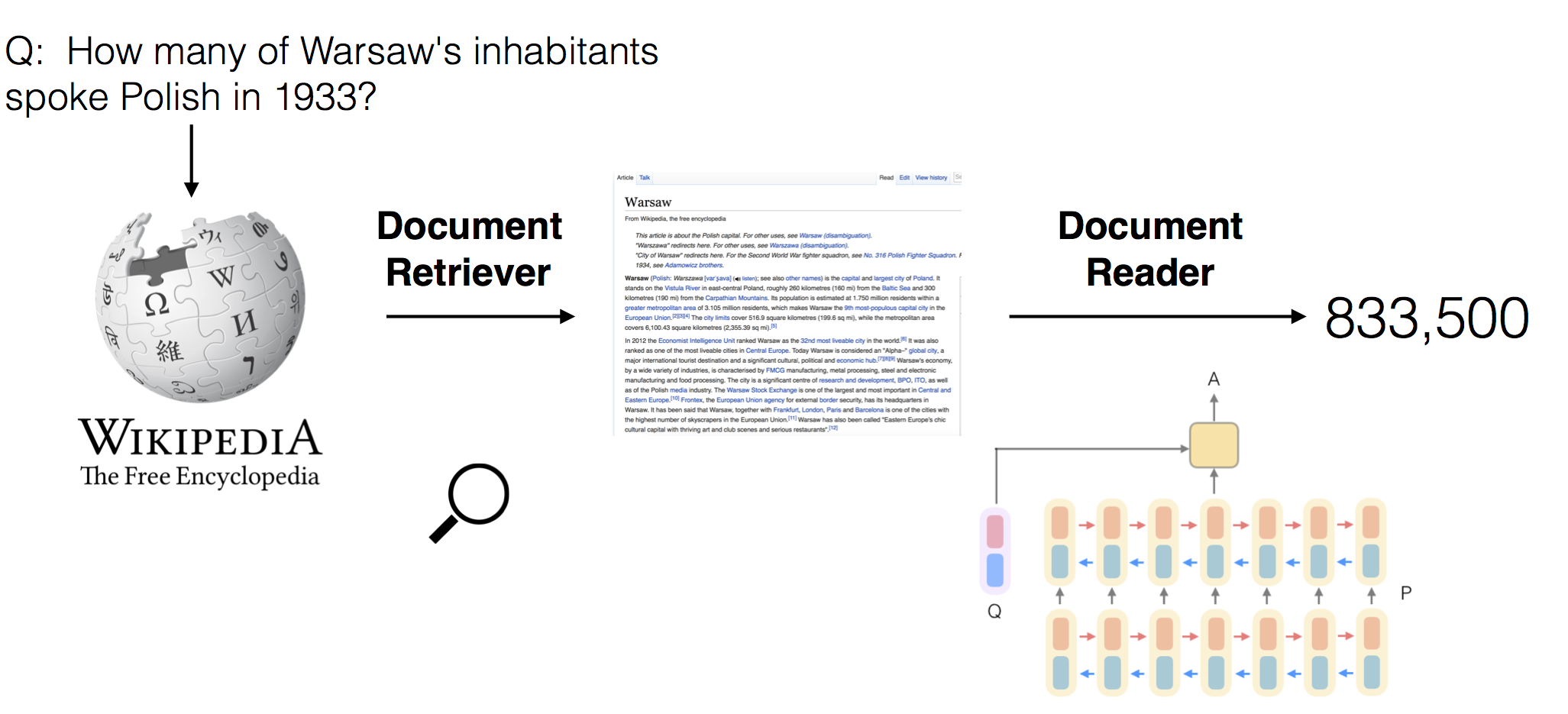}
    \caption{The two-stage \textit{retriever-reader} architecture of DrQA~\cite{chen2017reading}. Image courtesy: \url{https://github.com/facebookresearch/DrQA}.}
    \label{fig:drqa}
\end{figure*}

Some notable textual-based QA systems include Microsoft’s ASKMSR~\cite{brill2002analysis}, IBM’s DEEPQA~\cite{ferrucci2010building}, and DrQA~\cite{chen2017reading}. ASKMSR is a search-engine based QA system that relies on "web redundancy" rather than complicated linguistic analyses of questions or documents, while DEEPQA is the most representative modern QA system which consisting of many different pieces in a pipeline (as we show in Figure ~\ref{fig:deepQA}). The IBM DEEPQA's victory at the TV game-show JEOPARDY! in 2011 received a lot of attention, and rekindle the research interest in QA. Compared to DEEPQA's sophisticated system, the newly proposed DrQA system used a two-stage framework, integrating a classical retrieval module and a neural machine reading comprehension (MRC) component(as shown in Figure ~\ref{fig:drqa}). DrQA is designed
to answer questions from English Wikipedia and is also the first neural QA system. The two-stage \textit{retriever-reader} framework become dominated paradigm for open-domain QA~\cite{izacard2021leveraging,NEURIPS2020_6b493230}, and we have seen significant progress in recent years.

Different kinds of resources might be involved in order to answer the questions. These resources could be \textbf{unstructured natural language documents} (encyclopedias, dictionaries, news articles and general Web documents), \textbf{structured data} (e.g., knowledge bases), or \textbf{semi-structured data}(e.g., tables) or even \textbf{other modalities} such as videos or images.

The questions posed by humans could be \textbf{simple questions(single-hop questions)} seeking for factoid information, such as \textit{"Who is the president of the US?"}, or might be \textbf{multi-hop questions} which need to aggregate information from multiple places and reasoning over them to answer, like \textit{“Which novel by the author of ‘Armada’ will be adapted as a feature film by Steven Spielberg?”}, or even more \textbf{complicated questions} such as the \textit{how/why} type questions.

The types of answers also varies a lot. The answer might be a short-phrase span \textbf{extracted} from the given passage, or \textbf{multiple-choice} type, or even free-text form \textbf{generated} from the system.

\subsection{Different Open-domain QA Tasks}

According to the difference in resources involved to answer questions or question types, there is also a difference in the relationships between the questions and the corresponding resources, along with different answer types (e.g., a text span, a ranked list of passages, cloze style, multiple choice answer, free-form answer). As a result, a QA can be formulated and categorised into different types, where the different application scenarios may differ depending of the type of QA.

QA can be categorized into the following categories via the format of the \textit{knowledge resources}:
\begin{itemize}
    \item \textbf{Conversational QA} is a QA task involving comprehension of both passage and conversational QA histories~\cite{reddy2019coqa}.It is proposed to mimic the way human seek information in conversations ~\cite{ju2019technical}. It will play a crucial role in conversational AI systems. 
    \item \textbf{Text-based QA} aims to answer questions based on unstructured natural language documents. It puts more emphasis on text understanding with answering questions regarded as a way to measure language understanding, and can also be interpreted as machine reading comprehension problem. One of the most representative text-based QA task is the SQuAD ~\cite{rajpurkar2016squad}.
    \item \textbf{Knowledge-based QA} (KBQA) analyzes query question and then finds the answer from the knowledge base/ knowledge graphs (KG)~\cite{yao2014information}. It help end users more efficiently and more easily access the substantial and valuable knowledge in the KG without knowing its data structure. Examples of a few large KGs include Wikidata (Google, 2013 ~\cite{vrandevcic2014wikidata}), DBPedia (~\cite{auer2007dbpedia}), Yago (~\cite{suchanek2007yago}), and NELL~\cite{mitchell2018never}.
    \item \textbf{Table-based QA} aims to answer complex questions on semi-structured tables, instead of a fixed database ~\cite{pasupat2015compositional}. Different questions could be asked on different tables with different schemas. 
    % \item \textbf{Community-based QA} (cQA) are defined as dedicated platforms for users to respond to other users' questions, resulting in the building of a community where users share and interactively give ratings to questions and answers (Liu et al., 2008). CQA services are emerging as a valuable information resource that is rich not only in the expertise of the user community but also their interactions and insights. 
    \item \textbf{Visual QA} (VQA) aims to answer questions about images or vidoes~\cite{balanced_vqa_v2}. The questions require an an understanding of vision, language and commonsense knowledge to answer.

\end{itemize}

\noindent Depends on the formats of the \textit{answer}, QA can also be categorised into: 
\begin{itemize}
    \item \textbf{Cloze-style}: The question contains a placeholder, and the systems must guess which word or entity completes the sentence (question), based on the passage.
    \item \textbf{Multiple-choice}: The correct answer is chosen from k hypothesized answers(e.g., k = 4).
    \item \textbf{Span-based QA}: It can also be referred to as extractive QA and the answer must be a single span from the passage.
    \item \textbf{Free-form answer QA}: The answer is allowed to be any free-text form (i.e., a word sequence of arbitrary length).
\end{itemize}

\subsection{Open-domain QA Frameworks}

The frameworks for open-domain QA can be roughly categorized into two classes: the \textit{open-book} QA methods, which exploit an external knowledge source to answer the question, and the \textit{closed-book} QA (CBQA) methods ~\cite{roberts2020much}, which tries to directly answer the open-domain questions without access to any external knowledge sources, and leverages the parametric knowledge stored in large pretrained language models (LMs)~\cite{raffel2020exploring, brown2020language, ye2020studying}. 

\subsubsection{The Open-book QA Method}
\label{sec:two-stage}
For text-based open-domain QA, the majority of the systems consisted of two stages: a \textbf{document retriever}, which is normally an information retrieval (IR) system used to retrieve most related documents/paragraphs a question, and a \textbf{document reader} module, which normally cast as a machine reading comprehension (MRC) task, to predict the answer spans from the selected candidates paragraphs. Thus, most work in (text-based) open-domain QA are mainly focus on the two tasks: 
\begin{itemize}
    \item \textit{Information retrieval (IR)}: how to effectively and efficiently (considering both precision and recall metrics) retrieve the related candidate documents,
    \item \textit{Machine reading comprehension(MRC)}: accurate answer span prediction and ranking (by reasoning over multi-documents).
\end{itemize}

The \textit{open-book} QA method have obtained state-of-the-art results on open-domain QA benchmark datasets. The most representative work of the open-book QA methods, include RAG~\cite{NEURIPS2020_6b493230} and FiD~\cite{izacard2021leveraging}. 

\subsubsection{The Closed-book QA Method}
Another branch of work ~\cite{petroni2019language, radford2019language, brown2020language,roberts2020much} focused on the retrieval-free approach, trying to use the large pre-trained language models such as  BERT~\cite{petroni2019language}, GPT-2~\cite{radford2019language}, GPT-3~\cite{brown2020language}, T5~\cite{roberts2020much} as 'knowledge storage' to get the answer, instead of explicitly storing all the text and searching among their dense or sparse representations. Since the LMs were pre-trained on Wikipedia (and other textual corpora) so they should be able to memorize a fair amount of information. However, the \textit{closed-book} methods are not competitive with \textit{open-book} models in term of accuracy.

\section{Preliminaries}
In this part, we will introduce the techniques required for our work in building an effective LFQA system.

\subsection{Sequence-to-Sequence (Seq2Seq)}
Sequence-to-Sequence (or Seq2Seq) is a neural network that transforms a given input sequence into another sequence. For example, the input could be a sequence of English words, the output could be another sequence of German words. If we denote the input and output sequence as $X = \{x_1,..., x_n\}$, and $Y = \{y_1,..., y_m\}$, the objective of a Seq2Seq model is to approximate the conditional probability $P(Y|X)$. Formally, a Seq2Seq model generates a sequence of probability distributions, each of which is conditioned on the previously generated tokens and the input sequence X, and the model can be formulated as:
\begin{equation}
    P_\theta(Y|X) = \sum_{i=0}^mP_\theta(y_i|y_0,...,y_{i-1};x_0,...,x_n)
\end{equation}

where $\theta$ denotes the model parameter.

Seq2Seq models consist of an Encoder and a Decoder. The Encoder takes the input sequence and maps it into a higher dimensional space (n-dimensional vector),
\begin{equation}
   h_{enc} = \text{Encoder}(x_1,...,x_n) 
\end{equation}

where $h_{enc} = \{h_{enc}^1, ...h_{enc}^i,..., h_{enc}^n\}$, and $h_{enc}^i \in R^{d}$ is a d-dimensional vector. Then the encoder output is fed into the Decoder which turns it into an output sequence. The output sequence can be in another language, symbols, a copy of the input etc., and can be simply formulated as:
\begin{equation}
    p(y_i) = \text{Decoder}(y_0,...,y_{i-1};h_{enc})
    \label{eq: seq2seq}
\end{equation}

While the Encoder and Decoder can adopt different neural architectures, such as the Recurrent Neural Network(RNN), Long-Term Short Term Memory(LSTM)~\cite{hochreiter1997long} or Gated Recurrent Unit (GRU)~\cite{cho2014learning}, it can also employ the non-recurrent-based network, such as the Transformer we are going to introduce in Section ~\ref{sec:transformer}.

\subsection{Transformer}
\label{sec:transformer}
The Transformer was proposed in the paper \textit{Attention is All You Need}~\cite{NIPS2017_3f5ee243}. It is an only attention-mechanism-based Seq2Seq model, without any RNN (Recurrent Neural Networks), thus it can be computed via parallelizable operation.

\begin{figure*}[!ht]
    \centering
    \includegraphics[width=0.7\linewidth]{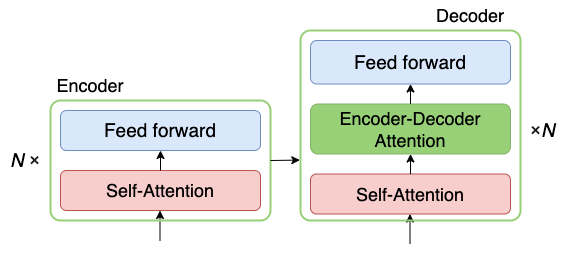}
    \caption{A simple illustration of Transformer (part).}
    \label{fig:transformer}
\end{figure*}

Both Encoder and Decoder are composed of modules that can be stacked on top of each other multiple times, which is described by $N$. We see that the modules consist mainly of Attention and Feed Forward layers ( as shown in Figure ~\ref{fig:transformer}). 

The attention mechanism in both the Encoder and Decoder can be illustrated as:

\begin{equation}
    Attention(Q, K, V) = softmax(\frac{QK^{T}}{ \sqrt{d}}V),
\end{equation}

$Q$ is a matrix that contains the query (vector representation of one word in the sequence), $K$ are all the keys (vector representations of all the words in the sequence) and $V$ are the values, which are again the vector representations of all the words in the sequence.

While for the \textbf{Self-Attention} in both the Encoder and the Decoder, the query $Q$ is the same as $K$, which is also equal to $V$. However, in the \textbf{Encoder-Decoder Attention} layer in the Decoder block, which is also called \textbf{Cross-Attention}, the query $Q$ will be from the Decoder, while $K$ and $V$ are from the Encoder sequence. The cross-attention mechanism can help the decoder focus on relevant parts of the input sentence.

\subsection{Pre-trained Language Models}
%BERT: for MRC
%BART: for Generation

In general, pre-trained models are divided into bi-directional such as BERT~\cite{devlin2019bert}, uni-directional or casual-decoder (left-to-right decoder) such as the GPT model~\cite{radford2019language, brown2020language}, and encoder-decoder generative model like BART~\cite{lewis2020bart} and T5~\cite{2019t5}. Those models have achieved state-of-the-art performances in many natural language understanding tasks~\cite{wang-etal-2018-glue}. 

The bi-directional pre-trained models are trained with a masked-language model (MLM) loss, which learns how to predict words that are randomly masked in the input. While the uni-directional and encoder-decoder generative models are usually trained using the likelihood function in Equation ~\ref{eq: seq2seq}. 

In this thesis, we mostly focus on encoder-decoder based generative models. The encoder-decoder based pre-trained LM models are trained on huge amount of unlabeled data. Many different pre-training strategies has been proposed based on the task objective. For example, BART used span-prediction and denoising pre-training~\cite{lewis2020bart}: the input sequence is corrupted and the model is taught to reconstruct the original sequence.

\subsection{Retrieval Methods}
As we mentioned in Section ~\ref{sec:two-stage}, the document retriever component will retrieve the most related documents/paragraphs given a question, from a large-scale external document collections. 

Traditional methods use the \textbf{sparse retrieval} based approach (TF-IDF or BM25) to get a candidate set. For example, the TF-IDF weighted term vector model over unigrams/bigrams can be formulated as follows:
\begin{align}
    \text{tf-idf}(t,D_i,D) &= \text{tf}(t,D_i) \cdot \text{idf}(t,D)
    \\
    \text{tf}(t,D_i) &= \text{log}(1+\text{freq}(t,D_i)) \\
    \text{idf}(t,D) &= \text{log}(\frac{|D|}{|d\in D: t \in D_i}|)
\end{align}

where we use $t$ denote the term(unigrams/bibrams) in query $Q$, and $D_i$ refer to one document such as one wikipedia article, and $D$ is the whole external knowledge sources. tf means term frequency, and idf represents the inverse document frequency.

More recently, better results~\cite{karpukhin2020dense,guu2020realm} have been obtained with \textbf{dense-only retrieval} methods. Basically the model pre-encodes all docoments $D$ with large-pretrained language models such as BERT~\cite{devlin2019bert}, then encodes question $Q$ into the same space at test time, retrieves with simple nearest-neighbor search: 

\begin{equation}
    sim(Q, D_i) = \text{BERT}_q(Q)^T \text{BERT}_d(D_i)
\end{equation}

Compared to the traditional sparse-based retrieval methods, the dense retrieval techniques enables: (1) a trainable retriever, (2) dense representations of the documents and the query, which leads to better retrieval performance. Currently there are techniques and tools to support fast maximum inner product search (MIPS), such as the approximate nearest neighbors search in the FAISS~\footnote{github.com/facebookresearch/faiss} library, which uses in-memory data structure and indexing schemes, the dense retrieval can be done quite efficiently.

\subsection{Machine Reading Comprehension}

The task of Machine Reading Comprehension (MRC) - answering comprehension questions over a passage of text, has a long history, and was first studied in 1970s~\cite{lehnert1977process}. Traditional methods used manually generated rules, or feature-based methods~\cite{chen2016thorough}. Recently, the field has witnessed great success, with the emergence and development of neural (deep-learning) reading comprehension models, especially the large pre-trained language models like BERT~\cite{devlin2019bert}, XLNET~\cite{yang2019xlnet}, and the creation of large-scale supervised datasets in the form of (passage, question, answer)
triples such as SQuAD~\cite{rajpurkar2016squad}. The dominant way is to fine-tune the large pre-trained language models by MRC datasets. Most recently, ~\cite{lovenia2022clozer} proposed Clozer, a sequence-tagging based cloze answer extraction method used to extende task adaptive pretraining (TAPT)~\cite{zhang2021hybrid} for adaptation on cloze-style machine reading comprehension. 

The pre-trained language model normally takes the concatenation of the passage $D$ and question $Q$ as input, and outputs the prediction of the start and end position of the potential evidence spans in the document $D$. Specifically, it outputs two probability distributions over the tokens in $D$: $P^s(i)$ and $P^e(i)$, where $P^s(i)$ / $P^e(i)$ is the probability that the $i$-th token is the start/end of the evidence span in $D$. The score of a candidate span from
position $i$ to position $j$ is defined as $P_i + P_j$, and the maximum scoring span where $j >= i$ is used as a prediction. The performance of the state-of-the-art model on the task has already surpass human~\cite{devlin2019bert,yang2019xlnet}.

\section{Related Work}

\subsection{Open-domain QA} 
Open-domain QA is the task of answering general-domain questions~\cite{chen2017reading}, where the evidence is usually not given. Models that explicitly exploit an external corpus are referred to as \textit{open-book} models~\cite{roberts2020much}. They typically index the corpus and then \textit{retrieve-and-read} to extract the answer span from documents~\cite{chen2017reading, lee2019latent, izacard2021leveraging, rag, lazaridou2022internet}. Another recently proposed class of methods is \textit{closed-book} QA models. ~\citet{ye2020studying,roberts2020much} finetune pretrained LMs such as T5~\cite{raffel2020exploring} or BART~\cite{lewis2020bart} with QA pairs without access to any external knowledge or context.

\subsection{Machine Reading Comprehension}
The goal of a Machine Reading Comprehension (MRC) system is to answer a question by reading one or more context documents. Many work has been done for MRC in recent years, fueled by the creation of many large-scale datasets~\cite{rajpurkar2016squad, lai2017race, saha2018duorc, trischler2017newsqa, Joshi_2017}. 

We have witnessed several breakthroughs in MRC models, such as bidirectional attention flow (BiDAF)~\cite{seo2017bidirectional}, the attention over attention mechanism (AoA)  ~\cite{cui2017attention}, and a multi-hop architecture using gated-attention readers ~\cite{dhingra2017gated}. Recently, fine-tuning pre-trained language models~\cite{devlin2019bert, yang2019xlnet} via supervised learning has achieved the state-of-the-art performance on many MRC datasets. 

While most previous methods focus on improving in-domain performance, ~\citet{su-etal-2019-generalizing} propose a multi-task learning framework that learns the shared representation across different tasks, to build a QA system which has general linguistic intelligence, so that the model can generalize to out-of-domain and unseen tasks.

\subsection{Query-focused Summarization}
Query-focused Summarization (QFS) aims to generate a summary according to the query and the provided relevant document(s)~\cite{tombros1998advantages}. The input can be either a \textit{single} document that has multiple views or \textit{multiple} documents that contain multiple topics, and the output summary should be focused on the given query. QFS has various applications (e.g., a personalized search engine that provides the user with an overview summary based on their query~\cite{su2020caire}).

Early work on the QFS task mainly focused on generating extractive summaries~\cite{davis2012occams, daume2006bayesian, feigenblat2017unsupervised, xu2020coarse}, which may contain unreadable sentence ordering and lack cohesiveness. Other work on abstractive QFS incorporated the query relevance into existing neural summarization models~\cite{nema2017diversity,baumel2018query}. \citet{nema2017diversity} proposed an encode-attend-decode system with an additional query attention mechanism and diversity-based attention mechanism to generate a more query-relevant summary. \citet{baumel2018query} incorporated query relevance into a pre-trained abstractive summarizer to make the model aware of the query. While~\citet{xu2020abstractive} discovered a new type of connection between generic summaries and QFS queries, and provided a universal representation for them which allows \textit{generic} summarization data to be further exploited for QFS. ~\citet{su2020caire}, meanwhile, built a query model for paragraph selection based on the answer relevance score and iteratively summarized paragraphs to a budget. However, they only used QA as distant supervision to retrieve relevant segments for generating the summary, but did not take into consideration the answer relevance in the generation model. 

\subsection{Few-shot LM Prompting} 

Language model prompting method has become a new paradigm of utilizing LM for many NLP tasks. Traditional \textit{pretrain then fine-tune} way of leveraging the LM, aims to train a model $P(y\mid x;\theta)$, where a large amount of supervised data is required. While LM prompting can help circumvent this issue by instead learning an LM that models the probability $P(x;\theta)$ of text $x$ itself and using this probability to predict $y$, reducing or obviating the need for large supervised datasets.

\citet{radford2019language, brown2020language} prompt GPT-2~\cite{radford2019language} and GPT-3~\cite{brown2020language} conditioned on several few-shot examples to predict the answer for ODQA. They constructed the prompt $Prompt(Q)$ for a given question $Q$ as:
\begin{align*}
    Prompt(Q) =& \texttt{Q:} q_m \backslash n \texttt{A:} a_m \backslash n \ldots \\ &\texttt{Q:} q_1 \backslash n \texttt{A:} a_1 \backslash n \texttt{Q:} Q \backslash n
\end{align*}

where the $\{(q_1,a_1),...(q_m,a_m)\}$ are examples selected from corresponding training datsets for question $Q$. Then pass $Prompt(Q)$ through a pretrained LM such as GPT-3, to generate the answer for question $Q$ as follows: 

\begin{equation*}
    a = \mathcal{LM}(Prompt(Q))
\end{equation*}

Most recent work by ~\citet{lazaridou2022internet} further empower LM's few-shot prompting abilities with information returned from the web using Google-Search API, and experimented on QA task. Their prompt format is:
\begin{align*}
    Evidence: ...
    Question: ...
    Answer: ...
\end{align*}

While ~\citet{wei2022chain, wang2022self} use \textit{chain of thought} few-shot prompting of LM to generate a coherent chain of short sentences that minic the reasoning process of human might employ to solve reasoning tasks.

% \chapter{A Long-form Question Answering System with Fluency}
\chapter{Generating Query-relevant, Long-form Answers}

In this chapter, we first propose a coarse-to-fine method to extract the document-level and sentence-level query-relevant information, to help a traditional Seq2Seq model to  handle long and multiple documents as input, and considering query-relevance. We further introduce QFS-BART, a model that incorporates the explicit answer relevance attention of the source documents into the generation model's encoder-decoder attention module, to further enhance the query-relevance.

% how we can build an effective long-form question answering system. Specifically, we present CAiRE-COVID, a real-time long-form question answering system for COVID-19. 

% Then, we introduce QFS-BART, a model that incorporates the explicit answer relevance of the source documents given a query into the generation model, to generate more query-relevant answers.

% The CORD-19 dataset represents the most extensive machine-readable coronavirus literature collection available for data mining to date. 

\begin{figure*}[t]
    \centering
    \includegraphics[trim=0cm 1.1cm 0cm 0cm, clip=true, scale = 0.32]{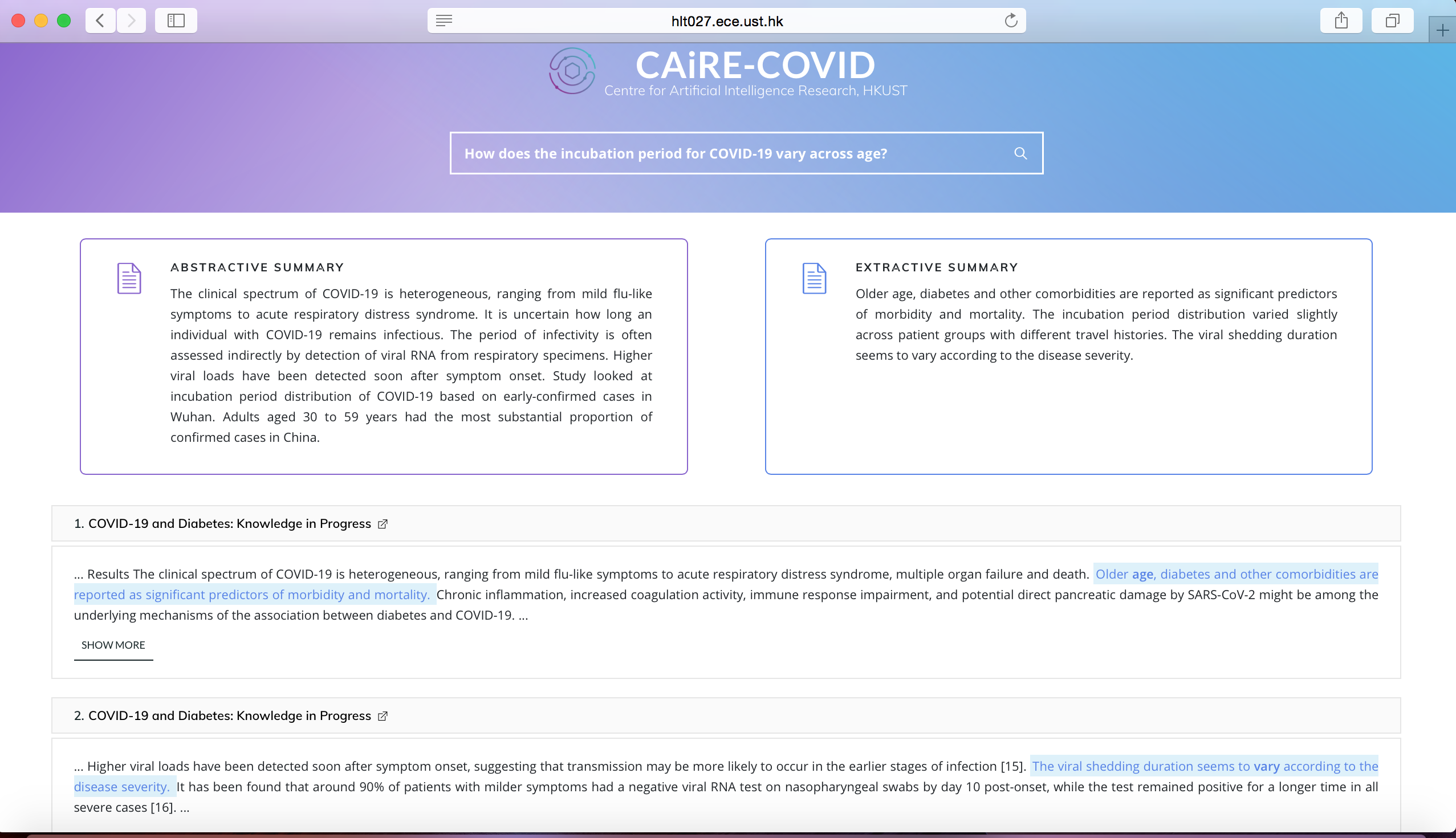}
    \caption{The user interface of our CAiRE-COVID website.}
    \label{fig:website}
\end{figure*}

\section{CAiRE-COVID: A Long-form Question Answering System}
\subsection{Backgroud}
Since the COVID-19 outbreak, there are emerging requests from both the medical research community and wider society for efficient management of the information about COVID-19. Many high priority questions need to be answered, e.g., \textit{What do we know about COVID-19 risk factors?} and \textit{What do we know about virus genetics, origin, and evolution?}. At the same time, a huge number of scientific articles have been published and made publicly available to the medical community everyday (such as \href{https://www.biorxiv.org/}{bioRxiv}, \href{https://www.medrxiv.org/}{medRxiv}, \href{https://www.who.int/}{WHO}, \href{https://www.ncbi.nlm.nih.gov/pmc/}{pubMed}). 

% While the current search engines can only return some unreliable web pages whose content can not be verified, an effective COVID-19 information management system, especially built over the scholarly documents, is in high demand.  

The release of the COVID-19 Open Research Dataset (CORD-19)\textsuperscript{\ref{foot:kaggle}} ~\cite{wang2020cord}, which consists of over 158,000 scholarly articles about COVID-19 and related coronaviruses, creates an opportunity for the natural language processing (NLP) community to address these requests. However, it also poses a new challenge since it is not easy to extract precise information regarding given scientific questions and topics from such a large set of unlabeled resources.

Almost all the systems built by the community fall in the search engine paradigm. CORD-19 Search\footnote{\tt \small https://cord19.aws/} is a search engine that utilizes the CORD-19 dataset processed using Amazon Comprehend Medical. Google released the \href{https://ai.googleblog.com/2020/05/an-nlu-powered-tool-to-explore-covid-19.html}{COVID19 Research Explorer} a semantic search interface on top of the CORD-19 dataset. Meanwhile, Covidex\footnote{\tt \small https://covidex.ai/} applies multi-stage search architectures, which can extract different features from data. An NLP medical relationship engine named the WellAI COVID-19 Research Tool\footnote{\tt \small https://wellai.health/} is able to create a structured list of medical concepts with ranked probabilities related to COVID-19, and the tmCOVID\footnote{\tt \small http://tmcovid.com/} is a bioconcept extraction and summarization tool for COVID-19 literature.

Therefore, we propose CAiRE-COVID\footnote{\tt \small https://caire.ust.hk/covid}, a real-time long form question answering system for COVID-19, to tackle the timely challenges of mining the numerous scientific articles being published on COVID-19 by \textit{answering} high priority questions from the community and \textit{summarizing} salient question-related information. The system won one of the 10 tasks in the Kaggle COVID-19 Open Research Dataset Challenge\footnote{\label{foot:kaggle}\tt \small  https://www.kaggle.com/allen-institute-for-ai/CORD-19-research-challenge}, judged by medical experts. 

\subsection{CAiRE-COVID Framework}

\begin{figure*}[t!]
    \centering
    \includegraphics[width=0.7\linewidth]{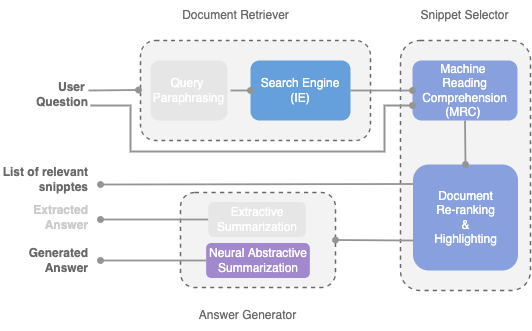}
    \caption{System architecture overview}
    \label{fig:system}
\end{figure*}

Figure~\ref{fig:system} illustrates the architecture of the CAiRE-COVID system, which consists of three major modules: 1) Document Retriever, 2) Relevant Snippet Selector, and 3) Query-focused Multi-Document Summarizer.

Given a user query, the system first \textit{selects} the most relevant documents from the CORD-19 dataset\textsuperscript{\ref{foot:kaggle}} with high coverage via a Document Retriever module. It then \textit{highlights} the answers or evidence (text spans) for the query, given the relevant paragraphs, by a Snippet Selector module via question answering (QA) models. Furthermore, to efficiently \textit{present} COVID-19 question-related information to the user, we propose a query-focused Multi-Document Summarizer to generate abstractive and extractive answers related to the question, from multiple retrieved answer-related paragraph snippets. We leverage the power of the generalization ability of pre-trained language models \cite{lewis2019bart, yang2019xlnet, lee2020biobert, su2019generalizing} by fine-tuning them for QA and summarization, and propose our own adaptation methods for the COVID-19 task.

\subsubsection{Document Retrieval}
To \textit{select} the most relevant document, i.e. article or paragraph, given a user query, we first apply the Document Retriever with the following two sub-modules. 

% \subsubsection{Search Engine}
We use Anserini~\cite{yang2018anserini} to create the search engine for retrieving a preliminary candidate set of documents. Anserini is an information retrieval module wrapped around the open source search engine Lucene\footnote{\tt \small https://lucene.apache.org/} which is widely used to build industry standard search engine applications. Anserini uses the Lucene indexing to create an easy-to-understand information retrieval module. Standard ranking algorithms (e.g, bag of words and BM25) have been implemented in the module. We use paragraph indexing for our purpose, where each paragraph of the full text of each article in the CORD-19 dataset is separately indexed, together with the title and abstract. For each query, the module can return $n$ top paragraphs matching the query.

\subsubsection{Relevant Snippet Selector}
The Relevant Snippet Selector outputs a list of the most relevant answer snippets from the retrieved documents while highlighting the relevant keywords. To effectively find the snippets of the paragraphs  relevant to a query, we build a neural QA system as an evidence selector given the queries. It aims at predicting answers or evidences (text spans) given relevant paragraphs and queries. The paragraphs are further re-ranked based on a well-designed score, and the answers are highlighted in the paragraphs.

\textbf{Extractive-based QA as Evidence Selector}

\begin{itemize}
    \item \textbf{Evidence Selection} To enhance both generalization and domain-expertise capability, we leverage an ensemble of two QA models: the HLTC-MRQA model~\cite{su2019generalizing} and the BioBERT~\cite{lee2020biobert} model. The HLTC-MRQA model is an XLNet-based~\cite{yang2019xlnet} QA model which is trained on six different QA datasets via multi-task learning. This helps reduce over-fitting to the training data and enable generalization to out-of-domain data and achieve promising results. To adopt the HLTC-MRQA model as evidence selector into our system, instead of fine-tuning the QA model on COVID-19-related datasets, we focus more on maintaining the generalization ability of our system and conducting zero-shot QA.

    To obtain a better performance, we also combine the HLTC-MRQA model with a \textit{domain-expert}: the BioBERT QA model, which is fine-tuned on the SQuAD dataset. 
    \item \textbf{Answer Fusion} To increase the readability of the answers, instead of only providing small spans of answers, we provide the sentences that contain the predicted answers as the outputs. When the two QA models find different evidence from the same paragraph, both pieces of evidence are kept. When the predictions from the two models are identical or there is an inclusion relationship between the two, the predictions will be merged together.
    \item \textbf{Answer Re-ranking and Highlight Generation}
    \label{sec:rerank}
    The retrieved paragraphs are further re-ranked based on the answer relevance to the query.
    
    \textbf{Answer Confidence Score} We leverage the prediction probability from the QA models as the answer's confidence score. The confidence score of an ensemble of two QA models is computed as in Equation \ref{eq:qa_score}.
\begin{equation}
\label{eq:qa_score}
\setlength\abovedisplayskip{3pt}%shrink space
\setlength\belowdisplayskip{3pt}
s_{conf}\!=\!\left\{
    \begin{array}{ll}
    \begin{aligned}
    0.5min\{|s_{m}|,|s_{b}|\}\\ -\!max\{|s_{m}|,|s_{b}|\} \end{aligned}
    & 
    if s_{m},s_{b}\!<\!0 \\
    s_{m}+s_{b}     & {otherwise,}\\
    \end{array} 
\right. 
\end{equation}
where the confidence score from each model is annotated as $s_{m}$ and $s_{b}$.

    \textbf{Keyword-based Score} We calculate the matching score between a query and the retrieved paragraphs based on word matching. To obtain this score, we first select important keywords from the query based on POS-tagging, only taking words with {NN (noun), VB (verb), JJ (adjective)} tags into consideration. By separately summing the term frequencies and the total number of important keywords that appear in the paragraph, we can get two matching scores, which are annotated as $s_{freq}$ and $s_{num}$, respectively. For the term-frequency matching score, we normalize shorter paragraphs using a sigmoid value computed from the paragraph length, and reward paragraphs with more diverse keywords from the query. The final matching score is computed as in Equation \ref{eq:s_match}.
\begin{equation} \label{eq:s_match}
\setlength\abovedisplayskip{3pt}%shrink space
\setlength\belowdisplayskip{3pt}
s_{match}\!=\!\lambda_1 s_{freq}\!\cdot\! \sigma(l-l_c)\!+\!\lambda_2s_{num},
\end{equation}
where $l$ is the length of the paragraph and $l_c$ is a length constraint. Because of the effect of the sigmoid function, for data samples whose paragraph length is shorter or similar to $l_c$, the penalty will be applied to the final matching score.

    \item \textbf{Re-rank and Highlight} The re-ranking score is calculated based on both the matching score and the confidence score, as shown in Equation \ref{eq:rerank}. The relevant snippets are then re-ranked together with the corresponding paragraphs and displayed via highlighting:
\begin{equation}
\label{eq:rerank}
score_{re-rank} = s_{match}+\alpha s_{conf}.
\end{equation}

\end{itemize}

\subsubsection{Answer Generation}
To efficiently present pertinent COVID-19 information to the user, we propose to generate abstractive and extractive answers related to COVID-19 questions.  

\textbf{Abstractive Answer Generation}
\begin{itemize}
    \item \textbf{BART Fine-tuning} Our abstractive answer generatino model is based on BART~\cite{lewis2019bart}, which obtained state-of-the-art results on the summarization tasks on the CNN/DailyMail datasets~\cite{hermann2015teaching} and XSUM~\cite{narayan2018don}. We use the BART model fine-tuned on the CNN/DailyMail dataset as the base model since we do not have other COVID-19 related data. 
    \item \textbf{Incorporating Answer Relevance}
In order to generate query-focused answers, we propose to incorporate answer relevance in the BART-based answer generation process in two aspects. First, instead of using the paragraphs list passed by the Document Retriever, we use the top $k$ paragraphs  $ \{para_1, para_2, .., para_k\}$ passed by the QA module as input to the Multi-document Summarizer, which are re-ranked according to their \textit{answer relevance} to the query, as shown in Equation \ref{eq:rerank}. Then, instead of using only the re-ranked answer-related paragraphs to generate an answer, we further incorporate \textit{answer relevance} by concatenating the predicted answer spans from the QA models with each corresponding paragraph. We also concatenate the query to the end of the input, since this has been proved to be effective for the QFS task \cite{savery2020question}. So input to the answer generation model is
$C = \{\hat{para}_1, \hat{para}_2, .., \hat{para}_k\} $, where 
\begin{equation}
\setlength\abovedisplayskip{3pt}%shrink space
\setlength\belowdisplayskip{3pt}
 \hat{para}_{i} = [para_{i}; ans\_spans_{i}; query]   
\end{equation}

\item \textbf{Multi-document Answer Generation} Considering that each $\hat{para}_i$ in $C$ may come from different articles and focus on different aspects of the query, we generate the multi-document summary by directly concatenating the summary of each $\hat{para}$, to form our final answer. Some redundancy might be included, but we think this is fine at the current stage.

\end{itemize}

\textbf{Extractive Answer Generation}

In order to generate a query-focused extractive answer, we first extract answer sentences which contain the answer spans generated from the QA module, from multiple paragraphs as candidates. Then we re-rank and choose the top-$k$ (k=3) according to their answer relevance score to form our final answer\footnote{The number of documents $k$ is constrained by the maximum input length of LMs, and is selected based on our empirical experience.}. The \textit{answer relevance} score is calculated in the following way:
\begin{itemize}
    \item \textbf{Sentence-level Representation}
To generate a sentence-level representation we sum the contextualized embeddings encoded by ALBERT\cite{lan2019albert}, then divide by the sentence length. This representation can capture the semantic meaning of the sentence to a certain degree through a stack of self-attention layers and feed-forward networks. For a sentence with \(n\) tokens $X = [{w}_1, {w}_2, .., {w}_n] $, the representation \(h\) is calculated by Equation \ref{eq:sen_emb}.
\begin{align}\label{eq:sen_emb}
\setlength\abovedisplayskip{3pt}%shrink space
\setlength\belowdisplayskip{3pt}
\begin{split}
    &e_{1:n} = ALBERT([{w}_1, {w}_2, .., {w}_n]) \\
    &h =\frac{\sum_{i=1}^{n}e_i}{n}
\end{split}
\end{align}

\item \textbf{Similarity Calculation}
After sentence representation extraction, we have embeddings for the answer sentences and the query. In this work, the cosine similarity function is used for calculating the similarity score between them. For each query, only the top 3 answer sentences are kept.

\end{itemize}

\begin{table*}[t!]
\centering
\resizebox{0.68\textwidth}{!}{
\begin{tabular}{lcccccc}
\toprule
\multirow{2}{*}{Model} & \multicolumn{3}{c}{NL Question} & \multicolumn{3}{c}{Keyword Question} \\ \cmidrule{2-7} 
& P@1 & R@3 & MRR & P@1 & R@3 & MRR \\
\midrule
T5($+$ MS MARCO)$^{\dagger}$ & 0.282 & 0.404 & 0.415 & 0.210 & 0.376 & 0.360 \\
\midrule
BioBERT & 0.177 & 0.423 & 0.288 & 0.162 & 0.354 & 0.311 \\
HLTC-MRQA &  0.169 & 0.415 & 0.291 & 0.185 & 0.431 & 0.274 \\
Ensemble &  0.192 & \textbf{0.477} & 0.318 & \textbf{0.215} & \textbf{0.446} & 0.329 \\
\bottomrule
\end{tabular}
}
\caption{Results of the QA models on the CovidQA dataset. $^{\dagger}$The T5 model~\cite{2019t5} which is fine-tuned on the MS MARCO dataset~\cite{DBLP:conf/nips/NguyenRSGTMD16} is the strongest baseline from \citet{tang2020rapidly}. However, due to the difference in experiment settings, the MRR values from our models and those from baseline models are not comparable.}
\label{tab:qa}
\end{table*}

\subsection{Experiments}
In order to quantitatively evaluate the performance of each module and show the effectiveness of our system, we conduct a series of respective experiments.

In Table ~\ref{tab:caire_covid_examples_1}, we show concrete examples of our system. 

\subsubsection{Experiments: Machine Reading Comprehension}
% \subsubsection{Quantity Evaluation}

\textbf{Dataset} We evaluate our QA module performance on the CovidQA dataset, which was recently released by \citet{tang2020rapidly} to bootstrap research related to COVID-19. The CovidQA dataset consists of 124 question-article pairs related to COVID-19 for zero-shot evaluation on transfer ability of the QA model. 

\textbf{Experiment Settings} The evaluation process on the CovidQA dataset is designed as a text ranking and QA task. 
% Given one question and the corresponding articles, the QA models are supposed to rank all the sentences from the article, where the higher the rank, the more likely the sentence is to contain the golden answer. However, our QA module performs in paragraph level in our system, which makes the experiment settings a bit different. 
For one article which contains $M$ sentences, we split it into $N (N<M)$ paragraphs. One sentence is selected as the evidence to the query from each of the paragraphs. The re-ranking scores for each sentences are meanwhile calculated. After evidence selection, we re-rank the $N$ sentences according to the re-ranking score (\S \ref{sec:rerank}). The QA results are evaluated with Mean Reciprocal Rank (MRR), precision at rank one (P@1) and recall at rank three (R@3). However, in our case, MRR is computed by:
\begin{equation}
\setlength\abovedisplayskip{3pt}%shrink space
\setlength\belowdisplayskip{3pt}
MRR=\frac{1}{|Q|}\sum_{i=1}^{|Q|}\{\frac{1}{rank_i},0\},
\end{equation}
where $rank_i$ is the rank position of the first sentence where the golden answer is located given one article (We assume it as the golden sentence). If there's no golden sentence selected in the $N$ candidates, we assign the score of the data sample as zero. 
% However, in the original settings, even though the golden sentence is not ranked in the first $N$ places, the score of the data sample will be $0<\frac{1}{M}\leq\frac{1}{rank_i}<\frac{1}{N}$.
% As a result, due to the different experiment settings, our MRR value tends to be lower than that with the original settings, while precision and recall fractions are not affected.
For the QA module, we conduct all the experiments with hyper-parameter $\lambda_1$ as 0.2, $\lambda_2$ as 10, $l_c$ as 50 and $\alpha$ as 0.5.

\textbf{Analysis} The results are shown in Table \ref{tab:qa}. We test our models on both natural language questions and keyword questions. Changes in the efficiency of different models indicate their preferences for different kinds of questions.
The HLTC-MRQA model with keyword questions shows better performance on precision and recall fractions, while the model with natural language questions is more likely to have relevant answers with a higher rank. The BioBERT model, however, performs under a different scheme. After making an ensemble of two QA models, the performance in terms of  precision, recall and MRR fractions is improved. Moreover, our QA module even outperforms the T5~\cite{2019t5} baseline on the recall metric, while for keyword questions, our model also marginally outperforms T5 on the precision fraction.

\begin{figure*}[t!]
    \centering
    \includegraphics[width=0.9\linewidth]{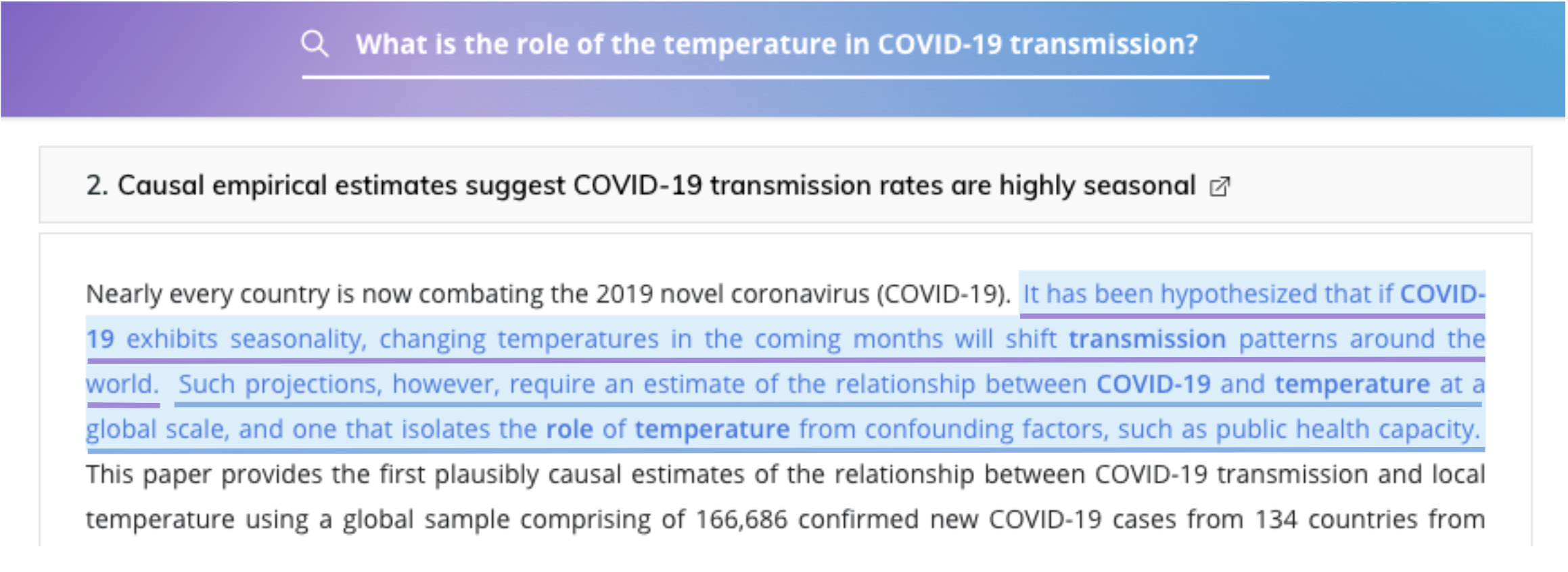}
    \caption{An example of QA output of our system. The output of the QA module is highlighted in the paragraph in blue. We also use purple and blue underlining to distinguish the outputs of the HLTC-MRQA model and the BioBERT model.}
    \label{fig:qa-case}
\end{figure*}

\textbf{Case Study}
Despite the fact that two models select the same sentence as the final answer given a question in most of the times when there is a reasonable answer in the paragraph, we observe that two models show different \textit{taste} on language style. Figure~\ref{fig:qa-case} shows a representative example of QA module output. The prediction of the BioBERT model shows its preference for an experimental style of expression, while the prediction of the MRQA model is more neutral to language style.

\subsubsection{Experiments: Query-relevant, long-form Answer Generation}

% In order to generate query-focused answers for COVID-19 questions, we propose to incorporate answer relevance with the help of a QA model into the generation process. 
To investigate the quality of the answer generator module, we leverage the query-focused summerization (QFS) task. Query focused summarization (QFS) aim to extract essential information from a source document(s) and organize it into a summary that can answer a query~\cite{dang2005overview}. 

\textbf{Datasets}
Since there are no existing QFS datasets for COVID-19, we choose the following two datasets to evaluate the performance of the answer generator. 
\begin{itemize}
    \item \textbf{DUC Datasets}
DUC 2005~\cite{dang2005overview} first introduced the QFS task. This dataset provides 50 queries paired with multiple related document collections. Each pair, has 4-9 human written summaries. The excepted output is a summary within 250 words for each document collection that can answer the query. DUC 2006~\cite{hoa2006overview} and DUC 2007 have a similar structure. We split the documents into paragraphs within 400 words to fit the QA model input requirement.

    \item \textbf{Debatepedia Dataset} This dataset is included in our experiments since it is very different from the DUC QFS datasets. Created by \cite{nema2017diversity}, it is the first large-scale QFS dataset, consisting  of 10,859 training examples, 1,357 testing and 1,357 validation samples. The data come from Debatepedia, an encyclopedia of pro and con arguments and quotes on critical debate topics, and the summaries are debate key points that are a single short sentence. The average number of words in summary, documents and query is 11.16, 66.4, and 10 respectively.
\end{itemize}

\textbf{Model Setting}
\begin{itemize}
    \item \textbf{Abstractive Answer Generation Model Setting}
    
We use BART ~\cite{lewis2019bart} fine-tuned on XSUM~\cite{narayan2018don} as the abstractive base model for the Debatepedia dataset, since XSUM is the most abstractive dataset containing the highest number of novel bi-grams. Meanwhile, we use BART fine-tuned on CNN/DM for the DUC dataset to generate longer summaries. Different input combination settings are tested. 
\begin{itemize}
    \item \textit{BART(C)}: We use the context only as the input to the BART model.

    \item \textit{BART(C,Q)}: We use the concatenation of the context and query as input to the BART model.
   \item \textit{BART(Q,C)}: We concatenate the query at the beginning of the context as the input to the BART model.

    \item \textit{BART(A,Q)}: We concatenate the answer sentences (sentences from the context that contain the answer spans) with the query as input to the BART model.

    \item \textit{BART(Q,A)}: We switch the position of query and answer sentences as input to the BART model.

% \textit{BART(C,A,Q)}: We concatenate the context, answer spans, and query as input, which is the configuration we adopted in our system.
    \item \textit{BART(C$_\textit{nr}$)}: We use the context only as the input to the BART model. However, we do not re-rank the paragraphs in the context.

    \item \textbf{\textit{BART(C,A,Q)}}: We concatenate the context, answer spans, and query as input, which is the input configuration we adopt in our system. 

\end{itemize}

For the DUC datasets, which contain multiple documents as context, we iteratively summarize the paragraphs which are re-ranked by the QA confidence scores till the budget of 250 words is achieved. 

\item \textbf{Extractive Model Setting}
We conduct extractive summarization on the DUC datasets. LEAD is our baseline \cite{xu2020query}. For each document collection, LEAD returns all leading sentences of the most recent document up to 250 words. Our answer relavance driven extractive method has been introduced. 

\end{itemize}

\begin{table*}[!h]
\centering
\resizebox{0.95\textwidth}{!}{%
\begin{tabular}{l|ccc|ccc|ccc}
\hline
\multicolumn{1}{c|}{\multirow{2}{*}{\textbf{Model Setting}}} & \multicolumn{3}{c|}{\textbf{ROUGE-1}} & \multicolumn{3}{c|}{\textbf{ROUGE-2}} & \multicolumn{3}{c}{\textbf{ROUGE-L}} \\ \cline{2-10} 
\multicolumn{1}{c|}{} & \textbf{Recall} & \textbf{Precision} & \textbf{F1} & \textbf{Recall} & \textbf{Precision} & \textbf{F1} & \textbf{Recall} & \textbf{Precision} & \textbf{F1} \\ \hline
BART(C) & 19.60 & 8.80 & 11.80 & 3.22 & 1.41 & 1.91 & 16.76 & 8.17 & 10.70 \\ \hline
BART(C,Q) & 20.43 & 9.27 & 12.36 & 3.56 & 1.60 & 2.13 & 17.50 & 8.58 & 11.19 \\ \hline
BART(Q,C) & 19.16 & 8.49 & 11.43 & 3.06 & 1.31 & 1.76 & 16.39 & 7.77 & 10.25 \\ \hline
BART(A,Q) & 20.15 & 8.93 & 12.04 & 3.37 & 1.43 & 1.95 & 17.29 & 8.25 & 10.88 \\ \hline
BART(Q,A) & 19.15 & 8.57 & 11.48 & 2.97 & 1.27 & 1.70 & 16.46 & 7.88 & 10.36 \\ \hline
\multicolumn{1}{c|}{\textbf{BART(C,A,Q)}} & \textbf{21.92} & \textbf{10.05} & \textbf{13.32} & \textbf{4.21} & \textbf{1.85} & \textbf{2.47} & \textbf{19.09} & \textbf{9.36} & \textbf{12.18} \\ \hline
\end{tabular}%
}
\caption{Results for Debatepedia QFS dataset}
\label{results-debate}
\end{table*}

\begin{table*}[!h]
\centering
\resizebox{0.95\textwidth}{!}
{
\begin{tabular}{l|ccc|ccc|ccc}
\hline
\multicolumn{1}{l|}{\multirow{2}{*}{\textbf{Model Setting}}} & \multicolumn{3}{c|}{\textbf{DUC 2005}} & \multicolumn{3}{c|}{\textbf{DUC 2006}} & \multicolumn{3}{c}{\textbf{DUC 2007}} \\ \cline{2-10} 
\multicolumn{1}{c|}{} & \textbf{1} & \textbf{2} & \textbf{SU4} & \textbf{1} & \textbf{2} & \textbf{SU4} & \textbf{1} & \textbf{2} & \textbf{SU4} \\ \hline
LEAD                 & 33.35 & 5.66 & 10.88 & 32.10  & 5.30  & 10.40   & 33.40  & 6.50  & 11.30 \\ \hline
Our Extractive Method & \textbf{35.19} & \textbf{6.28} & \textbf{11.61} & \textbf{34.46} & \textbf{6.51} & \textbf{11.23}  & \textbf{35.31} & \textbf{7.79} & \textbf{12.07} \\ \hline
\hline
BART($C_{\textit{nr}}$) & 32.41 & 4.62 & 9.86 & 35.78  & 6.25 & 11.37   & 37.87 & 8.11 & 12.96 \\ \hline
BART(C) & 34.25 & 5.60 & 10.88 & 37.99  & 7.64 & 12.81   & 40.66 & 9.33 & 14.43 \\ \hline
BART(C,Q) & 34.20 & \textbf{5.77} & 10.88 & 38.26  & 7.75 & \textbf{12.95}   & \textbf{40.74} & \textbf{9.60} & \textbf{14.63} \\ \hline
BART(C,A) & 34.29 & 5.70 & 10.93 & 38.31  & 7.60 & 12.90   & 40.71 & 9.11 & 14.30 \\ \hline
\textbf{BART(C,A,Q)} & \textbf{34.64} & 5.72 & 11.04 & \textbf{38.31} & \textbf{7.70} & 12.88 & 40.53 & 9.24 & 14.37 \\ \hline
\end{tabular}
}
\caption{Results for DUC datasets}
\label{results-DUC}
\end{table*}

\textbf{Results}

We use ROUGE as the evaluation metric for the performance comparison. 
Table \ref{results-debate} and Table \ref{results-DUC} show the results for the Debatepedia QFS dataset and DUC datasets respectively. As we can see from the two tables, by incorporating the answer relevance, consistent ROUGE score improvements of \textbf{BART(C,A,Q)} over all other settings are achieved on both datasets, which proves the effectiveness of our method. Furthermore, as shown in Table \ref{results-DUC}, consistent ROUGE score improvements are obtained by our extractive method over the LEAD baseline, and in the abstractive senario, BART(C) also outperforms BART(C$_\textit{nr}$) by a good margin, showing that re-ranking the paragraphs via their answer relevance can help improve multi-document QFS performance.

\begin{table*}[ht]
\centering
\begin{adjustbox}{width={0.90\textwidth},totalheight={\textheight},keepaspectratio}
\begin{tabular}{p{1\columnwidth}}
% {\textwidth}
\hline
\textbf{Question}: What are the risk factors for COVID-19? (\textit{from Task-2}) \\
\textbf{Abstractive Answer}: Reliably identifying patients at increased risk for COVID-19 complications could guide clinical decisions, public health policies, and preparedness efforts. The prevalence of diabetes patients hospitalized in intensive care units for COVID-19 is two- to threefold higher. An increased body mass index is a major risk factor for requiring respiratory assistance. The Center for Disease Control and Prevention (CDC) suggests that neurological comorbidities, including epilepsy, may be a risk factor for COVID-19. Presently, a medical history of epilepsy has not been reported. \\
\textbf{Extractive Answer}: The Center for Disease Control and Prevention (CDC) suggests that neurological comorbidities, including epilepsy, may be a risk factor for COVID-19, despite the lack of evidence. As such, it is unclear to what extent the prevalence of comorbidities in the studied population differs from that of same age (and sex) SARS-CoV-2 positive patients; and, accordingly, whether these comorbidities are significant risk factors for severe COVID-19 or merely a reflection of comorbidity prevalence in the wider population. What are the factors, genetic or otherwise, that influence interindividual variability in susceptibility to COVID-19, its severity, or clinical outcomes? \\
\hline
\textbf{Query}: What has been published about information sharing and inter-sectoral collaboration? (\textit{from Task-10}) \\
\textbf{Abstractive Answer}: Epidemiology and laboratory collaboration between the human health and animal health sectors is a fundamental requirement and basis for an effective One Health response. During the past decade, there has been significant investment in laboratory equipment and training. For example, a key determining factor relating to cross-border collaboration is whether or not the neighbour in question is a fellow member of the EU. Several system leaders called for further investment in knowledge sharing among a broad network of health system leaders.
\\
\textbf{Extractive Answer}: Criteria selected in order of importance were: 1)severity of disease in humans, 2)proportion of human disease attributed to animal exposure, 3)burden of animal disease, 4)availability of interventions, and 5)existing inter-sectoral collaboration. Various rules-in-use by actors for micro-processes (e.g. coordination, information sharing, and negotiation) within NPGH arenas establish ranks and relationships of power between different policy sectors interacting on behalf of the state in global health. For example, our findings suggest that a key determining factor relating to cross-border collaboration is whether or not the neighbour in question is a fellow member of the EU.\\
\hline
\end{tabular}
\end{adjustbox}
\caption{Example QA pairs and the abstractive and extractive answers given CORD-19 task questions from our system.}
\label{tab:caire_covid_examples_1}
\end{table*}

% \subsection{Summary}
% In this Chapter, we first propose a real-time LFQA system, CAiRE-COVID, with open-domain QA and query focused multi-document summarization techniques for efficiently mining scientific literature given a query. The system has shown its efficiency on the Kaggle CORD-19 Challenge, which was evaluated by medical researchers, and a series of experimental results also proved the effectiveness of our proposed methods and the competency of each module. The system is also easy to be generalized to general domain-agnostic literature information mining, especially for possible future pandemics. We have launched our website\textsuperscript{\ref{foot:website}} for real-time interactions and released our code\textsuperscript{\ref{foot:git}} for broader use. 

% \newcommand\BibTeX{B\textsc{ib}\TeX}

\section{Improving Query-relevance with Answer Relevance Score}

% Query focused summarization (QFS) models aim to generate summaries from source documents that can answer the given query. Most previous work on QFS only considers the query relevance criterion when producing the summary. However, studying the effect of answer relevance in the summary generating process is also important. 

In the CAiRE-COVID system, we propose to generate the query-relevant answers by concatenating the retrieved relevant segments predicted by an MRC module to the question as input, for abstractive answer generation. While it has demonstrate its effectiveness at generating somewhat relevance answers by both quantitative and qualitative evaluations, it did not take into consideration the query-relevance of the retrieved segments in the generation model. 

In this part, we try to further improve the query-relevance of the generated answer, i.e., given the retrieved documents, we try to make the generated answer more relevant to the query. Specifically, we work on a query-driven summarization(QFS) task, which aims to generate summaries that can answer the given query from provided source documents, and try to improve the query relevance of the generated summary.

we propose QFS-BART, a model that incorporates the explicit answer relevance of the source documents given the question via a MRC model, to generate coherent and 'answer-related' summaries. We leverage a state-of-the-art QA model~\cite{su2019generalizing} to predict the answer relevance of the given source documents to the query, then further incorporate the answer relevance into the BART-based generation model. We conduct empirical experiments on the Debatepedia dataset, one of the first large-scale QFS datasets~\cite{nema2017diversity}, and achieve the new state-of-the-art performance on the ROUGE metrics compared to all previously published work.

\subsection{Query Focused Summarization (QFS)}
Query focused summarization (QFS) models aim to extract essential information from a source document(s) and organize it into a summary that can answer a query~\cite{dang2005overview}. The input can be either a \textit{single} document that has multiple views or \textit{multiple} documents that contain multiple topics, and the output summary should be focused on the given query. QFS has various applications (e.g., a personalized search engine that provides the user with an overview summary based on their query~\cite{su2020caire}).

\begin{table}[!ht]
    \centering
    \begin{adjustbox}{width={0.95\textwidth},totalheight={0.9\textheight},keepaspectratio}
\begin{tabular}{|p{1\columnwidth}|}
\hline
\textbf{Document}: Interrogator Ali Soufan said in an April op-ed article in the New York Times: ``It is inaccurate to say that Abu Zubaydah had been uncooperative [and that enhanced interrogation techniques supplies interrogators with previously unobtainable information]. Along with another f.b.i. agent and with several c.i.a. officers present I questioned him from March to June before the harsh techniques were introduced later in August. \hlorange{Under traditional interrogation methods he provided us with important actionable intelligence.}'' \\ \hline
\textbf{Query}: Are traditional interrogation methods insufficient?                                                                                         \\ \hline
\textbf{Summary}: The same info can be obtained by traditional interrogations.                                                                                   \\ \hline
\end{tabular}
    \end{adjustbox}
    \caption{An example of QFS. The input is a document and a corresponding query, and the \hlorange{highlight} sentence is the answer from our QA module. We observe that the summary and the answers are very correlated.}
    \label{tab:introduction}
\end{table}

Early work on the QFS task mainly focused on generating extractive summaries~\cite{davis2012occams, daume2006bayesian, feigenblat2017unsupervised, xu2020coarse}, which may contain unreadable sentence ordering and lack cohesiveness. Other work on abstractive QFS incorporated the query relevance into existing neural summarization models~\cite{nema2017diversity,baumel2018query}. The closest work to ours was done by~\cite{su2020caire} and~\cite{xu2020abstractive, xu2020coarse}, who leveraged an external question answering (QA) module in a pipeline framework to take into consideration the answer relevance of the generated summary. However, they only used QA as distant supervision to retrieve relevant segments for generating the summary, but did not take into consideration the answer relevance in the generation model. As shown in the Table~\ref{tab:introduction}, the query focused summary is correlated to the answer extracted from the QA module.

On the other hand, recent neural summarization models~\cite{paulus2017deep, gehrmann2018bottom, zhang2020pegasus} have achieved remarkable performance in~\textit{generic} abstractive summarization by taking advantage of large pre-trained language models~\cite{lewis2019bart,zhang2020pegasus}. Yet, how to leverage these models and adapt them to the QFS task remains unexplored.

% Generalizing question answering (QA) systems \cite{yang2019end, su2019generalizing} have shown significant achievements in open-domain QA task especially in identifying answer spans. Given a query, the models can provide an answer relevance score for each token in the context. Yet, despite their practicality, very few studies have used QA systems to enhance the QFAS models. In recent years, some work \cite{xu2020query, xu2020abstractive} has been done to apply QA models to rank the text segments in the first stage and feed the selected text segments to summarization model in the second stage. It is essential to incorporate the answer relevance in the summary generating process where less attention is paid. To address this research gap, we propose QFS-BART, a model that incorporates explicit answer relevance, to produce more answer relevant and coherent sumamries. 

% Our contributions in this work are threefold: 1) our work demonstrates the effectiveness of the answer relevance score in neural abstractive QFS; 2) we propose an effective method to incorporate the answer relevance score into the pre-trained language models which can produce more query-relevant summaries; 3) our model reaches the state-of-the-art performance on a single-document QFS dataset (Debatepedia), and brings substantial improvements over several strong baselines on two multi-document QFS datesets (DUC 2006, 2007).

\subsection{Methodology}

\begin{figure}
    \centering
    \includegraphics[scale=0.7]{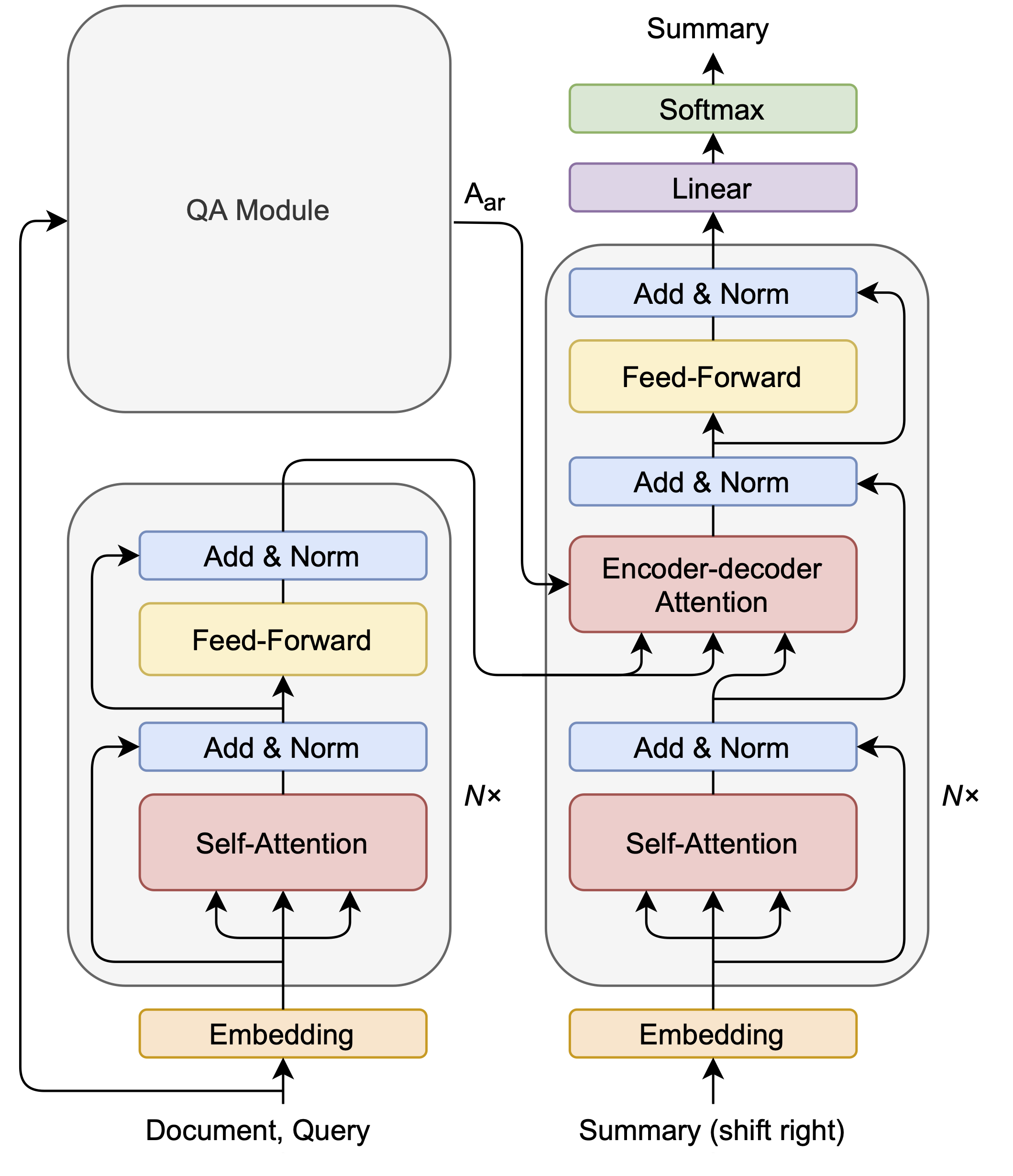}
    \caption{The framework of QFS-BART. The QA module calculates the answer relevance scores, and we incorporate the scores as explicit answer relevance attention to the encoder-decoder attention.}
    \label{fig:QFS-BART-model}
\end{figure}

In this section, we present our approach to incorporating the answer relevance into QFS. First, we introduce the method of generating answer relevance scores. Then, we describe our answer relevance attention in the Transformer-based model. Third, we introduce our QFS-BART model in which the decoder is composed of a stack of answer relevance decoding layers, as shown in Figure \ref{fig:QFS-BART-model}.

\subsubsection{Answer Relevance Generation}
In recent years, neural models~\cite{yang2019end, su2019generalizing} have shown remarkable achievements in QA tasks. In order to apply QA models to the QFS task, we use HLTC-MRQA~\cite{su2019generalizing} to generate the answer relevance score for each word in context. The reason for choosing HLTC-MRQA is twofold: 1) it shows robust generalization and transferring ability on different datasets, and 2) the model shows great performance in QA tasks and significantly outperforms the BERT-large baseline by a large margin. The HLTC-MRQA is introduced as follows.

Based on XLNet~\cite{yang2019xlnet}, HLTC-MRQA is fine-tuned on multiple QA datasets with an additional multilayer perceptron (MLP). Given a context that contains $n$ words, the model outputs a distribution $s \in (0, 1)$ for each word's probability of being the start word of the answer and a probability distribution $e \in (0, 1)$ to be the end word of answer. To generate the answer relevance score $r$ for each word, we calculate it by summing two distributions:
\begin{equation}
    r = s + e,
\end{equation}
where $r \in (0, 2)$.

\subsubsection{Answer Relevance Attention}
Scaled dot-product attention~\cite{vaswani2017attention} is the core-component of the Transformer-based model:
\begin{equation}
    Attention(Q, K, V) = softmax( \frac{QK^{T}}{ \sqrt{d}}V),
\end{equation}
where $d$ is the dimension of the query matrix $Q$, key matrix $K$ and value matrix $V$. The Transformer encoder is constructed by self-attention layers, where all of the keys, values and queries come from the input sequence. This makes each token in the input attend to all other tokens. The Transformer decoder layer is a combination of a self-attention layer and encoder-decoder attention layer. In the encoder-decoder attention layer, the query comes from the decoder's self-attention layer, and the key and value come from the output of the encoder. This allows every generated token to attend to all tokens in the input sequence.

In this work, we propose to incorporate the word-level answer relevance score as additional explicit encoder-decoder attention in the transformer decoder. Given a document with $n$ tokens, we generate a summary with a maximum length of $m$ tokens. Let $x^{l} \in \mathbb{R}^{n*d}$ denotes the output of the $l$-th transformer encoder layer and $y^{l} \in \mathbb{R}^{m*d}$ denotes the output of the $l$-th transformer decoder layer's self-attention layer. The encoder-decoder attention $\alpha^{l} \in \mathbb{R}^{m*n}$ can be computed as:
\begin{equation}
    \alpha^{l} = softmax(\frac{(y^{l}W_{Q})(x^{l}W_{K})}{\sqrt{d_{k}}} + A_{ar}),
\end{equation}
where $W_{Q}$ and $W_{K} \in \mathbb{R}^{d_{k}*d_{k}}$ are parameter weights and $A_{ar} \in \mathbb{R}^{m*n}$ is our explicit answer relevance score. Since the original answer relevance score is an $n$-dimensional vector, we repeat it $m$ times to generate an $m$ by $n$ attention matrix, which means our answer relevance attention is equal to all generated tokens.

\subsubsection{QFS-BART}

\begin{table*}[ht]
    \centering
    \begin{adjustbox}{width={0.7\textwidth},totalheight={\textheight},keepaspectratio}
    \begin{tabular}{lccc}
    \hline
    \textbf{Models}                                     & \textbf{ROUGE-1} & \textbf{ROUGE-2} & \textbf{ROUGE-L} \\ \hline \hline
    \multicolumn{4}{l}{\qquad \textit{Without Pre-training}}  \\
    % UT \cite{dehghani2018universal}                    & 36.21 & 25.75 & 35.53 \\
    % CSA Transformer (Mul)* \cite{xie2020conditional}    & 41.70 & 32.92 & 41.29 \\
    Transformer~\cite{vaswani2017attention}             & 28.16 & 17.48 & 27.28 \\
    Transformer (CONCAT)                                & 41.72 & 33.62 & 41.25 \\
    Transformer (ADD)                                   & 41.10 & 33.35 & 40.72 \\
    SD2*~\cite{nema2017diversity}                      & 41.26 & 18.75 & 40.43 \\ 
    CSA Transformer*~\cite{xie2020conditional}          & 46.44 & 37.38 & 45.85 \\ 
    \hdashline
    \multicolumn{4}{l}{\qquad \textit{With Pre-training}}  \\ 
    RSA Word Count*~\cite{baumel2018query}              & 53.09 & 16.10 & 46.18 \\
    QR-BERTSUM-TL*~\cite{laskar2020query}               & 57.96 & \textbf{45.20} & 57.05 \\ \hline
    \textbf{BART-FT}                                    & 57.98 & 43.62 & 56.30 \\
    \textbf{QFS-BART}                                   & \textbf{59.02} & 44.59 & \textbf{57.44} \\ \hline
    \end{tabular}
    \end{adjustbox}
    \caption{ROUGE-F1 scores for Debatepedia QFS dataset. Results with * mark are taken from the corresponding papers. The previous work can be divided into two categories: 1) training the models from scratch, and 2) using pre-trained models and fine-tuning on a QFS dataset.}
    \label{tab:qfs_main_results}
\end{table*}

Generative pre-trained models~\cite{dong2019unified, lewis2019bart, raffel2019exploring} have shown remarkable performance in natural language generation (NLG), including text summarization. We choose to combine our answer relevance attention with BART ~\cite{lewis2019bart}, a denoising autoencoder built with a sequence-to-sequence model, for two reasons: 1) BART achieves state-of-the-art performance on several summarization datasets (i.e. CNN/DailyMail ~\cite{hermann2015teaching} and Xsum ~\cite{narayan2018don}). 2) BART follows the standard Transformer encoder-decoder architecture, and we can easily combine the answer relevance as explicit attention to the encoder-decoder attention layers. In detail, we incorporate the same answer relevance attention for all Transformer decoder layers.

Domain adaption for natural language processing tasks is widely studied ~\cite{blitzer2007biographies, daume2009frustratingly, liu2020crossner, yu2021adaptsum}. ~\citet{hua2017pilot} first studied the adaptation of neural summarization models and showed that the models were able to select salient information, even when trained on out-of-domain data. Inspired by this, we leverage a two-stage fine-tuning method for our QFS-BART. In the first stage, we directly fine-tune the original BART model with the Xsum dataset, and in the second stage, we fine-tune our QFS-BART model with QFS datasets. All the parameters in the model are initialized from the first stage. In order to make the model capture both query relevance and answer relevance, the input text is formatted in the following way:

\centerline{[CLS] document [SEP] query.}
The answer relevance attention score for the document is generated by the QA model, and we take the maximum number in the document as the attention score for all the words in the query.

\subsection{Experimental Setup}
\textbf{Datasets} We use multiple QA datasets, including SQuAD~\cite{rajpurkar2016squad}, NewsQA~\cite{trischler2016newsqa}, TriviaQA~\cite{joshi2017triviaqa}, SearchQA~\cite{dunn2017searchqa}, HotpotQA~\cite{yang2018hotpotqa} and NaturalQuestions~\cite{kwiatkowski2019natural} to train HLTC-MRQA, following~\citet{su2019generalizing}. We evaluate our model on the Debatepedia dataset~\cite{nema2017diversity} and DUC2005-7 dataset (in Appendix).

\textbf{Training Details}
For all the experiments, we use the BART-large version to implement our models. We use a mini-batch size of 32 and train all the models on one V100 16G. During decoding, we use beam search with the beam size of 4. We decode until an end-of-sequence token is emitted and early stop when the generated summary reaches to 48 tokens.

\begin{table}[!ht]
    \centering
    \begin{adjustbox}{width={0.88\textwidth},totalheight={\textheight},keepaspectratio}
\begin{tabular}{p{1\columnwidth}}
\hline
\textbf{Document}: Interrogator Ali Soufan said in an April op-ed article in the New York Times: ``It is inaccurate to say that Abu Zubaydah had been uncooperative [and that enhanced interrogation techniques supplies interrogators with previously unobtainable information]. Along with another f.b.i. agent and with several c.i.a. officers present I questioned him from March to June before the harsh techniques were introduced later in August. Under traditional interrogation methods he provided us with important actionable intelligence.'' \\
\textbf{Query}: Are traditional interrogation methods insufficient?                                                                                             \\ \hline

\textbf{BART-FT}: Al Qaeda detainee Abu Zubaydah has been cooperative under traditional interrogation.                                                          \\ \hline

\textbf{QFS-BART}: \hlorange {The same info can be obtained by traditional interrogation.}                                                                                  \\ \hline

\textbf{Gold}: \hlgreen{The same info can be obtained by traditional interrogations.}                                                                                    \\ \hline
\end{tabular}
 \end{adjustbox}
    \caption{A example taken from Debatepedia test set. The generated summary from \textbf{QFS-BART} is almost the same as the gold summary.}
    \label{tab:case-study}
\end{table}

\subsection{Results \& Analysis}

% \subsection{Experimental Results}
We compare our proposed QFS-BART model with the following models: 1) \textbf{Transformer} does not consider the queries in the Debatepedia dataset. 2) \textbf{Transformer (CONCAT)} concatenates the query and the document. 3) \textbf{Transformer (ADD)} adds the query encoded vector to the document encoder. 4) \textbf{SD2} adds a query attention model and a new diversity-based attention model to the encode-attend-decode paradigm. 5) \textbf{CSA Transformer} combines conditional self-attention (CSA) with Transformer. 6) \textbf{RAS Word Count} incorporates query relevance into a pre-trained abstractive summarization model. 7) \textbf{QR-BERTSUM-TL} presents a transfer learning technique with the Transformer-based BERTSUM  model~\cite{liu2019text}. 8) \textbf{BART-FT} concatenates the document and query, and directly fine-tunes on the Debatepedia dataset.

We adopt ROUGE score~\cite{lin2004rouge} as the evaluation metric. As shown in Table \ref{tab:qfs_main_results}, QFS-BART significantly outperforms the models without pre-training. Compared with the models utilizing pre-training, ours improves the ROUGE-1 and ROUGE-L scores by a large margin.

\subsection{Case Study}
We present a case study comparing between the strong baseline BART-FT model, our QFS-BART model and the gold summary, shown in Table \ref{tab:case-study}. It's clear that the baseline model tends to copy spans from the document which are not directly related to the query and the QFS-BART model produces a more query- and answer- related summary.

\subsection{Adapting QFS-BART to Multi-document QFS Task}
DUC 2005-7 are datasets for the multi-document query focused summarization (QFS) task. As shown in the Table \ref{tab:length_of_dataset}, the documents and summaries of the DUC datasets are extremely longer than those in the Debatepedia \cite{nema2017diversity} dataset. We thus need to adapt the QFS-BART model to handle the multi-document scenario and produce longer output.

\begin{table}[!ht]
    \centering
    \begin{adjustbox}{width={0.5\textwidth},totalheight={\textheight},keepaspectratio}
    \begin{tabular}{l|ccc}
    \hline
    \textbf{Datasets}   & \textbf{Document(s)}      & \textbf{Query}        & \textbf{Summary}  \\ \hline
    Debatepedia         & 66.40                     & 11.16                 & 9.97              \\
    DUC 2005            & 20058.12                  & 26.60                 & 243.56            \\   
    DUC 2006            & 14330.14                  & 23.30                 & 246.84            \\
    DUC 2007            & 10759.17                  & 21.57                 & 243.94            \\ \hline
    \end{tabular}
    \end{adjustbox}
    \caption{Average length of the input documents, queries and output summaries for the Debatepedia and DUC 2005-7 datasets. For the DUC datasets, we add up the lengths of all the documents related the same query.}
    \label{tab:length_of_dataset}
\end{table}

In this work, we introduce a two-step architecture: 1) Retrieve answer-related sentences given the query, rank them by the confidence score (generated from Equation \ref{equ: confidence_score}) and concatenate them. 2) Use our QFS-BART to produce an abstractive summary.
\begin{equation}
    confidence\_score = P_{start}+P_{end},
    \label{equ: confidence_score}
\end{equation}

where $P_{start}$ and $P_{end}$ is two probability distributions over the tokens in the context. $P_{start}(i)$/$P_{end}(i)$ the probability of the $i$-th token is the start/end of the answer span in context.

\begin{itemize}
    \item \textbf{Answer Retrieving}
We split documents into paragraphs and feed each paragraph to the QA model to get answer-related sentences. Then the sentences are ranked by the confidence score.
    \item \textbf{Document Segmentation}
The QA model selects one answer span given an input document, and the sentences that contain the span will be chosen as the answer-related sentences. Since we only retain the answer-related sentences as input to the next step, we set the maximum paragraph length to 300 words to avoid missing too much information in this step. Specifically, we feed text to the paragraph sentence by sentence until it reaches the maximum length. 

    \item \textbf{Answer Relevance Ranking}
The paragraphs are fed to the QA model to generate answer-related sentences and the corresponding answer relevance scores. We align each sentence with a confidence score from the corresponding answer span. The confidence score is defined as:
    \item \textbf{Summary Generation}
We use the answer-related sentences and their answer relevance scores as the input to the QFS-BART model. The DUC 2005 dataset is used as a development set to optimize the model, and we evaluate the performance on the DUC 2006-7 dataset. We compare our QFS-BART with the following models:
    \begin{itemize}
        \item \textbf{LEAD.} \cite{xu2020query} returns all leading sentences of the most recent document up to 250 words.
        \item \textbf{TEXTRANK.} \cite{mihalcea2004textrank} is a graph-based ranking model that incorporate two unsupervised methods for keyword and sentence extraction.
        \item \textbf{HLTC-MRQA.} truncates the ranked answer related sentences from our first step as the extractive summary.
        \item \textbf{BART-CQA.} \cite{su2020caire} uses QA models for paragraph selection and iteratively summarizes paragraphs to 250 words.

    \end{itemize}

\end{itemize}

\begin{table}[!ht]
    \centering
    \begin{adjustbox}{width={0.6\textwidth},totalheight={\textheight},keepaspectratio}
    \begin{tabular}{l|cccccc}
    \hline
    \multicolumn{1}{l|}{\multirow{2}{*}{\textbf{Models}}} & \multicolumn{3}{c}{\textbf{DUC 2006}}                    & \multicolumn{3}{c}{\textbf{DUC 2007}}                    \\ \cline{2-7} 
    \multicolumn{1}{c|}{}                                 & \textbf{1} & \textbf{2} & \textbf{SU4} & \textbf{1} & \textbf{2} & \textbf{SU4} \\ \hline
    % \textbf{GOLD}                                         & 45.7             & 11.2             & 17.0               & 47.9             & 14.1             & 19.1               \\
    % \textbf{ORACLE}                                       & 40.6             & 9.1              & 15.8               & 41.8             & 10.4             & 16.0               \\
    % \multicolumn{7}{l}{\qquad \textit{Extractive approaches}}  \\
    \textbf{LEAD}                                         & 32.1             & 5.3              & 10.4               & 33.4             & 6.5              & 11.3               \\ 
    \textbf{TEXTRANK}                                     & 34.2             & 6.4              & 11.4               & 35.8             & 7.7              & 12.7               \\
    \textbf{HLTC-MRQA}                                    & 39.1             & 8.3              & 13.5               & \textbf{40.6}             & 9.6              & 14.7               \\ 
    \textbf{BART-CAQ}*                                    & 38.3             & 7.1              & 12.9               & 40.5             & 9.2              & 14.4               \\ \hline
    \textbf{BART-FT}                                      & 38.9             & 8.5              & 13.9               & 40.4             & \textbf{10.0}             & \textbf{15.1}               \\
    \textbf{QFS-BART}                                     & \textbf{39.4}             & \textbf{8.6}              & \textbf{14.1}               & 39.22              & 9.39              & 14.34                \\ \hline
    \end{tabular}
    \end{adjustbox}
    \caption{ROUGE-F1 scores for DUC 2006-7 dataset. Results with * mark are taken from the corresponding papers.}
    \label{tab: duc_results}
\end{table}

\textbf{Results and Discussion} We adopt ROUGE-F1 score \cite{lin2004rouge} as the evaluation metric. As shown in Table \ref{tab: duc_results}, HLTC-MRQA significantly outperforms the LEAD and TEXTRANK baselines, which indicates the effectiveness of our answer retrieval. However, QFS-BART does not perform well on DUC 2006-7 datasets. We conjecture the possible reason is because the answer relevance scores are not accurate on DUC dataset, thus affecting the performance of QFS-BART which highly replies on the relevance score. Including a trainable QA model together with QFS-BART might help improve that issue, and we explored it in the next Chapter.

\section{Summary}
In this Chapter, we first propose a real-time LFQA system, CAiRE-COVID, which combines information retrieval with state-of-the-art QA and query-focused multi-document summarization techniques, to answer high priority COVID-19 related questions. The system has shown its efficiency on the Kaggle CORD-19 Challenge, which was evaluated by medical researchers, and a series of experimental results also proved the effectiveness of our proposed methods and the competency of each module. The system is also easy to be generalized to general domain-agnostic literature information mining, especially for possible future pandemics. We have launched our website\footnote{https://caire.ust.hk/covid} for real-time interactions and released our code\footnote{https://github.com/HLTCHKUST/CAiRE-COVID} for broader use. 

Then, we propose QFS-BART, an abstractive summarization model for query focused summarization task. We use a generalizing QA model to make explicit answer relevance scores for all words in the document and combine them to the encoder-decoder attention. We also leverage pre-trained model (e.g. BART) and two-stage fine-tuning method which further improve the summarization performance significantly. Experimental results show the proposed model achieves state-of-the-art performance on Debatepedia dataset and outperforms several comparable baselines on DUC 2006-7 datasets.

% \chapter{An End-to-end Framework for Faithful Long-form Question Answering}
\chapter{Faithful to the Source: Generating Fact-aware Long-form Answers}

\begin{figure}[!ht]
 \centering
 \includegraphics[width=0.55\linewidth]{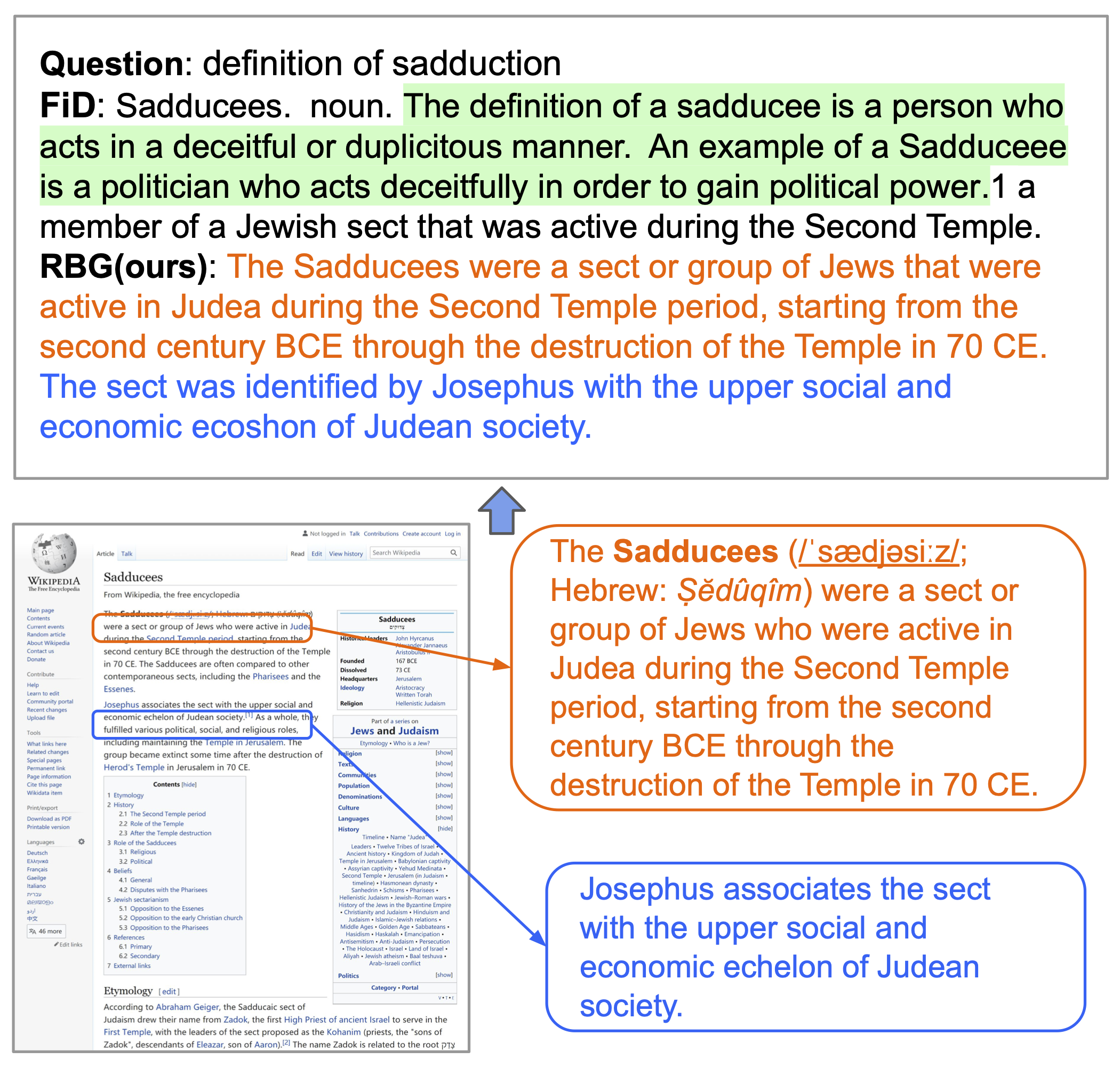}
  \caption{An example from MS MARCO~\cite{nguyen2016ms} dataset. We \hlgreen{highlight} the unfaithful snippets from other model. Our model(\textbf{RBG}) generate more factually accurate answer. }
  \label{fig:chap_rbg_example}
\end{figure}

While current work on LFQA using large pre-trained model for generation are effective at producing fluent and somehow relevant contents, one primary challenge lies in how to generate faithful answer that has less hallucinated facts (We show an example in Fig.~\ref{fig:chap_rbg_example}).

In this Chapter, we propose a novel end-to-end framework named RBG (\textbf{R}ead \textbf{B}efore \textbf{G}enerate) for LFQA to address the faithfulness challenge. The key idea is to augment the generation process with predicted salient information which can be viewed as an emphasis on answer-related facts. Specifically, we combine a Seq2Seq language model-based generator with a machine reading comprehension (\textit{reader}) module. The \textit{reader} produces an evidence probability score for each sentence, which will be integrated with the generator for final distribution prediction. We perform evidence fusion in a similar way to FiD~\cite{izacard2021leveraging} to equip the pre-trained language model with multiple input documents for generation. To further enhance the factual grounding ability of the generation model, we propose an additional pre-training task to encourage the model to rely more on retrieved documents to generate factual statements. The details are explained in Section \ref{sec:methodology}.

\section{A state-of-the-art Long-form Question Answering System}
\label{sec:methodology}
\begin{figure*}[!t]
 \centering
 \includegraphics[width=1.0\linewidth]{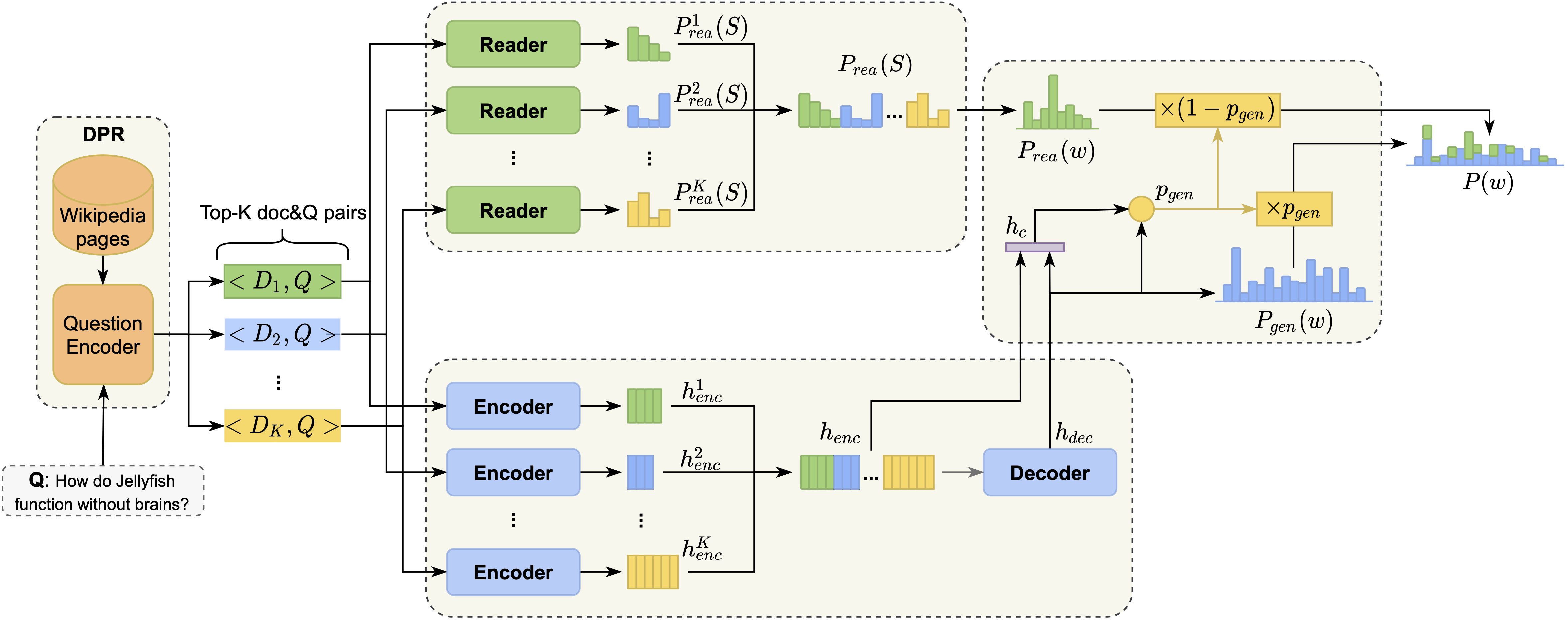}
  \caption{Overview architecture of our RBG framework. RBG comprises a supporting document retriever, a document reader and a generator.}
  \label{fig:chap_rbg_framework}
\end{figure*}

To generate in-depth, long-form answers for a given general domain question, we first use a retriever to search for relevant information from a large external knowledge source. Then our reader and the generation module take the multiple retrieved documents together with the question as input to generate the answer. Specifically, the reader module adopts a machine reading comprehension (MRC) model to produce an evidence score for each sentence in each document, while the generator, which adopts a large pre-trained Seq2Seq language model, fuses the sentence evidence score into its generation process. Our framework is shown in Figure~\ref{fig:chap_rbg_framework}.

\subsection{Supporting Document Retriever}

We use the dense passage retriever (DPR)~\cite{karpukhin2020dense} to retrieve the supporting documents following the typical methods in the state-of-the-art framework for open-domain QA~\cite{izacard2021leveraging,NEURIPS2020_6b493230}. 

The passage and question are represented as 768-dimensional dense vector representations, computed via the BERT-based bi-encoder networks of DPR. The retriever will rank the documents according to their similarity, calculated as
\begin{equation}
\label{dpr}
    sim(Q, D_i) = \text{BERT}_q(Q)^T \text{BERT}_d(D_i)
\end{equation}

Retrieval is performed using approximate nearest neighbors with the FAISS~\footnote{\url{github.com/facebookresearch/faiss}} library. We denote $D=\{D_1, D_2,..., D_k\}$ as the top-$K$ retrieved documents for question $Q$. 

As the question/answers in LFQA may cover different domains and topics, we use a multi-task variant of DPR to guarantee the retrieval performance. The retriever is trained jointly on the union of all knowledge-intensive training data in KILT benchmark~\cite{petroni2021kilt}, including TrivaQA~\cite{Joshi_2017}, kwiatkowski2019naturaluestion~\cite{kwiatkowski2019natural}, HotpotQA~\cite{yang2018hotpotqa}, Fever~\cite{thorne2018fever}, zsRE~\cite{levy2017zero}, AY2, T-REx~\cite{elsahar2018t} and WoW~\cite{dinan2018wizard} (More details can be found in Appendix~\ref{appendix:rbg_retriever}).

\subsection{Document Reader}
Since there are no golden retrievals for long-form answers, the retrieved documents may contain complementary, contradictory, or redundant information related to the answer. Thus, we propose to use a reader module to explicitly predict the sentence-level evidence probability in each document.  

\textbf{Evidence span prediction}
We use a machine reading comprehension (MRC) model to predict the evidence span in each document, as these models approach or even outperform human-level performance on many datasets~\cite{joshi2020spanbert}. The MRC model takes the concatenation of the retrieved document $D_i$ and question $Q$ as input, and outputs the prediction of the start and end position of the potential evidence spans in $D_i$. Specifically, it outputs two probability distributions over the tokens in $D_i$: $P^s_i(w_s)$ and $P^e_i(w_s)$, where $P^s_i(w_s)$ / $P^e_i(w_s)$ is the probability that the token $w_s$ is the start/end of the evidence span in $D_i$. 

\textbf{Sentence evidence probability}
Originally, the MRC model was designed to give accurate, short-phrase span prediction~\cite{rajpurkar2016squad}, but we argue that a sentence-level evidence probability will be better in our scenario. The supporting sentences can provide the minimum required context information for each answer span, which is quite important, especially in multi-document generation~\cite{xu2020coarse}. We define our sentence-level evidence probability score $P^i_{rea}(S)$ as the summation over all token-level evidence probabilities in that sentence, and it is calculated via
\begin{align}
    P^i_{rea}(S) &= \sum\nolimits_{w_s\in S}(P^s_i(w_s) + P^e_i(w_s)) \\
    P_{rea}(S) &= \text{Norm}(P^1_{rea}, P^2_{rea}, ..., P^K_{rea}) 
    % P_{rea}(w_i) &= P(S_i)    \text{if} w_i \in S_i 
\end{align}
where $P_{rea}(S)$ denotes the final sentence-level evidence probability in all the $K$ documents regarding the question. 

\textbf{Multi-task MRC}
As there are no golden answer spans for LFQA data, we need a MRC model that has enough generalization ability for open domain questions as a starting point. We choose SpanBERT~\cite{joshi2020spanbert}, and further fine-tune it in a multi-task way on six large-scale MRC datasets from the MRQA shared task ~\cite{fisch-etal-2019-mrqa} following work by ~\cite{su-etal-2019-generalizing}: SQuAD~\cite{rajpurkar2016squad}, NewsQA~\cite{trischler2017newsqa}, TriviaQA~\cite{Joshi_2017}, SearchQA~\cite{dunn2017searchqa}, HotpotQA~\cite{yang2018hotpotqa}, and NatualQuestions~\cite{kwiatkowski2019natural}. The MRC model $R$ is jointly trained with the generator, using the golden answer in a distantly supervised way.

\subsection{Generator}

\textbf{FiD-BART} We choose BART as our generation backbone because of its outstanding performance on many generation tasks, especially on long-form abstractive summarization task~\cite{lewis2020bart}. We propose FiD-BART, following the \textit{Fusion-in-Decoder} idea from ~\cite{izacard2021leveraging}, to empower BART to deal with multiple, long-document inputs. FiD-BART processes each document independently in the encoder, while performing the cross-attention in the decoder jointly. 

The encoder encodes the concatenation of each supporting document $D_i$ and the question $Q$. More precisely, we append the special tokens \textit{question:} before $Q$, \textit{title:} and \textit{context:} before the title and text of each document $D_i$. We denote the encoded final representation of the encoder as $h_{enc}$,  which is the concatenation of the $K$ encoder outputs $h^i_{enc}$ for the $i$th document:
\begin{align}
\setlength\abovedisplayskip{10pt}
\setlength\belowdisplayskip{-10pt}
    h^i_{enc} &= \text{Encoder}(Q;D_i) \\
    h_{enc} &= (h^1_{enc},..., h^i_{enc},..., h^K_{enc}) 
\end{align}
The partial structure of the decoder can be illustrated by Eq.(6)--(8), where $h_l$ is the representation for the $l$-th decoder layer. We denote $h_{dec}$ as the last layer decoder outputs: 
\begin{align}
\setlength\abovedisplayskip{5pt}
\setlength\belowdisplayskip{3pt}
    h_l^a &= \text{SelfAttention}(h_l, h_l, h_l) \\
    h_l^b &= \text{LayerNorm}(h_l + h_l^a) \\
\label{cross-attention}    h_l^c &= \text{CrossAttention}(h_l^b, h_{enc}, h_{enc})
\end{align}
As we can see, FiD-BART can scale to a large number of input documents within a linear computation time.

\subsection{Reader-Before-Generator}
To incorporate the evidence probability into generation, we apply the pointer-generator model (depicted in Figure~\ref{fig:chap_rbg_framework}). The attention distribution $\mathcal{A}$ and context vector $h_c$, and the generation probability $p_{gen}$ $\in$ [0,1] are calculated as follows:
\begin{align}
    \mathcal{A} &= softmax(h_{dec} h^T_{enc}) \\
    h_{c} &= \mathcal{A}  h_{enc} \\
    p_{gen} &= sigmod(W_{c}h_{c} + W_{g}h_{dec}) 
\end{align}
where $W_c$ and $W_g$ are learnable parameters. $p_{gen}$ is used as a soft switch to choose between generating a word from the generator by sampling from the vocab, or copying a word from the input sequence by sampling according to the evidence distribution $P_{rea}(w)$:
\begin{align}
&P_{gen}(w) = lm_{head}(h_{dec}) \\
&P_{rea}(w) = \sum\nolimits_{s:w_s=w\text{,}w_s \in S}{P_{rea}(S)} \\
&P(w) = p_{gen}P_{gen}(w) + (1-p_{gen})P_{rea}(w)
\end{align}

\def\mask{{\sc[mask]}}

\subsection{Pre-training}
To further improve the ability to ground on retrieved documents, we propose a pre-training task: retrieval-augmented recovery~(RAR). Instead of recovering the corrupted text through the internal knowledge memorized in model parameters~\cite{raffel2020exploring,lewis2020bart}, RAR encourages the model to rely more on external retrieved documents to generate factual statements. Specifically, given an original text $S$, we retrieve the top-$k$ documents ${D_1,D_2,...,D_N}$ from the knowledge corpus using BM25~(discarding $S$ itself), and we replace 30\% of the words in $S$ with \mask to form a pseudo query $Q$. The pre-training task asks our RBG model to recover $S$ with the input of the pseudo query $Q$ and $k$ retrieved documents, which can be formulated as
\vspace{-5pt}
\begin{equation}
    S=RBG(Q;D_1,D_2,...,D_k)
    \vspace{-5pt}
\end{equation}
\noindent To involve more factual information during the text corruption and recovery process, we sample 1 million sentences of $S$ corresponding to at least one knowledge base triplet from Wikipedia with the text-triple alignment of TREX~\cite{TREX}.  
%We find that different reference documents can provide factual evidence to recover different masked words. This is similar to the downstream LFQA situation that the model need to combine evidence in different retrieved documents to generate the final answer.(..tbd details)
\section{Experiment Setups}

\subsection{Datasets}
We conduct experiments on the two following datasets, both of which concentrate on long form generative QA.
\begin{itemize}
    \item \textbf{ELI5}~\citet{fan2019eli5} is the only publicly available large-scale LFQA dataset. It is a collection of
question-answer pairs extracted from the Reddit forum "Explain Like
I’m Five"(ELI5). We use the KILT~\cite{petroni2021kilt}  version of the dataset from its Github repository\footnote{\url{github.com/facebookresearch/KILT}}, which
has 272,634 training examples and 1,507 development examples. The average length of the answers is 130 words.
    \item \textbf{MS MARCO}~\citet{nguyen2016ms} is a dataset of crowdsourced responses to Bing
queries. We use the question-answer pairs of the MS MARCO passage ranking track for training and evaluation, as they are more abstract and reliant on multi-document information than those of the NLG track. The training example size is about 500,000 and the evaluation example size is 6980. 
    \item \textbf{Knowledge source} The external knowledge source of the retriever is the Wikipedia paragraphs, which are provided in the KILT benchmark as a unified knowledge source for knowledge-intensive tasks, including open-domain LFQA ~\cite{petroni2021kilt}. It is based on the 2019/08/01 Wikipedia snapshot, and contains 5.9M articles. 

\end{itemize}

\subsection{Baselines}
\begin{itemize}
    \item \textbf{BART and T5} We fine-tune BART~\cite{lewis2020bart} and T5~\cite{raffel2020exploring} using QA pairs without explicit retrieval, and include them as our baselines which rely only on parameterized internal knowledge~\cite{roberts2020much} to generate answers.
    \item \textbf{}{DPR-BART} is our retrieval-based LFQA baseline. We follow ~\citet{petronicontext} to retrieve and prepend the top-3 passages from DPR for each input sample, and use context-enhanced training data to fine-tune a BART model.
    \item \textbf{RAG}~\citet{NEURIPS2020_6b493230} is an end-to-end retrieval-augmented generation model which back-propagates to the retriever’s input encoder. We experiment with fine-tuning RAG on LFQA tasks, establishing a strong baseline on all of them.  At every generation step we retrieve the top-5 passages and use them as supporting documents.
    \item \textbf{FiD} ~\citet{izacard2021leveraging} encodes each passage independently and combines all outputs from the encoder before passing them to the decoder. FiD has achieved superior performance on a number of open-domain QA tasks~\cite{izacard2021leveraging}. We implement FiD-BART, using BART as the generation backbone, as our strongest baseline.

\end{itemize}

\subsection{Implementation Details}

\textbf{Document Retriever Model Details} As the question/answers in LFQA may cover different domains and topics, we use a multi-task variant of DPR to guarantee the retrieval performance. The retriever is trained jointly on the union of all knowledge-intensive training data in KILT benchmark~\cite{petroni2021kilt}, including TrivaQA~\cite{Joshi_2017}, naturaluestion~\cite{kwiatkowski2019natural}, HotpotQA~\cite{yang2018hotpotqa}, Fever~\cite{thorne2018fever}, zsRE~\cite{levy2017zero}, AY2, T-REx~\cite{elsahar2018t} and WoW~\cite{dinan2018wizard}.

% As there is no golden retrievals provided in current LFQA datasets.

\section{Experiment Results}

\subsection{Automatic Evaluation}

We use the metrics unigram F1 score and ROUGE-L~\cite{lin2004rouge} in previous work on LFQA ~\cite{petroni2021kilt, krishna2021hurdles} to evaluate and compare the generation quality of our method. 

\textbf{Overall Comparison} Table~\ref{results} shows the performance of various methods on the two datasets. As shown, our RBG method outperforms all baselines models with regard to both evaluation metrics on both datasets. The RBG method also outperforms the previous state-of-the-art method \textit{c-REALM+RT} on the KILT-ELI5 leaderboard\footnote{\url{ https://evalai.cloudcv.org/web/challenges/challenge-page/689/leaderboard/1908}} ~\cite{krishna2021hurdles}, as shown in Table~\ref{leaderboard}.

\begin{table}[!ht]
\centering
\resizebox{0.62\textwidth}{!}
{
\begin{tabular}{c|cc|cc}
\hline
Models  & \multicolumn{2}{c}{Eli5} &\multicolumn{2}{c}{MS MARCO}\\  
            & ROUGE-L      & F1         & ROUGE-L       & F1 \\ \hline
T5(base)    & 21.02        & 18.36       & 21.19 & 20.03 \\
BART(large) & 22.69        & 22.19      & 23.26 & 25.6  \\
DPR+BART    & 17.41        & 17.88      & 23.01 & 25.13     \\
RAG         & 16.11        & 17.24      &   -    & -  \\
FiD  &    25.70      &    28.55    &   24.64    & 27.08   \\
RBG(ours)   & \textbf{26.46}        & \textbf{29.04}      & \textbf{24.72} & \textbf{27.52} \\ \hline
\end{tabular}
}
\caption{Performance comparison between our RBG method and the baselines on the KILT-ELI5~\cite{petroni2021kilt} and MS MARCO~\cite{nguyen2016ms} evaluation sets.}
\label{results}
\end{table}

\begin{table}[!ht]
\centering
\resizebox{0.62\textwidth}{!}
{

\begin{tabular}{c|ccccc}
\hline
Model           & \multicolumn{2}{c}{Retrieval} & \multicolumn{2}{c}{Generation} &      \\
                & PRr.       & R@5        & F1           & R-L          & KRL  \\ \hline
RBG(ours)       & 10.83         & 27.25         & \textbf{24.53 }         & \textbf{27.13}         & \textbf{2.62} \\
DPR\_kilt\_wiki &        14.83       &   27.69            &      16.45          &       15.91        &    2.46  \\
c-REALM$^1$  &        10.67       &      24.56         &        23.19        &      22.88         &   2.36   \\
DPR+BART        &    10.67           &       26.92        &       17.41         &        17.88       &  1.90    \\
RAG             &        11.00       &      22.92         &      14.05          &      14.51         &   1.69   \\
BART-large      &          0.00     &         0.00     &     20.55            &    19.23            &     0.00  \\ 
T5-base         &      0.00         &       0.00        &       19.08         &      16.10         &   0.00   \\
\hline
\end{tabular}

}
\caption{Results on the ELI5 test set on the KILT leaderboard. Our RBG tops the leaderboard in terms of (1) retrieval performance, using R-precision(RPr.) and Recall@5(R@5), and (2) generation quality, using F1 and ROUGE-L(R-L). These scores are combined to produce the overall metric KILT R-L(KRL)~\cite{petroni2021kilt}. c-REALM$^1$ is from~\cite{krishna2021hurdles}}
\label{leaderboard}
\vspace{-10pt}
\end{table}

\textbf{Fine-grained Comparison} Intuitively, the quality of retrieved documents will affect the generation quality, thus we provide a fine-grained performance comparison. We split MS-MARCO evaluation set into different subset based on the quality of the retrieved documents\footnote{\label{foot:retrieve} We consider two metrics to measure the retrieval quality for a certain question: (1)~\textbf{Top-1 document retrieval score} which is the matching score output by the retriever (Equation.~\ref{eqa:retriever_score}) for the top-1 document to measure the corresponding semantic relevance to the given question, and (2)~\textbf{N-gram overlap}, which is the N-gram overlap between the golden answer and the top-k retrieved documents.}, and compare the ROUGE-L score between FiD and RBG under each subset.

As we can see from Table~\ref{tab:FidvsRBG}, even though RBG beats FiD by 0.1 Rouge-L score on the whole MS-MARCO evaluation set, the performance gap continue increasing as the retrieval quality of the evaluation subset increased. This indicates that RBG is especially effective when high-quality retrieval documents is provided, which matches with our intuition. 

\begin{table}[!h]
\centering
\resizebox{0.60\textwidth}{!}
{
\begin{tabular}{lc|cccc}
\hline
\multicolumn{2}{l|}{>ngram overlap} & 0     & 0.4        & 0.6       & 0.8    \\ \hline
\multicolumn{2}{l|}{\# of documents} & 6980  & 3493      & 1470      & 489    \\ \hline
\multirow{2}{*}{ROUGE-L}    & FiD    & 24.64 & 28.04  & 33.62  & 45.25 \\
                            & RBG    & 24.72 & 28.59  & 34.38   & 46.29  \\ \hline
\end{tabular}
}
\end{table}
\vspace{-20pt}
\begin{table}[!h]
\centering
\resizebox{0.60\textwidth}{!}
{
\begin{tabular}{lc|cccc}
\hline
\multicolumn{2}{l|}{\textgreater{}retrieval score} & 0.0   & 75    & 80    & 85    \\ \hline
\multicolumn{2}{l|}{\# of documents}               & 6980  & 5811  & 3188  & 1001  \\ \hline
\multirow{2}{*}{ROUGE-L}           & FiD           & 24.64 & 24.7  & 25.63 & 26.81 \\
                                   & RBG           & 24.72 & 25.46 & 26.53 & 27.96 \\ \hline
\end{tabular}
}
\caption{Fine-grained comparison between FiD and RBG on different subset of MS-MARCO evaluation data.}
\label{tab:FidvsRBG}
\end{table}

% As we can see in Table~\ref{retrieval_results}, better-retrieved documents always bring better generation quality, indicating the importance of high-quality supporting documents for the generation process.

\vspace{-10pt}
\subsection{Human Evaluation} 
We further evaluate our model using human annotators, who we ask to quantify three aspects of the generated answer, (1)~\textbf{fluency}, which measures whether the answer is coherent and less repetitive; (2)~\textbf{relevance}, which measures the amount of information relevant to answering the question, and (3)~\textbf{factual correctness}~(also briefly called correctness), which measures the correctness and faithfulness of all facts involved in the generated answer.

We select FiD, which is the strongest baseline in terms of automatic metrics, for comparison. We sample evaluation questions from the MS MARCO dev set, which are better supported by Wikipedia knowledge than ELI5. Table~\ref{tab:human eval abs} shows the absolute evaluation results of human annotation. To reduce the impact of scale selection inconsistency of different annotators, we also show the relative evaluation results in Table~\ref{tab:human eval rel}. We can see that both types of results indicate that RBG outperforms FiD in terms of all three aspects. RBG has more advantages over FiD on the metric of factual correctness, possibly benefited by the introduction of the reader module and additional pre-training. More details of the human evaluation setup and statistical analysis can be found in Appendix~\ref{appendix:rbg_human_eval}.

\begin{table}[htbp]
\centering
\resizebox{0.60\textwidth}{!}
{
\begin{tabular}{c|ccc}
\hline
Model    & Fluency      & Relevance      &  Correctness   \\ \hline
FiD  &    2.62      &    2.34    &    2.07    \\
RBG(ours)   & \textbf{2.70}        & \textbf{2.50}     & \textbf{2.41} \\ \hline
\end{tabular}
}
\caption{Absolute human evaluation results for RBG vs. FiD on MS MARCO. The table shows the mean value across all annotators and examples for each metric.}
\label{tab:human eval abs}
\end{table}

\begin{table}[ht]
\centering
\resizebox{0.58\textwidth}{!}
{
\begin{tabular}{c|ccc}
\hline
Aspect    & Prefer FiD      & Prefer RBG      &  Tie  \\ \hline
Fluency  &    12\%   & \textbf{26\%}   &    62\% \\
Relevance  &  18\%     & \textbf{48\%} & 34\% \\
Correctness &  4\%     & \textbf{62\%} & 34\% \\ \hline
\end{tabular}
}
\caption{Relative human evaluation results for RBG vs. FiD on MS MARCO. The percentages represent the ratio of one model being voted as preferred by multiple annotators on a metric.}
\label{tab:human eval rel}
\end{table}

\subsection{Ablation}
To further investigate the contribution and effect of each module in the proposed system, we conducted a systematic ablations on the MS-MARCRO evaluation dataset. 

\begin{table}[!th]
\centering
\resizebox{0.60\textwidth}{!}
{
\begin{tabular}{c|c|cc}
\hline
No. & models     & \multicolumn{2}{c}{MS MARCO} \\ 
  &   & ROUGE-L    & \multicolumn{1}{l}{F1} \\ \hline
0 & RBG(ours)    & 24.72 & 27.52                  \\ 
1 & w/o reader &  24.66 & 27.30 \\
2 & w/o pre-training  & 24.65  & 27.38   \\
3 & w/o reader + pre-training & 24.64  & 27.08 \\
4 & w/ reader frozen &  24.51 & 25.85 \\
5 & w/ random retrieval  & 22.84  & 25.23  \\
\hline
\end{tabular}
}
\caption{Ablation results on the MS MARCO evaluation set. A more fine-grained results comparison is shown with analysis in Section ~\ref{sec:further}.}
\label{ablation}
\vspace{-15pt}
\end{table}
\begin{itemize}
    \item \textbf{\textit{w/o} reader/pre-training:} We respectively remove the reader module (\textbf{w/o reader}), the pre-training (\textbf{w/o pre-training}), and both together (\textbf{w/o reader + pre-training}) from our model , to test the contribution of each part. As we can see from Table~\ref{ablation}, without the reader to predict the evidence probability, the generation performance decreases in both metrics, and the performance continues to drop without the pre-training. 
    \item \textbf{\textit{w/} reader frozen:}  We freeze the reader to investigate the benefit of distantly supervised end-to-end training of the reader module. As we can see from Table~\ref{ablation}, the results on both metrics drop, especially the F1 score, which proves the effectiveness of the end-to-end training.
    \item \textbf{\textit{w/} random retrieval:} To investigate whether and how much the generation process is grounded in the retrieved documents, we replace retrieved paragraphs with randomly sampled paragraphs from Wikipedia at \textit{inference} time for comparison. As we can see, the ROUGE-L drops significantly with randomly retrieved documents, and it is also worse than the baseline systems such as BART and DPR-BART (Table~\ref{results}).

\end{itemize}

\section{Further analysis} 
\vspace{-5pt}
\label{sec:further}
We conduct further analysis on the results, considering that LFQA is a complicated but less explored task, which deserves a complete investigation. 
\vspace{-5pt}
\subsection{How does retriever affect the generation quality?}
\vspace{-3pt}
We further investigate the effects of the quality of retrieved documents on the final generation. We split the evaluation sets of the two datasets via different thresholds for the two metrics\textsuperscript{\ref{foot:retrieve}} and calculate the corresponding ROUGE-L score for each subset. As we can see in Table~\ref{retrieval_results}, better-retrieved documents always bring better generation quality, indicating the importance of high-quality supporting documents for the generation process.

% We also measure the effects of the number of retrieved documents $K$ on the generation quality and find that the best $K$ from $\{5,10,20,50\}$ is 10 (see more details in Appendix~\ref{appendix:rbg_k}). More retrieved documents do not improve generation quality as in open-domain QA.

\subsection{Number of Retrieved Documents on Generation Quality}
\label{appendix:rbg_k}
\begin{table}[!ht]
\centering
\begin{tabular}{c|cc}
\hline
ndocs & ROUGE-L & F1    \\ \hline
5     & 24.63   & 27.29 \\
10    & \textbf{24.72}   & \textbf{27.52} \\
20    & 24.39   & 26.68 \\
50    & 23.43   & 25.94 \\ \hline
\end{tabular}
\caption{Generation performance versus the number of retrieved documents of our model on MS MARCO~\cite{nguyen2016ms}.}
\label{tab:results_k}
\end{table}

We also investigate the effects of number of retrieved documents $k$, on the answer generation quality. As we can see in Table~\ref{tab:results_k}, the generation quality in terms of ROUGE-L and F1, do not further improve as the number of $k$ increases, and the best performance are obtained when $k=10$ in our case.

\begin{table}[!h]
\centering
\resizebox{0.60\textwidth}{!}
{
\begin{tabular}{c|cc|cc}
\hline
\multicolumn{1}{l|}{\multirow{2}{*}{\begin{tabular}[c]{@{}l@{}}\textgreater{}retrieval\\ score(top-1)\end{tabular}}} & \multicolumn{2}{c|}{ELI5}                                    & \multicolumn{2}{c}{MS MARCO}                                 \\ \cline{2-5} 
\multicolumn{1}{l|}{}                                                                                                & \multicolumn{1}{l}{\# of data} & \multicolumn{1}{l|}{ROUGE-L} & \multicolumn{1}{l}{\# of data} & \multicolumn{1}{l}{ROUGE-L} \\ \hline
0.0 & 1570  & 26.35  & 6980  & 24.72 \\
75 & 1270 & 26.37 & 5811  & 25.46                      \\
80 & 479  & 26.38 & 3188  & 26.53                      \\
85  & 72  & 26.96   & 1001     & 27.96                      \\
90 & 11    & 27.25   & 161  & 27.61                      \\ \hline
\end{tabular}
}
\end{table}
\begin{table}[!h]
\centering
\resizebox{0.60\textwidth}{!}
{
\begin{tabular}{c|cc|cc}
\hline
\multicolumn{1}{l|}{\multirow{2}{*}{\begin{tabular}[c]{@{}l@{}}\textgreater{}ngram \\ overlap\end{tabular}}} & \multicolumn{2}{c|}{ELI5}                                    & \multicolumn{2}{c}{MS MARCO}                                 \\ \cline{2-5} 
\multicolumn{1}{l|}{}                                                                                        & \multicolumn{1}{l}{\# of data} & \multicolumn{1}{l|}{ROUGE-L} & \multicolumn{1}{l}{\# of data} & \multicolumn{1}{l}{ROUGE-L} \\ \hline
0.0  & 1570   & 26.35  & 6980  & 24.72 \\
0.4 & 460 & 27.09& 3493 & 28.59  \\
0.5 & 260& 27.31 & 2470 & 30.72 \\
0.6 & 109  & 27.52 & 1470 & 34.38 \\
0.7 & 48 & 27.63 & 845  & 39.64  \\
0.8& 27  & 27.17& 489  & 46.29  \\ \hline
\end{tabular}
}
\vspace{-5pt}
\caption{Fine-grained results of our RBG on ELI5 and MS MARCO. With high-quality retrieval (higher N-gram overlap or retrieval score threshold), the answer quality (ROUGE-L) increases on both datasets.}
\label{retrieval_results}
\end{table}
\vspace{-10pt}

\vspace{-5pt}
\subsection{How does the reader contribute to the generation?}
\vspace{-5pt}

As shown in the ablation study, the reader module improves the overall performance on the MS MARCO evaluation dataset. We further investigate its performance when retrieved documents with different quality levels are provided. 

\begin{figure}[!t]
	\begin{center}
		\includegraphics[width=0.66\textwidth]{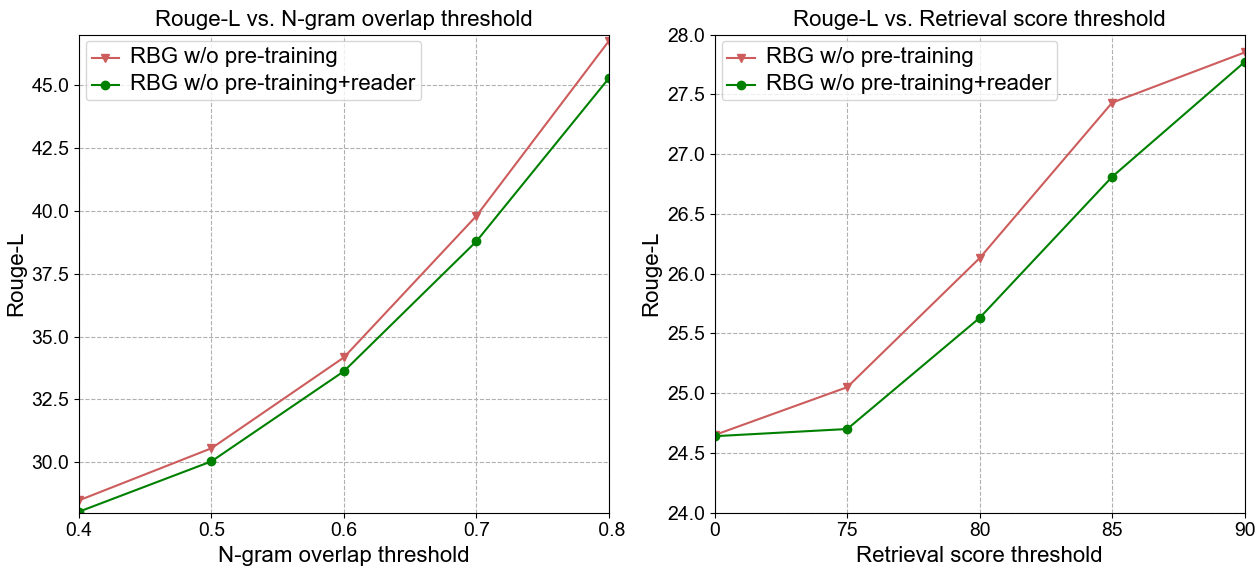}
		\caption{ROUGE-L versus document retrieval performance for reader analysis.}
		\label{Fig:decomp_for_reader}
	\end{center}
	\vspace{-0.2cm}
\end{figure}

We show in Figure~\ref{Fig:decomp_for_reader} the fine-grained comparison results between ablation models  \textbf{No.2}: \textit{RBG w/o pre-training} and \textbf{No.3}: \textit{RBG w/o pre-training + reader}. As we can see, the difference in ROUGE-L between the two models increases as the quality of the retrieved documents improves, indicating the reader's strong capability, especially on high-quality data. This also matches with our intuition. We also conduct a human evaluation for reader analysis, and we show the results in Table~\ref{tab:human eval reader}. 

\begin{table}[htbp]
\centering
\resizebox{0.60\textwidth}{!}
{
\begin{tabular}{c|ccc}
\hline
Aspect    & Prefer w/o reader      & Prefer w/ reader      &  Tie \\ \hline
Fluency  &    15\%    & \textbf{35\%}   &    50\% \\
Relevance  &  17\%     & \textbf{57\%} &26\% \\
Correctness &  25\%     & \textbf{45\%} & 30\% \\ \hline
\end{tabular}
}
\caption{Human evaluation results for RBG reader analysis on MS MARCO. The model with reader has better generation performance in terms of fluency, relevance and correctness.}
\label{tab:human eval reader}
\end{table}

\begin{figure}[t]
\vspace{-5pt}
	\begin{center}
		\includegraphics[width=0.66\textwidth]{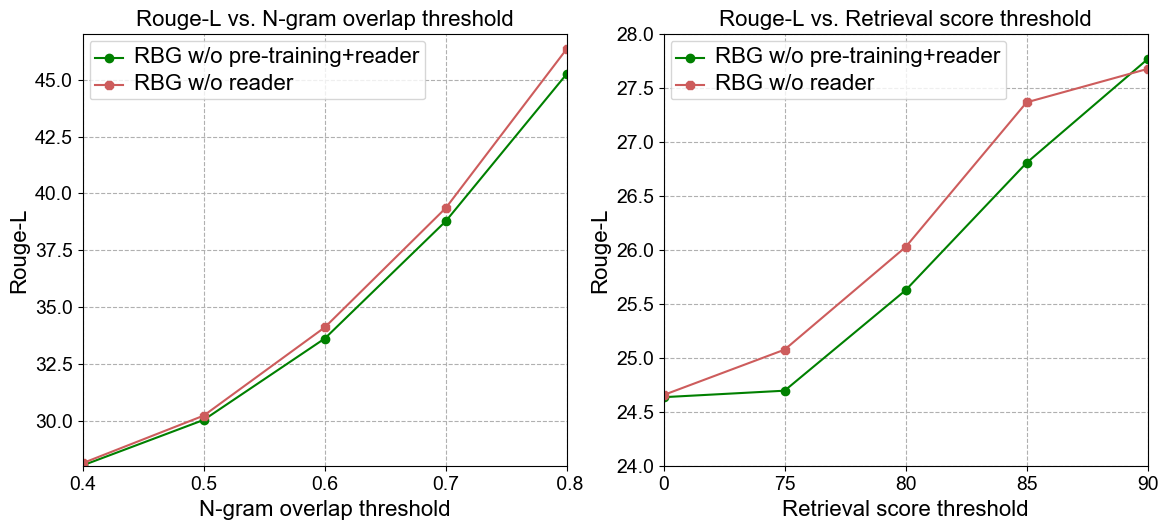}
		\vspace{-5pt}
		\caption{ROUGE-L versus Document retrieval performance for pre-training analysis.}
		\label{Fig:decomp_for_pretrain}
	\end{center}
	\vspace{-10pt}
\end{figure}

\vspace{-5pt}
\subsection{How does pre-training help?}
\vspace{-5pt}

\begin{table}[ht]
\centering
\resizebox{0.60\textwidth}{!}
{
\begin{tabular}{c|ccc}
\hline
Aspect    & Prefer w/o pre-training      & Prefer w/ pre-training      &  Tie \\ \hline
Fluency  &    40\%    & \textbf{43\%}   &    17\% \\
Relevance  &  20\%     & \textbf{33\%} & 47\% \\
Correctness &  23\%     & \textbf{47\%} & 30\% \\ \hline
\end{tabular}
}
\caption{Human evaluation results for RBG pre-training analysis on MS MARCO. The model with RAR pre-training has better generation performance in terms of relevance and correctness.}
\label{tab:human eval pretrain}
\end{table}

We also compare the models' performance in a fine-grained way, to quantify the contribution from our pre-training task. We show in Figure~\ref{Fig:decomp_for_pretrain} the fine-grained comparison results between ablation models \textbf{No.1}: \textit{RBG w/o reader} and \textbf{No.3}: \textit{RBG w/o pre-training + reader}. As we can see, the model with pre-training is better in most situations than that without pre-training. The human evaluation in Table~\ref{tab:human eval pretrain} also indicates the effectiveness of our pre-training task to improve the factual correctness and relevance of the generated answer. We conjecture that the pre-training task of retrieval-augmented recovery can facilitate the downstream LFQA model to combine multiple pieces of evidence from different retrieved documents to generate the final answer.

\subsection{Faithfulness analysis}

\textbf{Zero-shot on extractive QA tasks}  Inspired by previous work~\cite{wang2020asking, durmus2020feqa} which leverage a Question Generation(QG) and a QA model to generate question answer pairs, to evaluate the faithfulness of a summary\footnote{
They generate question answer pairs <$q$,$a_{sum}$> from the summary, and compare $a_{sum}$ with the answer $a_{sc}$ from source document for $q$, to evaluate faithfulness.}, we propose to evaluate answer faithfulness via evaluation on two simpler open-domain QA datasets: NaturalQuestions~\cite{kwiatkowski2019natural} and HotpotQA~\cite{yang2018hotpotqa}, which contain single-hop or multi-hop factual questions with golden answers ($\{(q_i, a^s_i)\}_{i=1}^m$) where $a_i^s$ can be extracted from Wikipedia-based documents. We use the trained models (based on MS MARCO) in Table~\ref{results} to do zero-shot long-form answer generation for these two datasets $\{a^l_i = \text{Model}_{\text{ms}}(q_i)\}$, and measure the short-answer recall~(the ratio of golden answer span $a^s$ contained in the generated long answer $a^l$) as an estimation of faithfulness of the generated long-answer:
\begin{equation}
Score(q, a^s, a^l) = \frac{\sum_{i=1}^m\mathbbm{1}[a^s_i \in a^l_i]}{m}
\label{eqa:faithfulness}
\end{equation}
We show the results in Table~\ref{nq_HotpotQA}. As we can see, our system achieves comparable performance with FiD on NQ, and it consistently outperforms other strong baselines on multi-hop dataset hotpotQA, indicating its capability in generating faithful answer especially on complex question that need to synthesis information. We also give concrete examples in Appendix~\ref{appendix:rbg_results} that show our model can generate more faithful snippets than FiD apart from automatic metrics. 

\begin{table}[!t]
\centering
\resizebox{0.55\textwidth}{!}
{
\begin{tabular}{c|cc}
\hline
          & NQ Recall   & HotpotQA Recall\\ \hline
T5         & 4.76  & 7.20     \\
BART-large & 10.44 & 9.13   \\
DPR+BART   & 16.37 & 11.57     \\
FiD      & 43.93 & 22.94    \\ \hline
RBG(ours)  & 43.93 & 23.36    \\ \hline
\end{tabular}
}
\caption{Faithfulness Analysis of the system generation quality via zero-shot evaluation on NQ~\cite{kwiatkowski2019natural} and HotpotQA~\cite{yang2018hotpotqa}.}
\label{nq_HotpotQA}
\end{table}

\textbf{Case Study} To have a concrete understanding of the reader's role to address faithfulness, we show two examples in Table~\ref{tab:reader_examples}. While both models use the same \textbf{ctxs}, RBG \textbf{w reader} generates a more \textit{relevant} answer, and \textbf{w/o reader} only gives some correct but unrelated statements (Case 1). The reader also contributes to generating more \textit{faithful} answers, as shown in Case 2. However, we notice that there is one unfaithful statement, which hallucinates the 'second-oldest' as 'oldest'. This might be because of the property of our MRC model: it gives a high evidential score to one sentence, and weakening the importance of other sentences in the same document.

\begin{table}[!ht]
% \small
\centering
\begin{adjustbox}{width={0.75\textwidth},totalheight={\textheight},keepaspectratio}
\begin{tabular}{p{1\columnwidth}}
\hline
\textbf{Question:} what does abc stand for in liquor store?  \\
\hline
\textbf{w/o reader answer:} ABC stores. A liquor store is a retail shop that predominantly sells prepackaged alcoholic beverages intended to be consumed off the store's premises. ...,or other similar terms. \\
\textbf{w/ reader answer}: ABC stands for Alcoholic Beverages Control (ABC Stores) and is a type of liquor store that is owned and operated exclusively by the state government, where liquor stores often sell only spirits or sometimes sell spirits and wine but not beer...\\
\textbf{ctxs:} [\{text: ...\textcolor[RGB]{5, 30, 255}{In seventeen alcoholic beverage control (ABC) states}, ...\},...,\{text:\textcolor[RGB]{5, 30, 255}{operated exclusively by the state government, where liquor stores often sell only spirits or sometimes sell spirits and wine but not beer.}, ...,because purchased liquor must be packaged in sealed bottles\},...,\{...\}]\\
\hline
\textbf{Question:} dow jones industrial average please? \\
\hline
\textbf{w/o reader answer:} The Dow Jones Industrial Average (DJIA) is an index of \hlgreen{30 major U.S. stock indexes}. The DJIA was created in 1896 by Charles Dow, a statistician and editor of The Wall Street Journal. \hlgreen{The index was first published in the mid-1880s, and was last updated on May 26, 1896.} \\
\textbf{w/ reader answer:} The Dow Jones Industrial Average (DJIA) is a stock market index that tracks the performance of the 30 largest U.S. industrial companies. It is owned by S\&P Dow Jones Indices, which is majority owned by Standard \& Poor's Global. The Dow is the \hlgreen{oldest} and most famous of the Dow averages. It was created by Charles Dow in 1896. \\
 \textbf{ctxs:} [\{text:...,was originally published on February 16, 1885. ...\textcolor[RGB]{5, 30, 255}{The industrial average was first calculated on May 26, 1896}..\},...\{text:...It is the \textbf{second-oldest} U.S. market index after the Dow Jones Transportation Average. \textcolor[RGB]{5, 30, 255}{Currently owned by S\&P Dow Jones Indices, which is majority owned by S\&P Global}..\},...,\{...\}]
 \\
 \hline
\end{tabular}
 \end{adjustbox}
\caption{Examples from MS MARCO dataset. We \textcolor[RGB]{5, 30, 255}{highlight} the sentences that have high evidential probability from the reader, and use \hlgreen{green} to mark out the unfaithful snippets. }
\label{tab:reader_examples}
\end{table}

\section{Summary}
In this Chapter, we propose a new end-to-end framework RBG that jointly models answer generation and machine reading to tackle the faithfulness issue in LFQA. Experiments on two LFQA datasets, ELI5 and MS MARCO, demonstrate the effectiveness of our method in comparison with strong baselines on automatic and human evaluation metrics. The detailed analysis further proves the competency of our method in generating fluent, relevant, and more faithful answers. We also propose to evaluate the factual correctness of LFQA model by answering questions of extractive QA tasks~(e.g., Natural Questions), which may be helpful to evaluate the faithfulness of LFQA model efficiently.
\chapter{Generating Succinct Answers from Long-form Answers}

A LFQA system should be an ‘empathetic machine’: the long-form answer should not only provide relevant and factual information, but also be succinct. While no prior work has been done on this direction, in this Chapter, we take the initial step by leveraging the generated long-form answers as an context to extract succinct short-phrase answers for extractive open-domain question answering task.

Specifically, we propose a framework named CGAP~\cite{su2022context}, which first generates a long-form answer leveraging the amount of parameterized knowledge stored in pre-trained language models, and then extract the short-span answer from the long-form answers without access to any external knowledge sources. Experimental results on three QA benchmarks show that our method significantly outperforms previous \textit{closed-book} QA methods, and is on par with traditional \textit{open-book} methods which extracts the answer from the retrieved documents.

\begin{figure}[!t]
 \centering
 \includegraphics[width=0.6\textwidth]{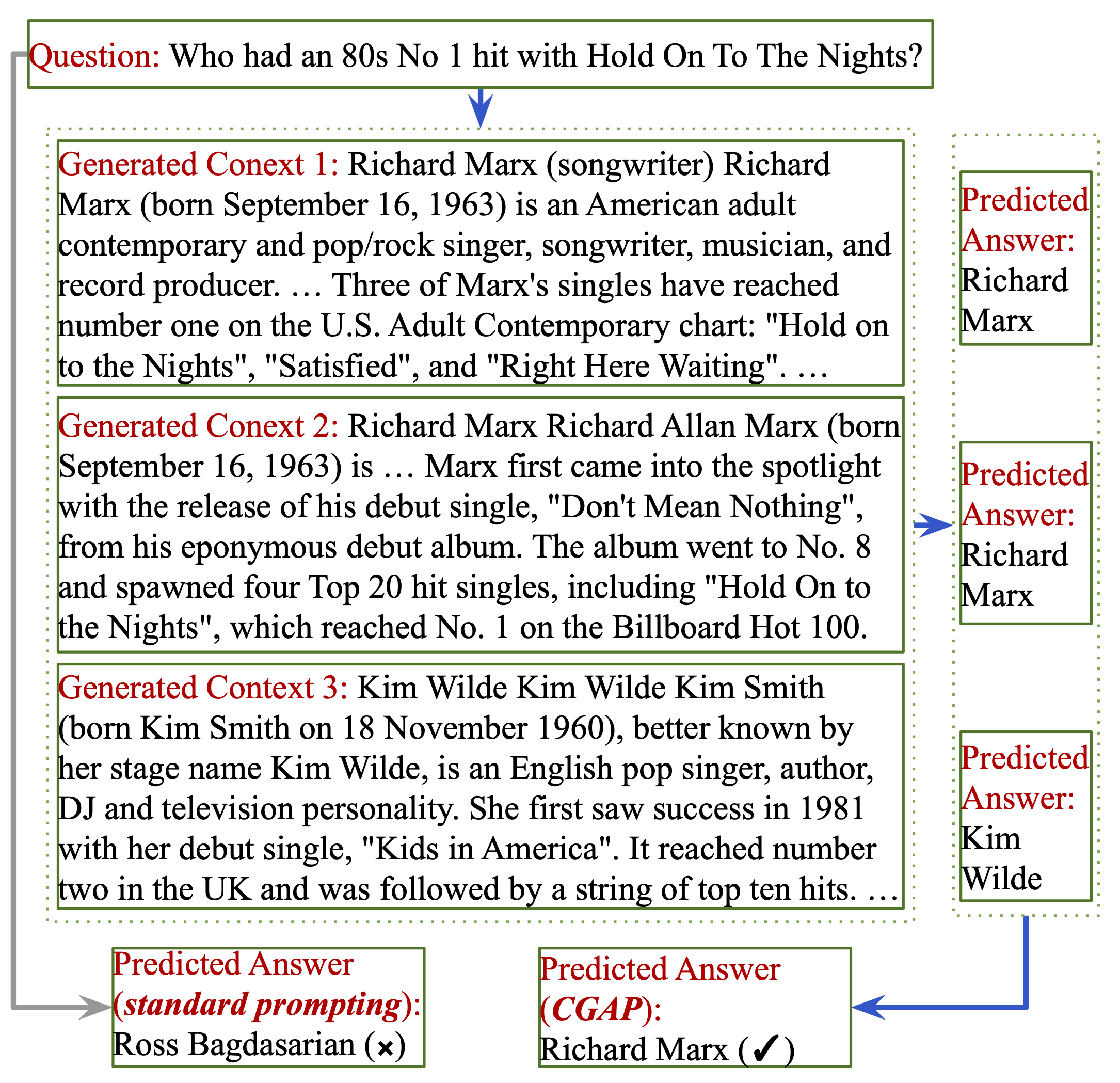}
  \caption{An example illustrating our two-stage, \textbf{CGAP} framework. CGAP generates more accurate answer (e.g. \textit{Richard Marx}) compared to standard few-shot prompting (e.g. \textit{Ross Bagdasarian}).}
  \label{fig:example}
\end{figure}

\section{Methodology}
\label{sec:cgap_methodology}
% \subsection{Main Idea}
% While it has been shown that large pretrained LMs store abundant knowledge~\cite{petroni2019language,roberts2020much}, we hypothesize the accuracy gaps are largely because the way of exploiting the parameterized knowledge are not sophisticated enough. Prior work on CBQA either finetunes pretrained LM models on entire QA datasets~\cite{roberts2020much, ye2020studying}, or they directly prompt those models using several few-shot QA pairs~\cite{brown2020language, radford2019language}. On the contrary, \textit{open-book} models use a two-stage pipeline. They first retrieve relevant contexts from external corpus, then they extract the answer based on the retrieved contexts.

% Therefore, to better exploit the parameterized knowledge in pretrained LMs and bridge the huge accuracy gaps between the \textit{closed-book} and \textit{open-book} methods, we propose a \textit{coarse-to-fine}, two-stage method for CBQA task. The main idea is to leverage generated contexts as the intermediate bridge between the huge amount of parameterized knowledge and answer. To the best of our knowledge, no previous investigation has been conducted on generating context from large pretrained LMs for CBQA and leveraging them to predict answer.

Our proposed framework \textbf{CGAP} consists of two stages. It first performs \textbf{C}ontext \textbf{G}eneration relevant to a given question by prompting a pretrained LM. Then it prompts the same LM for \textbf{A}nswer \textbf{P}rediction using the generated context and the question.  In order to mitigate failures caused by variability in the generated context, we generate multiple contexts, predicting the final answer by majority voting. Figure~\ref{fig:example} shows that CGAP generates 3 contexts and 3 predicted answers at the two stages respectively, and choose the most voted answer as the final answer. Note that we do not finetune the large pretrained LMs for context generation or answer prediction. This facilitates our approach to take advantage of gigantic LMs such as GPT-3~\cite{brown2020language}, PALM~\cite{chowdhery2022palm}  or Megatron-Turing NLG 530B~\cite{smith2022using}, which are only available through APIs.

\begin{figure*}[!ht]
 \centering
 \includegraphics[width=0.94\linewidth]{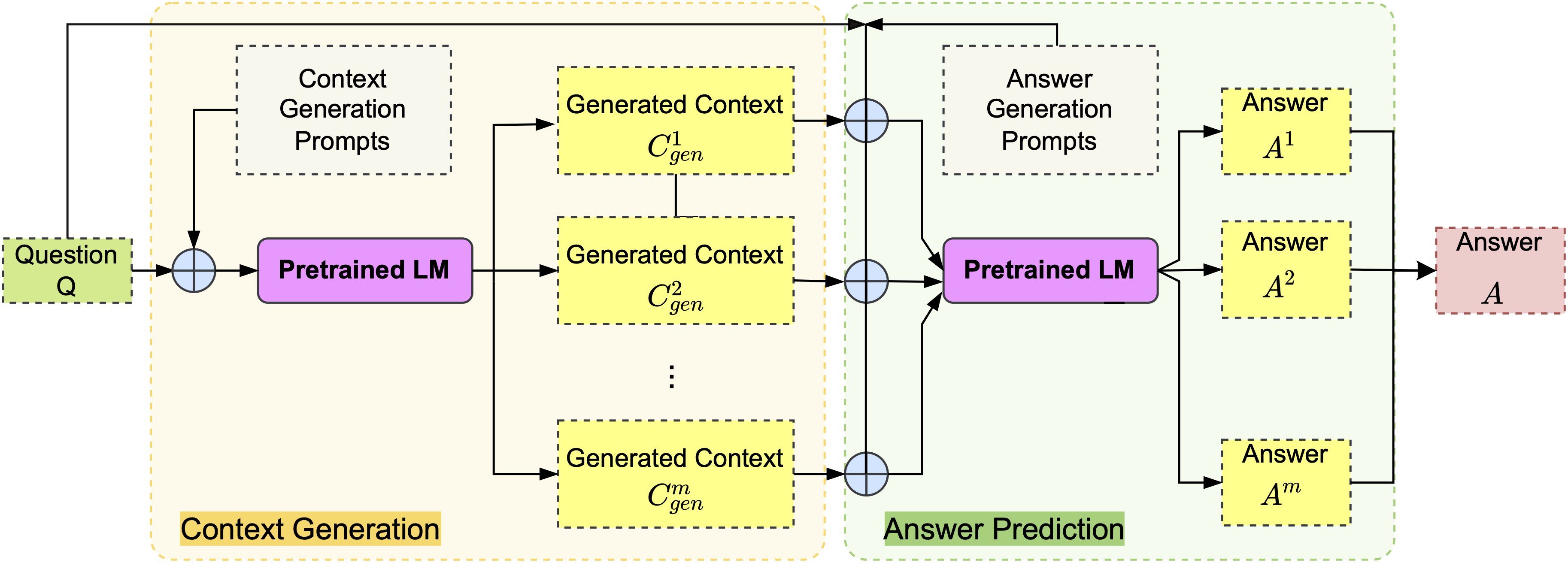}
  \caption{Overview architecture of our \textbf{CGAP} framework. It first does \textbf{C}ontext \textbf{G}eneration by prompting large pretrained LMs, then it further prompts the LMs for \textbf{A}nswer \textbf{P}rediction by feeding the generated context to the LM models alongside the question. $k$ contexts are generated and the final answer $A$ is chosen by majority voting. (\textit{If computation capability allows, it could prompt multiple ($k$) LMs in parallel at both two stages to speed up.}) }
  \label{fig:framework}
\end{figure*}

Our proposed \textbf{C}ontext \textbf{G}eration and \textbf{A}nswer \textbf{P}rediction, called \textbf{CGAP}, framework is illustrated in Figure~\ref{fig:framework}. CGAP consists of two stages. First, it generates relevant context to a given question by prompting a large pretrained LM. 
In the second stage, it predicts an answer using the generated context and the question by prompting the same LM. 
To accurately predict the answer, we generate multiple contexts. We run each of the two stages multiple times in parallel for the same question, generating different contexts for each, and use majority voting to select the final answer.

% Large LMs such as GPT-3~\cite{brown2020language}, Megatron-Turing-530B~\cite{smith2022using}, and PALM~\cite{chowdhery2022palm} have shown promising results in few-shot learning. They have also been shown as a good knowledge source~\cite{roberts2020much, liu2022multi}. However, finetuning them on downstream applications is challenging due to smaller datasets. Moreover, they are only available through API. In our \textbf{CGAP} framework, we therefore propose a modified few-shot prompting and demonstrate significant improvement over exiting finetuing or few-shot accuracies. 

% We denote the input question as $Q$, and the corresponding generated context as $C_{gen}$. We use  $C_{gen}^i$ to represent the $i$-th generated context samples that we marginalize on $(i=1.,,,k)$. The database of samples is denoted as $D$, and each data sample in $D$ is denoted by $d_i$ where $i=1,...N$, which consists N triples $<Q_i, C_i, A_i>$ , a question $Q_i$, a golden context passage $C_i$, and corresponding answer $A_i$.

Formally, for our task we have a question $Q$ to be answered, and a support repository $\mathcal{D} = \{(c_1, q_1, a_1), \ldots, (c_n, q_n, a_n)\}$ that consists of tuples of question $q_i$ and answer $a_i$ pairs with mapping to the context $c_i$.
In our experiments, we use the training sets of the corresponding datasets as $\mathcal{D}$.

As shown in Figure~\ref{fig:framework}, in the first stage, given question $Q$, we select $m$ the context generation prompts $S = \{(q_1, c_1), \ldots, (q_m, c_m)\}$ from the support repository $\mathcal{D}$.
We then use $S$ with $Q$ to prompt pretrained LM to generate $k$ contexts, which are denoted by $ C_{gen} = \{c^1_{gen}, c^2_{gen}, \ldots, c^k_{gen}\}$.
In the second stage, we select $m$ answer prediction prompts $S'=\{(q_1, a_1, c_1), \ldots, (q_m, a_m, c_m)\}$ from $\mathcal{D}$ and then we prompt the same LM using $C_{gen}$, $Q$ and $S'$.
The LM predicts a set of $k$ answers $A_p = \{a^1_p, a^2_p,..., a^k_p\}$ each corresponding to the $k$ contexts in $C_{gen}$. The final answer $A$ is selected by majority voting on $A_p$.

\subsection{Context Generation}
\label{sec:context_generation}

We would like to leverage the parameterized knowledge stored in large pretrained LMs to facilitate the answer generation process.
We first prompt the LM with few-shot examples to generate contexts that are relevant to the given question.

%For few-shot examples, we use the training dataset, $\mathcal{D}$ that consists of tuples of question ($\mathbf{Q}$), answer ($\mathbf{A}$), and context ($\mathbf{C}$). We select $m$ such tuples and denote them by $S=\{(Q_1, C_1), \ldots, (Q_m, C_m)\}$. %where $S_j = (Q_j, C_j)$. 

%First we select $m$ samples $S=\{S_1, S_2,..., S_m\}$ from a pool of database (i.e., the corresponding training dataset) for question $Q$, then we construct the prompts with the selected samples and prompt the LM for $C_{gen} = \{C^1_{gen}, C^2_{gen}, ..., C^k_{gen}\}$.
% \begin{equation}
%     C_{gen}^i = LM^i(Prompt(<S_1,...S_m>, Q))
% \end{equation}

% where each sample $S=<Q_i, C_i>$ contains a question $Q_i$ and a corresponding golden context passage $C_i$. 

\textbf{Sample Selection} Selecting appropriate samples as in-context prompts is the key to generate high-quality context relevant to the question. 
Previous work has shown that leveraging relevant samples helps the LM to generate contextually relevant and factually correct context~\cite{liu2021pre,liu2022multi}. We therefore use a similarity-based retriever to search relevant samples from the corresponding training dataset. 
Prior work has shown that DPR~\cite{karpukhin2020dense} achieves good performance in selecting the most relevant context for the question in ~\textit{open-book} models~\cite{izacard2021leveraging}. We therefore use DPR in our framework. In our DPR setup, we represent the question and the samples in the training dataset as $768$-dimensional dense vector representations, computed via the BERT-based bi-encoder networks. We rank the documents according to their similarity score, calculated as:
\begin{equation}
        Score(Q, (q_j, c_j)) = \text{BERT}(Q)^T \cdot \text{BERT}(q_j; c_j)
\label{eqa:retriever_score}
\end{equation}

where $;$ denotes concatenation of the tokens of the question $q_j$ and the context $c_j$. 
Finally, we get $S=\{(q_1, c_1), \ldots, (q_m, c_m)\}$ which are the top-m retrieved samples for question $Q$.

% feeding the pretrained LM with suitable and intuitive prompts can trigger it to generate relevant content
\textbf{Prompts Construction} Given the question $Q$ and the set of question-context pair samples $S$ selected, we use few-shot prompting to condition pretrained LMs on the samples. 
We use similar few-shot prompting technique for \textit{closed-book} QA as in ~\cite{brown2020language}, that considers multiple <question, answer> pairs. The template we used to construct prompts is: \texttt{Q:} ... \texttt{A:} .... 
Thus the constructed prompt $Prompt(Q)$ for a given question $Q$ becomes:
\begin{align*}
    Prompt(Q) =& \texttt{Q:} q_m \backslash n \texttt{A:} c_m \backslash n \ldots \\ &\texttt{Q:} q_1 \backslash n \texttt{A:} c_1 \backslash n \texttt{Q:} Q \backslash n
\end{align*}

We use '$\backslash n$' to separate the question, context and the samples. 
We investigated the order of samples to optimize the prompt and find that using the retrieved samples in reversed order of similarity yields better accuracies across all datasets. 

We now pass $Prompt(Q)$ through a pretrained LM to generate the context as follows: 

\begin{equation*}
    c_{gen} = \mathcal{LM}(Prompt(Q))
\end{equation*}

To generate a set of $k$ contexts, $\{c^1_{gen}, ..., c^k_{gen}\}$, we run this step $k$ times.

\subsection{Answer Prediction}
\label{sec:ans_pred}
In the answer prediction stage, we use the generated context $c_{gen}$ from the first stage along with the question $Q$.
%After we get the generated context for each question from the first stage, we further predict the answer conditioned on the context alongside with the question. 
Specifically, we prompt the same LM by few-shot examples
%<Context, Question, Answer> examples 
selected from the training set, $\mathcal{D}$, together with $c_{gen}$ and $Q$. 

\textbf{Sample Selection} 
Constrained by the maximum sequence length of the LM, we can feed the LM only a few $(c, q, a)$ samples. 
Thus, it could be difficult for the LM to learn how to predict the answer for the given question conditioned on the context, unless similar examples have been provided. 
For example, if we were asking the question \textit{'who is the current director of the us mint?}', the example that answering the question \textit{'who is the fbi director of the united states?'} from the provided context will be more helpful, than the example that is answering \textit{'how many episodes are there in `Dragon Ball Z'?'} from the given context. We therefore use the same criteria for answer prediction as has been used for context generation. 
We use the same set of samples as selected in the first stage as described in Equation~\ref{eqa:retriever_score} and denote as $S'=\{(q_1, c_1, a_1), \ldots, (q_m, c_m, a_m)\}$. %However, each $S'_i$ is composed of tuples $<Q_i, C_i, A_i>$, where $A_i$ is the corresponding answer for question $Q_i$ and given context $C_i$. 
%Thus, we use the same similarity based criteria to select the answer prediction prompt samples, i.e., we used the same set of samples selected from the first stage as described in Equation~\ref{eqa:retriever_score}, we denote as $S'=\{S'_1, S'_2,..., S'_m\}$, except that now $S'_m$ is composed of triples of $<Q_m, C_m, A_m>$ where $A_m$ is the corresponding answer for question $Q_m$ given context $C_m$. 

\textbf{Prompt Construction} We are prompting LMs with few-shot examples to predict answer for the question conditioned on the generated context. 
To equip the LM with this capability, we constructed intuitive prompts for the selected examples and feed them into the LM. 
Specifically, the template we used to construct answer prediction prompts is: \texttt{C:} ... \texttt{Q:} ... \texttt{A:} ... . 
Thus, the constructed prompt for a given question Q and the $i$-th generated context $c^i_{gen}$ is:
\vspace{-5pt}

\begin{align}
\begin{split}
    \textrm{Prompt}(c^i_{gen}, Q) =&
    \texttt{C:} c_m \backslash n 
    \texttt{Q:} q_m \backslash n
    \texttt{A:} a_m \backslash n\\
    & \ldots \\
    &\texttt{C:} c_1 \backslash n 
    \texttt{Q:} q_1 \backslash n 
     \texttt{A:} a_1 \backslash n \\
    &\texttt{C:} c^i_{gen} \backslash n
    \texttt{Q:} Q \backslash n 
    \label{equ:cgen_q}
\end{split}
\end{align}

We then feed $Prompt(c^i_{gen}, Q)$ into the pretrained LM to predict the answer:
\begin{equation}
    a_p^i = \mathcal{LM}(\textrm{Prompt}(c^i_{gen}, Q)))
    \label{equ:answer_prediciton}
\end{equation}
where we use $a^i_p$ to denote the $i$-th answer predicted by the LM. The $k$ generated contexts in $c_{gen}$ will yield a set of answers $A_p=\{a^1_p, ...,a^k_p\}$.

\subsection{Context Marginalization}
\label{sec:methods_margin}
The large pretrained LM can generate impressively fluent and relevant context given input, it also has a tendency to generate factually incorrect statements, ranging from subtle inaccuracies to wild hallucinations~\cite{shuster2021retrieval, krishna2021hurdles, su2022read}. 
Answers conditioned solely on hallucinated or erroneous statements are likely to be incorrect (Equation~\ref{equ:answer_prediciton}). Thus, we would like to remove the variability in the answer due to any particular generated context.

Ideally, we could marginalize over this unknown context by producing an answer for every possible context, weighting each answer by the probability of the context. Here we approximate this by generating a set of contexts, and selecting the final answer based on majority voting. Suppose there are $T$ unique answers $\{A^1_p,...,A^T_p\}$ from the $k$ predicted answer from Equation~
\ref{equ:answer_prediciton} where $T<=k$, then we select the $J$-th answer that receives the highest number of votes from the $T$ different answers via:
\begin{equation}
    % J = \argmax_{j\in \{1,2,...,T\}} \sum^k_{i=1}(\mathbbm{1}(a^i_p = A^j_p))
    J = \argmax_{j\in \{1,2,...,T\}}\sum^k_{i=1}(\mathbbm{1}(a^i_p = A^j_p))
    \label{equ:marginalization}
\end{equation}

as the final answer $A$.  As $k$ gets larger, the final answer $A$ will converge to the answer that would be produced marginalizing over all possible contexts.  We refer to this majority vote over multiple generated contexts as context marginalization.

\begin{table*}[!th]
\centering
\begin{adjustbox}{width=0.95\textwidth}
{
\begin{tabular}{lllccc}
\specialrule{.08em}{.1em}{.1em} 
\textbf{Model Type}  & \multicolumn{1}{c}{\textbf{Model}} & \textbf{Method} &\textbf{NQ} & \textbf{TQA} & \textbf{WQ}\\ 
\hline 
\multirow{3}{*}{Open-book} & RAG~\cite{lewis2020retrieval} & \textit{Finetuned} & 44.5  & 68.0    & \textbf{45.5}  \\
                            & Fusion-in-Decoder (large) ~\cite{izacard2021leveraging} & \textit{Finetuned}  & \textbf{51.4}  & 67.6  & -     \\
                              & $OB_{Google}^{PoE}$~\cite{lazaridou2022internet} & \textit{Few-shot} & 38.4  & - & - \\ 
                              \hline
\multirow{5}{*}{Closed-book}  & T5-11B ~\cite{roberts2020much} & \textit{Finetuned} & 32.6             & 42.3     & 37.2         \\
                              & T5-11B+SSM ~\cite{roberts2020much} & \textit{Finetuned}   & 34.8  & 51.0    & 40.8  \\
                            & BART-large, pre-finetuned on PAQ ~\cite{lewis2021paq} & \textit{Finetuned} & 32.7  & 33.2  & -  \\
                            %   & GPT-3-175B (Few-shot, paper) *~\cite{brown2020language}              & 29.9  & 71.2  & 41.5  \\
                              & LM-530B (API) & \textit{Few-shot}           & 23.0 & 55.3 & 23.6 \\ 
                              \cdashline{2-6}
                            % \cline[dashed]{2-6}
                              & \textbf{CGAP} (ours)   & \textit{Few-shot} & \underline{42.0} & \textbf{\underline{68.6}}  & \underline{41.8} \\ 
\specialrule{.08em}{.1em}{.1em} \\
\end{tabular}
}
\end{adjustbox}
\vspace{-10pt}
\caption{Exact Match (EM) score for \textbf{CGAP} (highest accuracy configurations) in comparison to recent state-of-the-art \textit{open-book} and \textit{closed-book} based systems. Highest score indicated in \textbf{bold}, highest \textit{closed-book} model
\underline{underlined}.}
\vspace{-5pt}
\label{tab:main_results}
\end{table*}

\section{Experimental Setup}

\subsection{Datasets}
\label{sec:datasets}
We evaluated our experiments on three open-domain QA benchmark datasets: Natural Questions (NQ)~\cite{kwiatkowski2019natural}, TriviaQA (TQA)~\cite{joshi2017triviaqa}, and WebQuestions 
(WQ)~\cite{berant2013semantic}, using the same data splits for train, validation and test as in~\citet{lee2019latent, izacard2021leveraging}.

NQ contains questions from Google search queries; TQA contains a collection of questions from trivia and quiz-league websites, and we use their unfiltered set; while questions of WQ were from Google Suggest API. 
For NQ and TQA, we use the processed data provided by~\citet{izacard2021leveraging}, in which each question-answer pair is accompanied by a 100-words Wikipedia passage containing the answer. 
For WQ, we retrieved the corresponding context passage for each question from 2019/08/01 Wikipedia dump, using the DPR-based retriever that is trained jointly on the union of knowledge-intensive training data in KILT benchmark~\cite{petroni2021kilt}.

%WE NEED TO MENTION ABOUT THE PRETRAINED MODELS as well.

\subsection{Baselines}
\label{sec:baselines}
We compare our \textbf{CGAP} framework with the following baseline methods for \textit{closed-book} QA.

\textbf{Standard Few-shot Prompting} 
We use the standard few-shot prompting technique
similar to GPT-3~\cite{brown2020language} in our evaluation on the \textit{closed-book} QA datasets as described in Section ~\ref{sec:datasets}. We consider this technique as the few-shot baseline in all our experiments. The baseline that is experimented using 530 billion (530B) parameterized LM is refferred as \textbf{LM-530B}.
%For fair comparison of this technqiue with our method, we experiment the same 530 billion parameterized LM (referred as \textbf{LM-530B}).
%adopted the standard few-shot prompting on GPT-3 for \textit{closed-book} QA, and evaluated on the same datasets as described in Section ~\ref{sec:datasets}.
%However, we were unable to reproduce the results reported in their paper when we query GPT-3 via API~\footnote{Results of querying OpenAI GPT-3 API using standard prompting are shown in Appendix~\ref{sec:appendix-gpt3}}. 
%Thus, we use their standard prompting format and experimented on a 530 billion parameterized LM as our baseline (referred as \textbf{LM-530B}), for fair comparison. 
%Greedy

% They use beam search with a beam width of 4 and a length penalty of 0.6 for answer prediction. (reffered as \textbf{GPT-3 (Few-shot, paper)})
% They first randomly draw $K$ (=64) question-answer pairs from the corresponding training set, and use 'Q: ' and 'A: ' respectively as prefix before each question and answer, to form the conditioning prompts.

\textbf{LM Fune-tuning}~\citet{roberts2020much} first proposed the \textit{closed-book} QA task for open domain QA, and they directly fine-tuned T5~\cite{raffel2019exploring} using the entire QA pairs in the training data, without access to any external knowledge corpus (referred as \textbf{T5-11B}). They also experimented with using ’Salient Span-Masking’ (SSM) to continue pretraining the T5 checkpoints before fine-tuning for QA (referred as \textbf{T5-11B+SSM}). ~\citet{lewis2021paq} pre-finetuned BART-large~\cite{lewis2020bart} on \textit{Probably Asked Questions} (PAQ), a very large resource of 65M automatically generated QA-pairs, then further finetuned the model on corresponding training data (referred as \textbf{BART-large, pre-finetuned on PAQ}).

% \boldmath
\textbf{Open-book Few-shot Prompting} ~\citet{lazaridou2022internet} used few-shot prompting for open domain QA task, but they generate the answer via conditioning on retrieved documents from Google Search API. (referred as \textbf{$OB^{PoE}
_{Google}$})
% They proposed to do answer ranking via Product-of-Experts(PoE) to combine together all probabilities for the final answer probability. 
% \unboldmath

\subsection{State-of-the-art Open-book QA Models}
We compare the state-of-the-art \textit{open-book} QA models with \textbf{CGAP}. \textbf{Fusion-in-Decoder} (FiD)~\cite{izacard2021leveraging} uses DPR~\cite{karpukhin2020dense} to retrieve 100 passages from Wikipedia. Then they encode each passage independently and combine all outputs from the T5 encoder before passing them to the T5 decoder to generate a final answer. \textbf{RAG}~\cite{rag} is an end-to-end retrieval-augmented generation model. 
% which back-propagates to the retriever’s input encoder, learning to adapt the input embedding to retrieve more relevant results.

% \begin{figure*}[!t]
% 	\begin{center}
% 		\includegraphics[width=0.95\textwidth]{figs/ablation-1-new.pdf}
% 		\caption{Ablation on context generation LM Size. Answer prediction LM size: 357M (\textit{left}), 1.3B(\textit{right}). The colored dash lines represent the standard prompting few-shot baselines.}
% 		\vspace{-5pt}
% 		\label{Fig:ablation-CG-357m-1.3b-ans}
% 	\end{center}
% \end{figure*}

\begin{figure*}[!t]
	\begin{center}
		\includegraphics[width=0.95\textwidth]{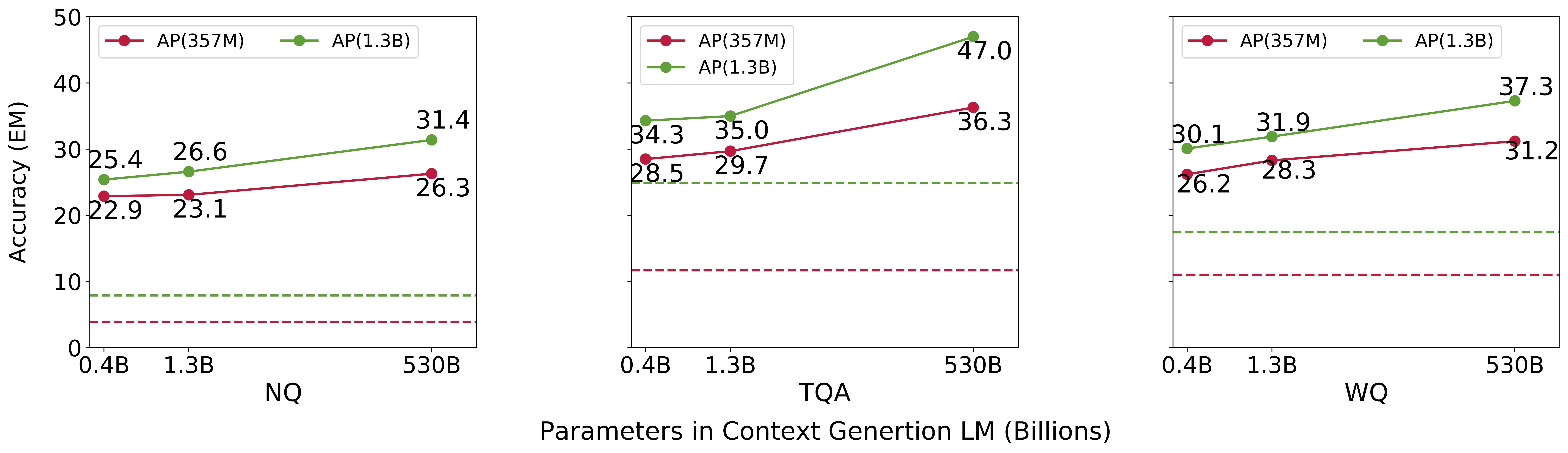}
		\caption{Ablation on context generation LM Size. The colored dash lines represent the standard prompting few-shot baselines.}
		\vspace{-5pt}
		\label{Fig:ablation-CG-357m-1.3b-ans}
	\end{center}
\end{figure*}
\subsection{Implementation Details}
To test how different model scales affect the performance of our approach, we experiment on a collection of GPT-style LMs, with 357 million (357m),
1.3 billion (1.3b), and 530 billion (530b)~\cite{smith2022using} parameters, at both context generator and answer prediction stage. 
We use top-$p$ sampling with a value of 0.9 to generate diversified contexts. However, to handle the deterministic generation (e.g. short answer),
we use greedy decoding at the answer prediction stage, similar to ~\cite{chowdhery2022palm, wang2022self}.

For the prompt configuration at both stages, we choose 10 samples, constrained by the maximum sequence length of the LMs. We use DPR checkpoint from Huggingface\footnote{\url{https://huggingface.co/facebook/dpr-ctx_encoder-multiset-base}} to select samples from the training set.

\subsection{Evaluation}
Following the standard evaluation procedures in previous work~\cite{rajpurkar2016squad,lee2019latent, izacard2021leveraging}, we use Exact Match (EM) as our answer accuracy evaluation metric, where each predicted answer is compared to the ground-truth after both are lowercased and stripped of articles, punctuation, and duplicate whitespace.

\section{Results and Ablation Studies}
In this section, we show our main results as well as ablations to further analyze the effectiveness of our approach.
% ompare our CCAP approach with baselines, and also state-of-the-art \textit{open-book} models for open-domain QA. 

\begin{table*}[t]
\centering
% \resizebox{.5\textwidth}{!}{
\begin{adjustbox}{totalheight=0.35\textheight-2\baselineskip,}
{
\begin{tabular}{ccclll}
\specialrule{.08em}{.1em}{.1em} 
\begin{tabular}[c]{@{}c@{}}\textbf{AP} \\ \textbf{LM Size}\end{tabular} &
  \begin{tabular}[c]{@{}c@{}}\textbf{CG} \\ \textbf{LM Size}\end{tabular} &
  \begin{tabular}[c]{@{}c@{}}\textbf{Margin-} \\\textbf{alization} \end{tabular} &
  \textbf{NQ} &
  \textbf{TQA} &
  \textbf{WQ} \\ \hline
\multirow{6}{*}{357M} & \multirow{2}{*}{357M} & \xmark & 22.9 & 28.5 & 26.2 \\
                      &  &  \checkmark   & 25.7 \textcolor{mypink2}{\small{(+2.8)}} & 33.4 \textcolor{mypink2} {\small{(+4.9)}} & 29.6 \textcolor{mypink2}{\small{(+3.4)}} \\ \cline{2-6} 
                      & \multirow{2}{*}{1.3B} & \xmark & 23.1 & 29.7 & 28.3  \\
                      &  &  \checkmark  & 26.1 \textcolor{mypink2}{\small{(+3.0)}} & 34.8 \textcolor{mypink2}{\small{(+5.1)}} & 31.3 \textcolor{mypink2}{\small{(+3.0)} } \\ \cline{2-6} 
                      & \multirow{2}{*}{530B} & \xmark & 26.3 & 36.3 & 31.2  \\
                      &  &  \checkmark   & 28.9 \textcolor{mypink2}{\small{(+2.7)}} & 45.7 \textcolor{mypink2}{\small{(+9.4)}} & 34.0 \textcolor{mypink2}{\small{(+2.8)}} \\ \hline
\multirow{2}{*}{530B} & \multirow{2}{*}{530B}  & \xmark & 29.5 & 56.3 & 28.3  \\
                      &  & \checkmark    & \textbf{42.0} \textcolor{mypink2}{\small{(+12.5)}} & \textbf{68.6} \textcolor{mypink2}{\small{(+12.4)}} & \textbf{41.8} \textcolor{mypink2}{\small{(+13.5)}} \\ 
\specialrule{.08em}{.1em}{.1em} 
\end{tabular}
}
\end{adjustbox}{}
\caption{Ablation on context marginalization. (AP and GP represent Answer Prediction and Context Generation, respectively.)}

\label{tab:ablation-2-margin}
\end{table*}

% \subsection{Comparison to Prior Work}
\subsection{Main Results}
\label{sec:main_results}
Table~\ref{tab:main_results} shows the EM score comparison between our CGAP-based method with existing \textit{closed-book} baseline approaches~\footnote{GPT-3 API shows different results than reported in the paper~\cite{brown2020language}. We therefore didn't compare to it. 
%We were unable to reproduce the results reported in paper~\cite{brown2020language} when we query GPT-3 via API, so we didn't compare to it. 
Detailed on these are shown in Section~\ref{sec:appendix-gpt3}}. 
We also compare with state-of-the-art \textit{open-book} models at the upper section of the table. 

% While GPT-3 (175B) few-shot method reported the highest answer accuracy on TQA~\cite{brown2020language}, we can not reproduce that score when we call the GPT-3 API by ourselves~\footnote{Results of querying openAI GPT-3 API using standard prompting are shown in Appendix}. 
As we can see, our CGAP based method outperforms other existing \textit{closed-book} methods by large margin, especially on NQ and TQA datasets. 
The CGAP also outperforms the standard few-shot prompting baseline LM-530B on all three datasets (at least by $13.3$ EM point). 

Furthermore, CGAP obtains highest score on TriviaQA. 
The scores are also very close to the state-of-the-art \textit{open-book} method RAG on NQ and WebQuestions, and only lose few points on NQ to FiD. While FiD uses 100 retrieved passages for answer prediction, CGAP rather only uses $8$ generated contexts for the approximate context marginalization.

\subsection{Ablation Studies}
We conducted a systematic ablation study to further investigate the contribution of the context generation model and the effect of context marginalization.

\subsubsection{Context Generation}
\label{sec: ablation_cg}
While previous work~\cite{roberts2020much, brown2020language} demonstrated that the scale of the model sizes improves the answer accuracy of \textit{closed-book} QA, there are also other findings showing that simply increasing the model size does not lead to substantive accuracy gains~\cite{RaeGopher2022}. Thus, we would like to investigate \textbf{how will the context generation LM affect the answer accuracy}. 

We experimented by varying the LM sizes for context generation, and fix the answer generation LM. We used context generation LM sizes of 357m, 1.3B and 530B, and answer generation LM with 357m and 1.3B parameters. 
We also compare with standard few-shot prompting which has no context generation.
 
We plot the results in Figure~\ref{Fig:ablation-CG-357m-1.3b-ans}. As we can see, there are huge accuracy gains from standard prompting, to CGAP method that has context generation. The accuracy increases by absolute $19.00$\% for NQ, $16.87$\% for TQA and $15.26$\% for WQ, when using 357M model for both standard prompting and CGAP approach.
%For example, when we use the same answer prediction LM (357m), the answer accuracy increases from 3.85\% to 22.85\% for NQ, 10.97\% to 26.23\% for WebQuesions, and 11.66\% to 28.53\% for TriviaQA respectively, when we use only a 357M LM for context generation in CGAP. 
The answer accuracy continues to increase when we increase the LM size for context generation. 
Furthermore, we notice that the slopes of the accuracy gain curve using larger answer prediction model is steeper than using smaller one on all three datasets. This suggests the use of larger answer prediction LM to fully exploit the knowledge in generated context.   

\begin{figure}[!th]
	\begin{center}	\includegraphics[width=0.36\textwidth]{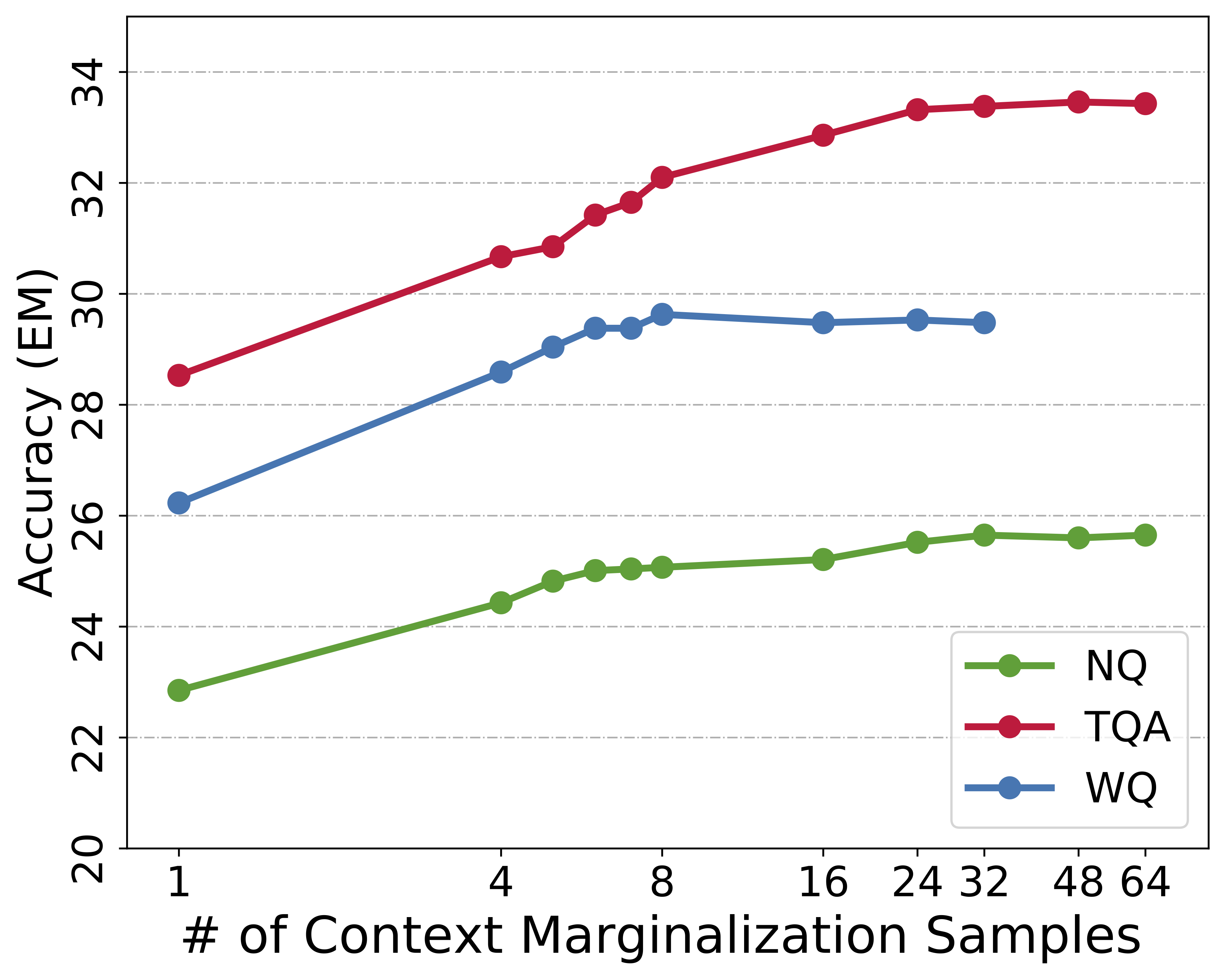}
		\caption{Ablation on $k$, the number of contexts for marginalization.}
		\vspace{-15pt}
		\label{Fig:ablation-margin-k}
	\end{center}
\end{figure}

\subsubsection{Context Marginalization}
\label{sec:margin}
Since there will be some hallucinated content or erroneous statements in the generated context, we approximate context marginalization by sampling multiple contexts and selecting the final answer based on majority voting, as introduces in Section~\ref{sec:methods_margin}. Here, we investigate the performance gains brought in by context marginalization, and also the accuracy curves with varied number of sampled contexts used in the approximate marginalization $k$.

In Table~\ref{tab:ablation-2-margin}, we show the accuracy comparisons w/ and w/o using marginalization (k=8), with different LM sizes. 
As we can see, \textbf{context marginalization improves the answer accuracy consistently on the three datasets}\footnote{We show a concrete example in Appendix~\ref{sec:appendix_example} Table~\ref{tab:appendix-gcap}}, under all settings. 
Notably, there is much larger performance gains using marginalization when we scale up the model sizes to 530 billion parameters (i.e. increase EM score by 12.8\% averaged on three datasets). 

The larger the number of context samples $k$, the more accurately the majority vote reflects the true marginalization over all possible contexts.  Therefore, we perform further ablation by changing the value $k$ for 357M LM for both context generation and answer prediction. 
%We use 357M LM , but vary the number of context samples $k$ used for marginalization. 
We plot the accuracy curves in Figure~\ref{Fig:ablation-margin-k}. We see that there are accuracy improvements when we use more context samples.  As expected and curves plateau for larger values of $k$ as the approximation approaches the true marginalization over all possible contexts.

\section{Analysis}
Considering that it is the first time leveraging context generated by large pretrained LMs for ODQA, we also conducted further analysis. 

We compare generated context with retrieved context in the two-stage, few-shot prompting based CBQA framework. It is a dominant paradigm to use retrieved context from external corpus together with the question for answer prediction for \textit{open-book} QA~\cite{chen2017reading, lewis2020retrieval,izacard2021leveraging, lazaridou2022internet}. 

\subsection{Retrieved vs. Generated Context}
% Therefore, we investigate using the retrieved context in the few-shot prompting setting for CBQA, and compare its performances with using generated context. 

In CBQA setting, we are not allowed to retrieve context from external knowledge sources. 
However, we can retrieve the contexts from the training dataset based on their relevance to the given question. 
We use $c_r=\{c^1_r, c^2_r,..., c^m_r\}$ to represent the top-$m$ relevant context for question $Q$. 
It can be obtained via Equation~\ref{eqa:retriever_score}.
%, where $c^j_r$ is the context of the $j$-th samples $S_j$ in $S=\{S_1, S_2,... S_m\}$. 

Let the top-1 retrieved context be $c^{\text{top-1}}_r$ for question $Q$.
We use $c^{\text{top-1}}_r$ to compare with the generated context, $c_{gen}$. 
We use the same top-$m$ prompts $S'$ for answer prediction as introduced in Section~\ref{sec:ans_pred}. 
The answer $a^r_p$ for the $c^{\text{top-1}}_r$ will be:
\begin{equation}
    a^r_p = \mathcal{LM}(Prompt(c^{\text{top-1}}_r, Q)))
    \label{equ:c1}
\end{equation}

where $Prompt(c^{\text{top-1}}_r,Q)$ can be obtained via Equation~\ref{equ:cgen_q}. 

% Equation~\ref{equ:answer_prediciton} is used to calculated the answer $A_p$ using generated context $C_{gen}$ in CGAP approach. 

% We experimented using different context generation and answer prediction LM sizes, to compare $C_{gen}$ with $C^{\text{top-1}}_r$ thoroughly. 
% We show that the accuracy gains from $C_{gen}$ are largely correlated with LM size and the the model size for answer prediction (AP) in XX.

The comparison between $c^{\text{top-1}}_r$ and $c_{gen}$ is shown in Table~\ref{tab:further_analysis_c1}. From the upper part of the table, we see that using $c^{\text{top-1}}_r$ gives slightly higher EM score than using $c_{gen}$ generated by 357M and 1.3B LMs. However, $c_{gen}$ gives higher EM scores than $c^{\text{top-1}}_r$ on all three datasets when we scale up the context generation LM size to 530B. This suggests the use of large pretrained LM for a better generated context.

\begin{table}[htbp]
\centering
\begin{adjustbox}{width=0.46\textwidth}
{
\begin{tabular}{lc|ccc}
\specialrule{.08em}{.1em}{.1em} 
\multicolumn{1}{c}{\textbf{AP}}          & \textbf{Context}          & \textbf{NQ}    & \multicolumn{1}{l}{\textbf{TQA}} & \multicolumn{1}{l}{\textbf{WQ}} \\ \hline
\multirow{4}{*}{357M}  & $c^{\text{top-1}}_r$  & 25.1 & 32.2 & 28.3 \\
                      & $c_{gen}$ \small{(357M LM)}  & 22.9  & 28.5  & 26.2 \\
                      & $c_{gen}$ \small{(1.3B LM)}  & 23.1  & 29.7  & 28.3 \\
                      & $c_{gen}$ \small{(530B LM)}  & \textbf{26.3}  & \textbf{36.3} & \textbf{31.2}  \\ \hline
% \multirow{4}{*}{1.3B} & $C^1_r$   & 28.4 & 41.0                 & 34.5 \\
%                       & $C_{gen}$ \small{(357M LM)}  & 25.4 & 34.3 & 30.1\\
%                       & $C_{gen}$ \small{(1.3B LM)} & 26.6 & 35.0  & 31.9 \\
%                       & $C_{gen}$ \small{(530B LM)} & \textbf{31.4} & \textbf{47.0} & \textbf{37.3} \\ \hline
\multicolumn{1}{c}{\multirow{2}{*}{530B}} & $c^{\text{top-1}}_r$  & \textbf{30.8} & \textbf{58.1}  & \textbf{29.5} \\
\multicolumn{1}{c}{}  & $c_{gen}$\small{(530B LM)}  & 29.5 & 56.3  & 28.3 \\ 
\specialrule{.08em}{.1em}{.1em} 
\end{tabular}
}
\end{adjustbox}
\caption{Comparison of using retrieved top-1 context $c^{\text{top-1}}_r$, with few-shot generated context $c_{gen}$ on \textit{closed-book} QA task.}
\vspace{-10pt}
\label{tab:further_analysis_c1}
\end{table}
\vspace{-10pt}
\subsection{Multiple Retrievals vs. Context Marginalization}
We notice that in Table~\ref{tab:further_analysis_c1}, $c^{\text{top-1}}_r$ performs slightly better than $c_{gen}$ when using 530B LM for answer prediction. 
We argue that this might be caused by the hallucination in $c_{gen}$. 
While we have shown in Section~\ref{sec:margin} that context marginalization could mitigate the problem and improve answer accuracy, we further facilitate $c_{gen}(530B)$ with context marginalization and compare with retrieved context. 

For fair comparison, we perform majority voting using the top-$k$ retrieved context $c_r$, since ~\citet{karpukhin2020dense} showed that the quality of the retrieved documents will also affect the final answer accuracy. 
Specifically, we replace $c^{\text{top-1}}_r$ with each retrieved context $c^i_r$ in Equation~\ref{equ:c1} to predict answer $a^{r(i)}_p$ ($i=1,...,k$), and use Equation~\ref{equ:marginalization} to select the most frequent answer as the final answer.
 
Furthermore, we replace $c^{\text{top-1}}_r$ with golden context $c_{golden}$ in Equation~\ref{equ:c1}. 
This will be the upper-bound of using retrieved/generated context in the two-stage, few-shot prompting CBQA task.

We show the results\footnote{Concrete comparison examples are shown in Appendix~\ref{sec:appendix_example} Table~\ref{tab:example-NQ}, Table~\ref{tab:example-TQA} and Table~\ref{tab:example-WQ}.} in Table~\ref{tab:further_analysis_margin}. 
As we can see, using marginalization over $c_{gen}$ consistently outperforms $c^{\text{top-1}}_r$, and also better than majority voting over multiple retrieved contexts $c_r$ for answer prediction, on all different model size combinations on three datasets. 
Notably, marginalization over $c_{gen}$ yields higher EM score than using $c_{golden}$ when using 530B LM for answer prediction. 
% The possible reason for this might be for some $C_{golden}$ that can not lead to a correct answer, the $\{c^1_{gen},..., c^k_{gen}\}$ generated by the 530B model can suprisingly .  

\begin{table}[!h]
\centering
\begin{adjustbox}{width=0.46\textwidth}{
\begin{tabular}{cl|ccc}
\hline
\textbf{AP} & \multicolumn{1}{c}{\textbf{Context}} & \textbf{NQ} & \textbf{TQA} & \textbf{WQ} \\ \hline
\multirow{5}{*}{357M} & $c_{golden}$ & \textbf{31.6} & \textbf{41.5} & 33.4 \\
 & $c^{\text{top-1}}_r$ & 25.1 & 32.2 & 28.3 \\
 & ($c^1_r$,...,$c^k_r$) & 24.6 & 32.4 & 27.8 \\
& $c_{gen}$  & 26.3 & 36.3 & 31.2 \\
 & ($c^1_{gen}$,..., $c^k_{gen}$) & 28.9 & 45.7 & \textbf{34.0} \\ \hline
%  &  &  &  &  \\
\multirow{5}{*}{1.3B} & $c_{golden}$ & \textbf{35.1} & 51.4 & 38.1 \\
 & $c^{\text{top-1}}_r$ & 28.4 & 41.0 & 34.5 \\
 & ($c^1_r$,...,$c^k_r$) & 28.1 & 42.9 & 34.5 \\
  & $c_{gen}$   & 31.4 & 47.0 & 37.3 \\
 & ($c^1_{gen}$,..., $c^k_{gen}$) & 33.5 & \textbf{55.5} & \textbf{41.5} \\ \hline
%  &  &  &  &  \\
\multirow{5}{*}{530B} & $c_{golden}$ & 36.0 & 61.3 & 30.2 \\
 & $c^{\text{top-1}}_r$ & 30.8 & 58.1 & 29.5 \\
 & ($c^1_r$,...,$c^k_r$) & 29.5 & 56.3 & 28.3 \\
 & $c_{gen}$   & 23.0 & 55.3 & 23.6 \\
 & ($c^1_{gen}$,..., $c^k_{gen}$) & \textbf{42.0} & \textbf{68.6} & \textbf{41.8} \\ 
 \specialrule{.08em}{.1em}{.1em} 
\end{tabular}
}
\end{adjustbox}
\caption{Comparison of using context marginalization ($c^1_{gen}$,..., $c^k_{gen}$), multiple retrievals ($c^1_r$,...,$c^k_r$), and golden context $c_{golden}$ on \textit{closed-book} QA task.}
\vspace{-10pt}
\label{tab:further_analysis_margin}
\end{table}

\section{Summary}
We propose a simple yet effective framework named \textbf{CGAP} to extract succinct answers from long-form answers. CGAP performs \textbf{C}ontext \textbf{G}eneration followed by  \textbf{A}nswer \textbf{P}rediction via two-stage prompting using large pretrained LMs. 
It does not rely on external knowledge sources, and does not need finetuning or add extra learnable parameters.
To the best of our knowledge, we are the first to leverage generated context from large pretrained LMs for succinct answer extraction.
Experimental results on three open-domain QA benchmarks show that our method significantly outperforms previous \textit{closed-book} QA methods and is par with \textit{open-book} methods. We demonstrate our method up to 530B parameter models and showcase that larger models boost the accuracy by huge margins.

\chapter{Conclusion}

In this thesis, we investigated the relevance, faithfulness, and succinctness aspects of Long Form Question Answering (LFQA). LFQA aims to generate an in-depth, paragraph-length answer for a given question, to help bridge the gap between real scenarios and the existing open-domain QA models which can only extract short-span answers.

We are among the first to research the LFQA task. We pioneered the research direction to improve the answer quality in terms of 1) query-relevance, 2) answer faithfulness, and 3) answer succinctness.

We investigated the core challenges to high answer quality in the LFQA task, in terms of the three aspects. Specifically,

\begin{itemize}
    \item Since traditional Seq2Seq models are not good at handling long and multiple documents as input, we propose to use a coarse-to-fine method to extract the document-level and sentence-level query-relevant information hierarchically, and incorporate them into the generation process, for better answer relevance. We further introduce QFS-BART, a model that incorporates the explicit answer relevance attention of the source documents into the Seq2Seq model’s encoder-decoder attention module, to further enhance the query-relevance.

    \item As multiple relevant documents are needed for long-form answer generation:  they contain a considerable amount of redundant, complementary, or contradictory information; and the longer input also challenges the pre-trained LM which has limited input length. So we propose an end-to-end framework named RBG (read before generate) to do dynamic global salient information prediction from multiple source documents in parallel and fusion-in-decoder, and augment the generation model with the salient information, to improve and generate fact-aware answer that is more faithful to the source documents.
    
    \item 
    The generated answer also should be succinct, other than providing relevant and factual information. While no prior investigation has been conducted, we propose to leverage the generated long-form answers as a context to further extract succinct, short-phrase answers. The proposed CGAP method first generates a long-form answer leveraging the amount of parameterized knowledge stored in pre-trained language models ~\cite{raffel2020exploring, brown2020language, ye2020studying}, and extracts a short-phrase span answer from the generated long-form answer without access to any external knowledge sources. 

\end{itemize}

% We obtained state-of-the-art results on large-scale LFQA datasets, demonstrating the effectiveness of our proposed method in comparison with strong baselines, on automatic and human evaluation metrics. The proposed RBG method also topped the only public leaderboard~\footnote{\tt \small https://evalai.cloudcv.org/web/challenges/challenge-page/689/leaderboard/1908} on the LFQA task! We are also the first to build an LFQA system for COVID-19. The CAiRE-COVID system has won one of the ten tasks in the Kaggle COVID-19 Open Research Dataset Challenge\footnote{\tt \small  https://www.kaggle.com/allen-institute-for-ai/CORD-19-research-challenge}, by generating relevant and fluent answers to salient COVID-19 related questions, judged by medical experts. 

The effectiveness of the coarse-to-fine method and the QFS-BART model to improve query-relevance has been proved by obtaining state-of-the-art results on large-scale datasets by automatic evaluation. The proposed RGB method to improve answer faithfulness obtains state-of-the-art results on large-scale LFQA datasets ELI5 and MS MARCO compared to strong baselines such as FiD and RAG, by both automatic and human evaluation metrics. We experimented on the extractive open-domain QA task, to generate succinct answers with the CGAP method. CGAP significantly outperforms previous \textit{closed-book} QA methods (e.g., exact matching 68.6\% vs. 55.3\%) and is on par with \textit{open-book} methods that exploit external knowledge sources (e.g., 68.6\% vs. 68.0\%), on three QA benchmarks.

The proposed RBG method also topped the only public leaderboard on the LFQA task, outperforming the previous state-of-the-art method by Google! We leverage the coarse-to-fine method to build the first LFQA system for COVID-19. The CAiRE-COVID system has won one of the ten tasks in the Kaggle COVID-19 Open Research Dataset Challenge, by generating relevant and fluent answers to salient COVID-19-related questions, judged by medical experts.

In future work, we expect to develop LFQA systems that can provide more query-relevant and fact-aware answers, and are also succinct enough. In fact, in the three chapters of this thesis, we open new exciting research directions for LFQA system: 1) there are still possible ways worth exploring to build a more query-relevant LFQA system. In Chapter 3, we try to improve query-relevance via the QFS task, while the quality of the relevant document retrieved from external corpus also affects the generation model. In future works, We would like to expand the external knowledge sources such as the knowledge bases (KBs), visual materials, etc., so that the coverage of relevant knowledge from the retriever will be increased. 2) building a fact-aware LFQA system is still in its infancy. In Chapter 4, we leverage a machine reading comprehension model to highlight the salient information in the retrieved documents and incorporate the global-wise information into the generation. In the future, more sophisticated retrieval techniques such as iterative retrieval should be explored on the task. 3) generating succinct answers by an LFQA system. In Chapter 5, we explored generating succinct
answers for those simple question types such as the who/when/which/where types from the long-form answers. In the future, generating succinct answers from general question types such as the why/how should be explored.

\newpage
\addcontentsline{toc}{chapter}{Reference}
\bibliographystyle{IEEEtranN}
\bibliography{reference}

% Generated by IEEEtranN.bst, version: 1.14 (2015/08/26)
\begin{thebibliography}{144}
\providecommand{\natexlab}[1]{#1}
\providecommand{\url}[1]{#1}
\csname url@samestyle\endcsname
\providecommand{\newblock}{\relax}
\providecommand{\bibinfo}[2]{#2}
\providecommand{\BIBentrySTDinterwordspacing}{\spaceskip=0pt\relax}
\providecommand{\BIBentryALTinterwordstretchfactor}{4}
\providecommand{\BIBentryALTinterwordspacing}{\spaceskip=\fontdimen2\font plus
\BIBentryALTinterwordstretchfactor\fontdimen3\font minus
  \fontdimen4\font\relax}
\providecommand{\BIBforeignlanguage}[2]{{%
\expandafter\ifx\csname l@#1\endcsname\relax
\typeout{** WARNING: IEEEtranN.bst: No hyphenation pattern has been}%
\typeout{** loaded for the language `#1'. Using the pattern for}%
\typeout{** the default language instead.}%
\else
\language=\csname l@#1\endcsname
\fi
#2}}
\providecommand{\BIBdecl}{\relax}
\BIBdecl

\bibitem[Ferrucci et~al.(2010)Ferrucci, Brown, Chu-Carroll, Fan, Gondek,
  Kalyanpur, Lally, Murdock, Nyberg, Prager, et~al.]{ferrucci2010building}
D.~Ferrucci, E.~Brown, J.~Chu-Carroll, J.~Fan, D.~Gondek, A.~A. Kalyanpur,
  A.~Lally, J.~W. Murdock, E.~Nyberg, J.~Prager \emph{et~al.}, ``Building
  watson: An overview of the deepqa project,'' \emph{AI magazine}, vol.~31,
  no.~3, pp. 59--79, 2010.

\bibitem[Chen et~al.(2017)Chen, Fisch, Weston, and Bordes]{chen2017reading}
D.~Chen, A.~Fisch, J.~Weston, and A.~Bordes, ``Reading wikipedia to answer
  open-domain questions,'' in \emph{Proceedings of the 55th Annual Meeting of
  the Association for Computational Linguistics (Volume 1: Long Papers)}, 2017,
  pp. 1870--1879.

\bibitem[Nguyen et~al.(2016{\natexlab{a}})Nguyen, Rosenberg, Song, Gao, Tiwary,
  Majumder, and Deng]{nguyen2016ms}
T.~Nguyen, M.~Rosenberg, X.~Song, J.~Gao, S.~Tiwary, R.~Majumder, and L.~Deng,
  ``Ms marco: A human generated machine reading comprehension dataset,'' in
  \emph{CoCo@ NIPS}, 2016.

\bibitem[Fan et~al.(2019)Fan, Jernite, Perez, Grangier, Weston, and
  Auli]{fan2019eli5}
A.~Fan, Y.~Jernite, E.~Perez, D.~Grangier, J.~Weston, and M.~Auli, ``Eli5: Long
  form question answering,'' in \emph{Proceedings of the 57th Annual Meeting of
  the Association for Computational Linguistics}, 2019, pp. 3558--3567.

\bibitem[Yang et~al.(2018{\natexlab{a}})Yang, Qi, Zhang, Bengio, Cohen,
  Salakhutdinov, and Manning]{yang2018hotpotqa}
Z.~Yang, P.~Qi, S.~Zhang, Y.~Bengio, W.~Cohen, R.~Salakhutdinov, and C.~D.
  Manning, ``Hotpotqa: A dataset for diverse, explainable multi-hop question
  answering,'' in \emph{Proceedings of the 2018 Conference on Empirical Methods
  in Natural Language Processing}, 2018, pp. 2369--2380.

\bibitem[Raffel et~al.(2019{\natexlab{a}})Raffel, Shazeer, Roberts, Lee,
  Narang, Matena, Zhou, Li, and Liu]{2019t5}
C.~Raffel, N.~Shazeer, A.~Roberts, K.~Lee, S.~Narang, M.~Matena, Y.~Zhou,
  W.~Li, and P.~J. Liu, ``Exploring the limits of transfer learning with a
  unified text-to-text transformer,'' \emph{arXiv e-prints}, 2019.

\bibitem[Nguyen et~al.(2016{\natexlab{b}})Nguyen, Rosenberg, Song, Gao, Tiwary,
  Majumder, and Deng]{DBLP:conf/nips/NguyenRSGTMD16}
\BIBentryALTinterwordspacing
T.~Nguyen, M.~Rosenberg, X.~Song, J.~Gao, S.~Tiwary, R.~Majumder, and L.~Deng,
  ``{MS} {MARCO:} {A} human generated machine reading comprehension dataset,''
  in \emph{Proceedings of the Workshop on Cognitive Computation: Integrating
  neural and symbolic approaches 2016 co-located with the 30th Annual
  Conference on Neural Information Processing Systems {(NIPS} 2016), Barcelona,
  Spain, December 9, 2016}, ser. {CEUR} Workshop Proceedings, T.~R. Besold,
  A.~Bordes, A.~S. d'Avila Garcez, and G.~Wayne, Eds., vol. 1773.\hskip 1em
  plus 0.5em minus 0.4em\relax CEUR-WS.org, 2016. [Online]. Available:
  \url{http://ceur-ws.org/Vol-1773/CoCoNIPS\_2016\_paper9.pdf}
\BIBentrySTDinterwordspacing

\bibitem[Tang et~al.(2020)Tang, Nogueira, Zhang, Gupta, Cam, Cho, and
  Lin]{tang2020rapidly}
R.~Tang, R.~Nogueira, E.~Zhang, N.~Gupta, P.~Cam, K.~Cho, and J.~Lin, ``Rapidly
  bootstrapping a question answering dataset for covid-19,'' \emph{arXiv
  preprint arXiv:2004.11339}, 2020.

\bibitem[Petroni et~al.(2021)Petroni, Piktus, Fan, Lewis, Yazdani, De~Cao,
  Thorne, Jernite, Karpukhin, Maillard, et~al.]{petroni2021kilt}
F.~Petroni, A.~Piktus, A.~Fan, P.~Lewis, M.~Yazdani, N.~De~Cao, J.~Thorne,
  Y.~Jernite, V.~Karpukhin, J.~Maillard \emph{et~al.}, ``Kilt: a benchmark for
  knowledge intensive language tasks,'' in \emph{Proceedings of the 2021
  Conference of the North American Chapter of the Association for Computational
  Linguistics: Human Language Technologies}, 2021, pp. 2523--2544.

\bibitem[Krishna et~al.(2021)Krishna, Roy, and Iyyer]{krishna2021hurdles}
K.~Krishna, A.~Roy, and M.~Iyyer, ``Hurdles to progress in long-form question
  answering,'' in \emph{Proceedings of the 2021 Conference of the North
  American Chapter of the Association for Computational Linguistics: Human
  Language Technologies}, 2021, pp. 4940--4957.

\bibitem[Kwiatkowski et~al.(2019{\natexlab{a}})Kwiatkowski, Palomaki, Redfield,
  Collins, Parikh, Alberti, Epstein, Polosukhin, Devlin, Lee,
  et~al.]{kwiatkowski2019natural}
T.~Kwiatkowski, J.~Palomaki, O.~Redfield, M.~Collins, A.~Parikh, C.~Alberti,
  D.~Epstein, I.~Polosukhin, J.~Devlin, K.~Lee \emph{et~al.}, ``Natural
  questions: a benchmark for question answering research,'' \emph{Transactions
  of the Association for Computational Linguistics}, vol.~7, pp. 453--466,
  2019.

\bibitem[Su et~al.(2019{\natexlab{a}})Su, Xu, Winata, Xu, Kim, Liu, and
  Fung]{su2019generalizing}
D.~Su, Y.~Xu, G.~I. Winata, P.~Xu, H.~Kim, Z.~Liu, and P.~Fung, ``Generalizing
  question answering system with pre-trained language model fine-tuning,'' in
  \emph{Proceedings of the 2nd Workshop on Machine Reading for Question
  Answering}, 2019, pp. 203--211.

\bibitem[Fleiss(1971)]{fleiss_kappa}
J.~L. Fleiss, ``Measuring nominal scale agreement among many raters.''
  \emph{Psychological bulletin}, p. 378, 1971.

\bibitem[Landis and Koch(1977)]{landis1977measurement}
J.~R. Landis and G.~G. Koch, ``The measurement of observer agreement for
  categorical data,'' \emph{biometrics}, 1977.

\bibitem[Joshi et~al.(2017{\natexlab{a}})Joshi, Choi, Weld, and
  Zettlemoyer]{joshi2017triviaqa}
M.~Joshi, E.~Choi, D.~S. Weld, and L.~Zettlemoyer, ``Triviaqa: A large scale
  distantly supervised challenge dataset for reading comprehension,''
  \emph{arXiv preprint arXiv:1705.03551}, 2017.

\bibitem[Berant et~al.(2013)Berant, Chou, Frostig, and
  Liang]{berant2013semantic}
J.~Berant, A.~Chou, R.~Frostig, and P.~Liang, ``Semantic parsing on freebase
  from question-answer pairs,'' in \emph{Proceedings of the 2013 conference on
  empirical methods in natural language processing}, 2013, pp. 1533--1544.

\bibitem[Simmons et~al.(1964)Simmons, Klein, and
  McConlogue]{simmons1964indexing}
R.~F. Simmons, S.~Klein, and K.~McConlogue, ``Indexing and dependency logic for
  answering english questions,'' \emph{American Documentation}, vol.~15, no.~3,
  pp. 196--204, 1964.

\bibitem[Kupiec(1993)]{kupiec1993murax}
J.~Kupiec, ``Murax: A robust linguistic approach for question answering using
  an on-line encyclopedia,'' in \emph{Proceedings of the 16th annual
  international ACM SIGIR conference on Research and development in information
  retrieval}, 1993, pp. 181--190.

\bibitem[Voorhees et~al.(1999)]{voorhees1999trec}
E.~M. Voorhees \emph{et~al.}, ``The trec-8 question answering track report,''
  in \emph{Trec}, vol.~99.\hskip 1em plus 0.5em minus 0.4em\relax Citeseer,
  1999, pp. 77--82.

\bibitem[Moldovan et~al.(2000)Moldovan, Harabagiu, Pasca, Mihalcea, Girju,
  Goodrum, and Rus]{moldovan2000structure}
D.~Moldovan, S.~Harabagiu, M.~Pasca, R.~Mihalcea, R.~Girju, R.~Goodrum, and
  V.~Rus, ``The structure and performance of an open-domain question answering
  system,'' in \emph{Proceedings of the 38th annual meeting of the Association
  for Computational Linguistics}, 2000, pp. 563--570.

\bibitem[Harabagiu et~al.(2000)Harabagiu, Pasca, and
  Maiorano]{harabagiu2000experiments}
S.~Harabagiu, M.~Pasca, and S.~J. Maiorano, ``Experiments with open-domain
  textual question answering,'' in \emph{COLING 2000 Volume 1: The 18th
  International Conference on Computational Linguistics}, 2000.

\bibitem[Brill et~al.(2002)Brill, Dumais, and Banko]{brill2002analysis}
E.~Brill, S.~Dumais, and M.~Banko, ``An analysis of the askmsr
  question-answering system,'' in \emph{Proceedings of the 2002 Conference on
  Empirical Methods in Natural Language Processing (EMNLP 2002)}, 2002, pp.
  257--264.

\bibitem[Baudi{\v{s}}(2015)]{baudivs2015yodaqa}
P.~Baudi{\v{s}}, ``Yodaqa: a modular question answering system pipeline,'' in
  \emph{POSTER 2015-19th International Student Conference on Electrical
  Engineering}, 2015, pp. 1156--1165.

\bibitem[Lin and Katz(2003)]{lin2003question}
J.~Lin and B.~Katz, ``Question answering techniques for the world wide web,''
  \emph{EACL-2003 Tutorial}, 2003.

\bibitem[Gliozzo et~al.(2012)Gliozzo, Kalyanpur, and Fan]{gliozzo2012natural}
A.~M. Gliozzo, A.~Kalyanpur, and J.~Fan, ``Natural language processing in
  watson,'' in \emph{Proceedings of the 2012 Conference of the North American
  Chapter of the Association for Computational Linguistics: Human Language
  Technologies: Tutorials}, 2012, pp. 1--3.

\bibitem[Yih and Ma(2016)]{yih2016question}
W.-t. Yih and H.~Ma, ``Question answering with knowledge base, web and
  beyond,'' in \emph{Proceedings of the 39th International ACM SIGIR conference
  on Research and Development in Information Retrieval}, 2016, pp. 1219--1221.

\bibitem[Sachan et~al.(2018)Sachan, Seo, Hajishirzi, and
  Xing]{sachan2018standardized}
M.~Sachan, M.~Seo, H.~Hajishirzi, and E.~Xing, ``Standardized tests as
  benchmarks for artificial intelligence,'' in \emph{Proceedings of the 2018
  Conference on Empirical Methods in Natural Language Processing: Tutorial
  Abstracts}, 2018.

\bibitem[Izacard and Grave(2021)]{izacard2021leveraging}
G.~Izacard and {\'E}.~Grave, ``Leveraging passage retrieval with generative
  models for open domain question answering,'' in \emph{Proceedings of the 16th
  Conference of the European Chapter of the Association for Computational
  Linguistics: Main Volume}, 2021, pp. 874--880.

\bibitem[Lewis et~al.(2021)Lewis, Wu, Liu, Minervini, K{\"u}ttler, Piktus,
  Stenetorp, and Riedel]{lewis2021paq}
P.~Lewis, Y.~Wu, L.~Liu, P.~Minervini, H.~K{\"u}ttler, A.~Piktus, P.~Stenetorp,
  and S.~Riedel, ``Paq: 65 million probably-asked questions and what you can do
  with them,'' \emph{arXiv preprint arXiv:2102.07033}, 2021.

\bibitem[Metzler et~al.(2021)Metzler, Tay, Bahri, and
  Najork]{metzler2021rethinking}
D.~Metzler, Y.~Tay, D.~Bahri, and M.~Najork, ``Rethinking search: Making
  experts out of dilettantes,'' \emph{arXiv preprint arXiv:2105.02274}, 2021.

\bibitem[Guu et~al.(2020)Guu, Lee, Tung, Pasupat, and Chang]{guu2020realm}
K.~Guu, K.~Lee, Z.~Tung, P.~Pasupat, and M.-W. Chang, ``Realm:
  Retrieval-augmented language model pre-training,'' \emph{arXiv preprint
  arXiv:2002.08909}, 2020.

\bibitem[Karpukhin et~al.(2020)Karpukhin, Oguz, Min, Lewis, Wu, Edunov, Chen,
  and Yih]{karpukhin2020dense}
V.~Karpukhin, B.~Oguz, S.~Min, P.~Lewis, L.~Wu, S.~Edunov, D.~Chen, and W.-t.
  Yih, ``Dense passage retrieval for open-domain question answering,'' in
  \emph{Proceedings of the 2020 Conference on Empirical Methods in Natural
  Language Processing (EMNLP)}, 2020, pp. 6769--6781.

\bibitem[Lee et~al.(2019)Lee, Chang, and Toutanova]{lee2019latent}
K.~Lee, M.-W. Chang, and K.~Toutanova, ``Latent retrieval for weakly supervised
  open domain question answering,'' in \emph{Proceedings of the 57th Annual
  Meeting of the Association for Computational Linguistics}, 2019, pp.
  6086--6096.

\bibitem[Lewis et~al.(2020{\natexlab{a}})Lewis, Liu, Goyal, Ghazvininejad,
  Mohamed, Levy, Stoyanov, and Zettlemoyer]{lewis2020bart}
M.~Lewis, Y.~Liu, N.~Goyal, M.~Ghazvininejad, A.~Mohamed, O.~Levy, V.~Stoyanov,
  and L.~Zettlemoyer, ``Bart: Denoising sequence-to-sequence pre-training for
  natural language generation, translation, and comprehension,'' in
  \emph{Proceedings of the 58th Annual Meeting of the Association for
  Computational Linguistics}, 2020, pp. 7871--7880.

\bibitem[Raffel et~al.(2020)Raffel, Shazeer, Roberts, Lee, Narang, Matena,
  Zhou, Li, and Liu]{raffel2020exploring}
C.~Raffel, N.~Shazeer, A.~Roberts, K.~Lee, S.~Narang, M.~Matena, Y.~Zhou,
  W.~Li, and P.~J. Liu, ``Exploring the limits of transfer learning with a
  unified text-to-text transformer,'' \emph{Journal of Machine Learning
  Research}, vol.~21, pp. 1--67, 2020.

\bibitem[Su et~al.(2021)Su, Yu, and Fung]{su2021improve}
D.~Su, T.~Yu, and P.~Fung, ``Improve query focused abstractive summarization by
  incorporating answer relevance,'' in \emph{Findings of the Association for
  Computational Linguistics: ACL-IJCNLP 2021}, 2021, pp. 3124--3131.

\bibitem[Yu et~al.(2020)Yu, Su, Dai, and Fung]{yu2020dimsum}
T.~Yu, D.~Su, W.~Dai, and P.~Fung, ``Dimsum@ laysumm 20,'' in \emph{Proceedings
  of the First Workshop on Scholarly Document Processing}, 2020, pp. 303--309.

\bibitem[Su et~al.(2022{\natexlab{a}})Su, Xu, and Fung]{su2022qa4qg}
D.~Su, P.~Xu, and P.~Fung, ``Qa4qg: Using question answering to constrain
  multi-hop question generation,'' in \emph{ICASSP 2022-2022 IEEE International
  Conference on Acoustics, Speech and Signal Processing (ICASSP)}.\hskip 1em
  plus 0.5em minus 0.4em\relax IEEE, 2022, pp. 8232--8236.

\bibitem[Su et~al.(2020{\natexlab{a}})Su, Xu, Dai, Ji, Yu, and
  Fung]{su2020multi}
D.~Su, Y.~Xu, W.~Dai, Z.~Ji, T.~Yu, and P.~Fung, ``Multi-hop question
  generation with graph convolutional network,'' in \emph{Proceedings of the
  2020 Conference on Empirical Methods in Natural Language Processing:
  Findings}, 2020, pp. 4636--4647.

\bibitem[Lin et~al.(2021)Lin, Hilton, and Evans]{lin2021truthfulqa}
S.~Lin, J.~Hilton, and O.~Evans, ``Truthfulqa: Measuring how models mimic human
  falsehoods,'' \emph{arXiv preprint arXiv:2109.07958}, 2021.

\bibitem[Nakano et~al.(2021)Nakano, Hilton, Balaji, Wu, Ouyang, Kim, Hesse,
  Jain, Kosaraju, Saunders, et~al.]{nakano2021webgpt}
R.~Nakano, J.~Hilton, S.~Balaji, J.~Wu, L.~Ouyang, C.~Kim, C.~Hesse, S.~Jain,
  V.~Kosaraju, W.~Saunders \emph{et~al.}, ``Webgpt: Browser-assisted
  question-answering with human feedback,'' \emph{arXiv preprint
  arXiv:2112.09332}, 2021.

\bibitem[Ji et~al.(2022)Ji, Lee, Frieske, Yu, Su, Xu, Ishii, Bang, Madotto, and
  Fung]{ji2022survey}
Z.~Ji, N.~Lee, R.~Frieske, T.~Yu, D.~Su, Y.~Xu, E.~Ishii, Y.~Bang, A.~Madotto,
  and P.~Fung, ``Survey of hallucination in natural language generation,''
  \emph{arXiv preprint arXiv:2202.03629}, 2022.

\bibitem[Brown et~al.(2020)Brown, Mann, Ryder, Subbiah, Kaplan, Dhariwal,
  Neelakantan, Shyam, Sastry, Askell, et~al.]{brown2020language}
T.~B. Brown, B.~Mann, N.~Ryder, M.~Subbiah, J.~Kaplan, P.~Dhariwal,
  A.~Neelakantan, P.~Shyam, G.~Sastry, A.~Askell \emph{et~al.}, ``Language
  models are few-shot learners,'' \emph{arXiv preprint arXiv:2005.14165}, 2020.

\bibitem[Ye et~al.(2020)Ye, Li, Wang, Bolte, Ma, Yih, Ren, and
  Khabsa]{ye2020studying}
Q.~Ye, B.~Z. Li, S.~Wang, B.~Bolte, H.~Ma, W.-t. Yih, X.~Ren, and M.~Khabsa,
  ``Studying strategically: Learning to mask for closed-book qa,'' \emph{arXiv
  preprint arXiv:2012.15856}, 2020.

\bibitem[Chen(2018)]{chen2018neural}
D.~Chen, \emph{Neural reading comprehension and beyond}.\hskip 1em plus 0.5em
  minus 0.4em\relax Stanford University, 2018.

\bibitem[Lewis et~al.(2020{\natexlab{b}})Lewis, Perez, Piktus, Petroni,
  Karpukhin, Goyal, K\"{u}ttler, Lewis, Yih, Rockt\"{a}schel, Riedel, and
  Kiela]{NEURIPS2020_6b493230}
\BIBentryALTinterwordspacing
P.~Lewis, E.~Perez, A.~Piktus, F.~Petroni, V.~Karpukhin, N.~Goyal,
  H.~K\"{u}ttler, M.~Lewis, W.-t. Yih, T.~Rockt\"{a}schel, S.~Riedel, and
  D.~Kiela, ``Retrieval-augmented generation for knowledge-intensive nlp
  tasks,'' in \emph{Advances in Neural Information Processing Systems},
  H.~Larochelle, M.~Ranzato, R.~Hadsell, M.~F. Balcan, and H.~Lin, Eds.,
  vol.~33.\hskip 1em plus 0.5em minus 0.4em\relax Curran Associates, Inc.,
  2020, pp. 9459--9474. [Online]. Available:
  \url{https://proceedings.neurips.cc/paper/2020/file/6b493230205f780e1bc26945df7481e5-Paper.pdf}
\BIBentrySTDinterwordspacing

\bibitem[Reddy et~al.(2019)Reddy, Chen, and Manning]{reddy2019coqa}
S.~Reddy, D.~Chen, and C.~D. Manning, ``Coqa: A conversational question
  answering challenge,'' \emph{Transactions of the Association for
  Computational Linguistics}, vol.~7, pp. 249--266, 2019.

\bibitem[Ju et~al.(2019)Ju, Zhao, Chen, Zheng, Yang, and Liu]{ju2019technical}
Y.~Ju, F.~Zhao, S.~Chen, B.~Zheng, X.~Yang, and Y.~Liu, ``Technical report on
  conversational question answering,'' \emph{arXiv preprint arXiv:1909.10772},
  2019.

\bibitem[Rajpurkar et~al.(2016)Rajpurkar, Zhang, Lopyrev, and
  Liang]{rajpurkar2016squad}
P.~Rajpurkar, J.~Zhang, K.~Lopyrev, and P.~Liang, ``Squad: 100,000+ questions
  for machine comprehension of text,'' \emph{arXiv preprint arXiv:1606.05250},
  2016.

\bibitem[Yao and Van~Durme(2014)]{yao2014information}
X.~Yao and B.~Van~Durme, ``Information extraction over structured data:
  Question answering with freebase,'' in \emph{Proceedings of the 52nd Annual
  Meeting of the Association for Computational Linguistics (Volume 1: Long
  Papers)}, 2014, pp. 956--966.

\bibitem[Vrande{\v{c}}i{\'c} and Kr{\"o}tzsch(2014)]{vrandevcic2014wikidata}
D.~Vrande{\v{c}}i{\'c} and M.~Kr{\"o}tzsch, ``Wikidata: a free collaborative
  knowledgebase,'' \emph{Communications of the ACM}, vol.~57, no.~10, pp.
  78--85, 2014.

\bibitem[Auer et~al.(2007)Auer, Bizer, Kobilarov, Lehmann, Cyganiak, and
  Ives]{auer2007dbpedia}
S.~Auer, C.~Bizer, G.~Kobilarov, J.~Lehmann, R.~Cyganiak, and Z.~Ives,
  ``Dbpedia: A nucleus for a web of open data,'' in \emph{The semantic
  web}.\hskip 1em plus 0.5em minus 0.4em\relax Springer, 2007, pp. 722--735.

\bibitem[Suchanek et~al.(2007)Suchanek, Kasneci, and Weikum]{suchanek2007yago}
F.~M. Suchanek, G.~Kasneci, and G.~Weikum, ``Yago: a core of semantic
  knowledge,'' in \emph{Proceedings of the 16th international conference on
  World Wide Web}.\hskip 1em plus 0.5em minus 0.4em\relax ACM, 2007, pp.
  697--706.

\bibitem[Mitchell et~al.(2018)Mitchell, Cohen, Hruschka, Talukdar, Yang,
  Betteridge, Carlson, Dalvi, Gardner, Kisiel, et~al.]{mitchell2018never}
T.~Mitchell, W.~Cohen, E.~Hruschka, P.~Talukdar, B.~Yang, J.~Betteridge,
  A.~Carlson, B.~Dalvi, M.~Gardner, B.~Kisiel \emph{et~al.}, ``Never-ending
  learning,'' \emph{Communications of the ACM}, vol.~61, no.~5, pp. 103--115,
  2018.

\bibitem[Pasupat and Liang(2015)]{pasupat2015compositional}
P.~Pasupat and P.~Liang, ``Compositional semantic parsing on semi-structured
  tables,'' \emph{arXiv preprint arXiv:1508.00305}, 2015.

\bibitem[Goyal et~al.(2017)Goyal, Khot, Summers{-}Stay, Batra, and
  Parikh]{balanced_vqa_v2}
Y.~Goyal, T.~Khot, D.~Summers{-}Stay, D.~Batra, and D.~Parikh, ``Making the {V}
  in {VQA} matter: Elevating the role of image understanding in {V}isual
  {Q}uestion {A}nswering,'' in \emph{Conference on Computer Vision and Pattern
  Recognition (CVPR)}, 2017.

\bibitem[Roberts et~al.(2020)Roberts, Raffel, and Shazeer]{roberts2020much}
A.~Roberts, C.~Raffel, and N.~Shazeer, ``How much knowledge can you pack into
  the parameters of a language model?'' in \emph{Proceedings of the 2020
  Conference on Empirical Methods in Natural Language Processing (EMNLP)},
  2020, pp. 5418--5426.

\bibitem[Petroni et~al.(2019)Petroni, Rockt{\"a}schel, Lewis, Bakhtin, Wu,
  Miller, and Riedel]{petroni2019language}
F.~Petroni, T.~Rockt{\"a}schel, P.~Lewis, A.~Bakhtin, Y.~Wu, A.~H. Miller, and
  S.~Riedel, ``Language models as knowledge bases?'' \emph{arXiv preprint
  arXiv:1909.01066}, 2019.

\bibitem[Radford et~al.(2019)Radford, Wu, Child, Luan, Amodei, and
  Sutskever]{radford2019language}
A.~Radford, J.~Wu, R.~Child, D.~Luan, D.~Amodei, and I.~Sutskever, ``Language
  models are unsupervised multitask learners,'' \emph{OpenAI blog}, vol.~1,
  no.~8, p.~9, 2019.

\bibitem[Hochreiter and Schmidhuber(1997)]{hochreiter1997long}
S.~Hochreiter and J.~Schmidhuber, ``Long short-term memory,'' \emph{Neural
  computation}, vol.~9, no.~8, pp. 1735--1780, 1997.

\bibitem[Cho et~al.(2014)Cho, van Merri{\"e}nboer, Gulcehre, Bahdanau,
  Bougares, Schwenk, and Bengio]{cho2014learning}
K.~Cho, B.~van Merri{\"e}nboer, C.~Gulcehre, D.~Bahdanau, F.~Bougares,
  H.~Schwenk, and Y.~Bengio, ``Learning phrase representations using rnn
  encoder--decoder for statistical machine translation,'' in \emph{Proceedings
  of the 2014 Conference on Empirical Methods in Natural Language Processing
  (EMNLP)}, 2014, pp. 1724--1734.

\bibitem[Vaswani et~al.(2017{\natexlab{a}})Vaswani, Shazeer, Parmar, Uszkoreit,
  Jones, Gomez, Kaiser, and Polosukhin]{NIPS2017_3f5ee243}
\BIBentryALTinterwordspacing
A.~Vaswani, N.~Shazeer, N.~Parmar, J.~Uszkoreit, L.~Jones, A.~N. Gomez, L.~u.
  Kaiser, and I.~Polosukhin, ``Attention is all you need,'' in \emph{Advances
  in Neural Information Processing Systems}, I.~Guyon, U.~V. Luxburg,
  S.~Bengio, H.~Wallach, R.~Fergus, S.~Vishwanathan, and R.~Garnett, Eds.,
  vol.~30.\hskip 1em plus 0.5em minus 0.4em\relax Curran Associates, Inc.,
  2017. [Online]. Available:
  \url{https://proceedings.neurips.cc/paper/2017/file/3f5ee243547dee91fbd053c1c4a845aa-Paper.pdf}
\BIBentrySTDinterwordspacing

\bibitem[Devlin et~al.(2019)Devlin, Chang, Lee, and Toutanova]{devlin2019bert}
J.~Devlin, M.-W. Chang, K.~Lee, and K.~Toutanova, ``Bert: Pre-training of deep
  bidirectional transformers for language understanding,'' in \emph{Proceedings
  of the 2019 Conference of the North American Chapter of the Association for
  Computational Linguistics: Human Language Technologies, Volume 1 (Long and
  Short Papers)}, 2019, pp. 4171--4186.

\bibitem[Wang et~al.(2018)Wang, Singh, Michael, Hill, Levy, and
  Bowman]{wang-etal-2018-glue}
\BIBentryALTinterwordspacing
A.~Wang, A.~Singh, J.~Michael, F.~Hill, O.~Levy, and S.~Bowman, ``{GLUE}: A
  multi-task benchmark and analysis platform for natural language
  understanding,'' in \emph{Proceedings of the 2018 {EMNLP} Workshop
  {B}lackbox{NLP}: Analyzing and Interpreting Neural Networks for {NLP}}.\hskip
  1em plus 0.5em minus 0.4em\relax Brussels, Belgium: Association for
  Computational Linguistics, Nov. 2018, pp. 353--355. [Online]. Available:
  \url{https://aclanthology.org/W18-5446}
\BIBentrySTDinterwordspacing

\bibitem[Lehnert(1977)]{lehnert1977process}
W.~G. Lehnert, \emph{The process of question answering.}\hskip 1em plus 0.5em
  minus 0.4em\relax Yale University, 1977.

\bibitem[Chen et~al.(2016)Chen, Bolton, and Manning]{chen2016thorough}
D.~Chen, J.~Bolton, and C.~D. Manning, ``A thorough examination of the
  cnn/daily mail reading comprehension task,'' in \emph{54th Annual Meeting of
  the Association for Computational Linguistics, ACL 2016}.\hskip 1em plus
  0.5em minus 0.4em\relax Association for Computational Linguistics (ACL),
  2016, pp. 2358--2367.

\bibitem[Yang et~al.(2019{\natexlab{a}})Yang, Dai, Yang, Carbonell,
  Salakhutdinov, and Le]{yang2019xlnet}
Z.~Yang, Z.~Dai, Y.~Yang, J.~Carbonell, R.~Salakhutdinov, and Q.~V. Le,
  ``Xlnet: Generalized autoregressive pretraining for language understanding,''
  \emph{arXiv preprint arXiv:1906.08237}, 2019.

\bibitem[Lovenia et~al.(2022)Lovenia, Wilie, Chung, Min, Cahyawijaya, Su, and
  Fung]{lovenia2022clozer}
H.~Lovenia, B.~Wilie, W.~Chung, Z.~Min, S.~Cahyawijaya, D.~Su, and P.~Fung,
  ``Clozer”:" adaptable data augmentation for cloze-style reading
  comprehension,'' in \emph{Proceedings of the 7th Workshop on Representation
  Learning for NLP}, 2022, pp. 60--66.

\bibitem[Zhang et~al.(2021)Zhang, Lyu, Ding, Shen, Jia, Han, and
  Knight]{zhang2021hybrid}
B.~Zhang, Y.~Lyu, N.~Ding, T.~Shen, Z.~Jia, K.~Han, and K.~Knight, ``A hybrid
  task-oriented dialog system with domain and task adaptive pretraining,''
  \emph{arXiv preprint arXiv:2102.04506}, 2021.

\bibitem[Lewis et~al.(2020{\natexlab{c}})Lewis, Perez, Piktus, Petroni,
  Karpukhin, Goyal, K\"{u}ttler, Lewis, Yih, Rockt\"{a}schel, Riedel, and
  Kiela]{rag}
\BIBentryALTinterwordspacing
P.~Lewis, E.~Perez, A.~Piktus, F.~Petroni, V.~Karpukhin, N.~Goyal,
  H.~K\"{u}ttler, M.~Lewis, W.-t. Yih, T.~Rockt\"{a}schel, S.~Riedel, and
  D.~Kiela, ``Retrieval-augmented generation for knowledge-intensive nlp
  tasks,'' in \emph{Advances in Neural Information Processing Systems},
  H.~Larochelle, M.~Ranzato, R.~Hadsell, M.~F. Balcan, and H.~Lin, Eds.,
  vol.~33.\hskip 1em plus 0.5em minus 0.4em\relax Curran Associates, Inc.,
  2020, pp. 9459--9474. [Online]. Available:
  \url{https://proceedings.neurips.cc/paper/2020/file/6b493230205f780e1bc26945df7481e5-Paper.pdf}
\BIBentrySTDinterwordspacing

\bibitem[Lazaridou et~al.(2022)Lazaridou, Gribovskaya, Stokowiec, and
  Grigorev]{lazaridou2022internet}
A.~Lazaridou, E.~Gribovskaya, W.~Stokowiec, and N.~Grigorev,
  ``Internet-augmented language models through few-shot prompting for
  open-domain question answering,'' \emph{arXiv preprint arXiv:2203.05115},
  2022.

\bibitem[Lai et~al.(2017)Lai, Xie, Liu, Yang, and Hovy]{lai2017race}
G.~Lai, Q.~Xie, H.~Liu, Y.~Yang, and E.~Hovy, ``Race: Large-scale reading
  comprehension dataset from examinations,'' in \emph{Proceedings of the 2017
  Conference on Empirical Methods in Natural Language Processing}, 2017, pp.
  785--794.

\bibitem[Saha et~al.(2018)Saha, Aralikatte, Khapra, and
  Sankaranarayanan]{saha2018duorc}
A.~Saha, R.~Aralikatte, M.~M. Khapra, and K.~Sankaranarayanan, ``Duorc: Towards
  complex language understanding with paraphrased reading comprehension,'' in
  \emph{Proceedings of the 56th Annual Meeting of the Association for
  Computational Linguistics (Volume 1: Long Papers)}, 2018, pp. 1683--1693.

\bibitem[Trischler et~al.(2017)Trischler, Wang, Yuan, Harris, Sordoni, Bachman,
  and Suleman]{trischler2017newsqa}
A.~Trischler, T.~Wang, X.~Yuan, J.~Harris, A.~Sordoni, P.~Bachman, and
  K.~Suleman, ``Newsqa: A machine comprehension dataset,'' in \emph{Proceedings
  of the 2nd Workshop on Representation Learning for NLP}, 2017, pp. 191--200.

\bibitem[Joshi et~al.(2017{\natexlab{b}})Joshi, Choi, Weld, and
  Zettlemoyer]{Joshi_2017}
\BIBentryALTinterwordspacing
M.~Joshi, E.~Choi, D.~Weld, and L.~Zettlemoyer, ``Triviaqa: A large scale
  distantly supervised challenge dataset for reading comprehension,''
  \emph{Proceedings of the 55th Annual Meeting of the Association for
  Computational Linguistics (Volume 1: Long Papers)}, 2017. [Online].
  Available: \url{http://dx.doi.org/10.18653/v1/p17-1147}
\BIBentrySTDinterwordspacing

\bibitem[Seo et~al.(2017)Seo, Kembhavi, Farhadi, and
  Hajishirzi]{seo2017bidirectional}
M.~J. Seo, A.~Kembhavi, A.~Farhadi, and H.~Hajishirzi, ``Bidirectional
  attention flow for machine comprehension,'' in \emph{5th International
  Conference on Learning Representations, {ICLR} 2017, Toulon, France, April
  24-26, 2017, Conference Track Proceedings}, 2017.

\bibitem[Cui et~al.(2017)Cui, Chen, Wei, Wang, Liu, and Hu]{cui2017attention}
Y.~Cui, Z.~Chen, S.~Wei, S.~Wang, T.~Liu, and G.~Hu, ``Attention-over-attention
  neural networks for reading comprehension,'' in \emph{Proceedings of the 55th
  Annual Meeting of the Association for Computational Linguistics (Volume 1:
  Long Papers)}, 2017, pp. 593--602.

\bibitem[Dhingra et~al.(2017)Dhingra, Liu, Yang, Cohen, and
  Salakhutdinov]{dhingra2017gated}
B.~Dhingra, H.~Liu, Z.~Yang, W.~Cohen, and R.~Salakhutdinov, ``Gated-attention
  readers for text comprehension,'' in \emph{Proceedings of the 55th Annual
  Meeting of the Association for Computational Linguistics (Volume 1: Long
  Papers)}, 2017, pp. 1832--1846.

\bibitem[Su et~al.(2019{\natexlab{b}})Su, Xu, Winata, Xu, Kim, Liu, and
  Fung]{su-etal-2019-generalizing}
\BIBentryALTinterwordspacing
D.~Su, Y.~Xu, G.~I. Winata, P.~Xu, H.~Kim, Z.~Liu, and P.~Fung, ``Generalizing
  question answering system with pre-trained language model fine-tuning,'' in
  \emph{Proceedings of the 2nd Workshop on Machine Reading for Question
  Answering}.\hskip 1em plus 0.5em minus 0.4em\relax Hong Kong, China:
  Association for Computational Linguistics, Nov. 2019, pp. 203--211. [Online].
  Available: \url{https://aclanthology.org/D19-5827}
\BIBentrySTDinterwordspacing

\bibitem[Tombros and Sanderson(1998)]{tombros1998advantages}
A.~Tombros and M.~Sanderson, ``Advantages of query biased summaries in
  information retrieval,'' in \emph{Proceedings of the 21st annual
  international ACM SIGIR conference on Research and development in information
  retrieval}, 1998, pp. 2--10.

\bibitem[Su et~al.(2020{\natexlab{b}})Su, Xu, Yu, Siddique, Barezi, and
  Fung]{su2020caire}
D.~Su, Y.~Xu, T.~Yu, F.~B. Siddique, E.~Barezi, and P.~Fung, ``Caire-covid: A
  question answering and query-focused multi-document summarization system for
  covid-19 scholarly information management,'' in \emph{Proceedings of the 1st
  Workshop on NLP for COVID-19 (Part 2) at EMNLP 2020}, 2020.

\bibitem[Davis et~al.(2012)Davis, Conroy, and Schlesinger]{davis2012occams}
S.~T. Davis, J.~M. Conroy, and J.~D. Schlesinger, ``Occams--an optimal
  combinatorial covering algorithm for multi-document summarization,'' in
  \emph{2012 IEEE 12th International Conference on Data Mining
  Workshops}.\hskip 1em plus 0.5em minus 0.4em\relax IEEE, 2012, pp. 454--463.

\bibitem[Daum{\'e}~III and Marcu(2006)]{daume2006bayesian}
H.~Daum{\'e}~III and D.~Marcu, ``Bayesian query-focused summarization,'' in
  \emph{Proceedings of the 21st International Conference on Computational
  Linguistics and 44th Annual Meeting of the Association for Computational
  Linguistics}, 2006, pp. 305--312.

\bibitem[Feigenblat et~al.(2017)Feigenblat, Roitman, Boni, and
  Konopnicki]{feigenblat2017unsupervised}
G.~Feigenblat, H.~Roitman, O.~Boni, and D.~Konopnicki, ``Unsupervised
  query-focused multi-document summarization using the cross entropy method,''
  in \emph{Proceedings of the 40th International ACM SIGIR Conference on
  Research and Development in Information Retrieval}, 2017, pp. 961--964.

\bibitem[Xu and Lapata(2020{\natexlab{a}})]{xu2020coarse}
Y.~Xu and M.~Lapata, ``Coarse-to-fine query focused multi-document
  summarization,'' in \emph{Proceedings of the 2020 Conference on Empirical
  Methods in Natural Language Processing (EMNLP)}, 2020, pp. 3632--3645.

\bibitem[Nema et~al.(2017)Nema, Khapra, Laha, and Ravindran]{nema2017diversity}
P.~Nema, M.~Khapra, A.~Laha, and B.~Ravindran, ``Diversity driven attention
  model for query-based abstractive summarization,'' \emph{arXiv preprint
  arXiv:1704.08300}, 2017.

\bibitem[Baumel et~al.(2018)Baumel, Eyal, and Elhadad]{baumel2018query}
T.~Baumel, M.~Eyal, and M.~Elhadad, ``Query focused abstractive summarization:
  Incorporating query relevance, multi-document coverage, and summary length
  constraints into seq2seq models,'' \emph{arXiv preprint arXiv:1801.07704},
  2018.

\bibitem[Xu and Lapata(2020{\natexlab{b}})]{xu2020abstractive}
Y.~Xu and M.~Lapata, ``Abstractive query focused summarization with query-free
  resources,'' \emph{arXiv preprint arXiv:2012.14774}, 2020.

\bibitem[Wei et~al.(2022)Wei, Wang, Schuurmans, Bosma, Chi, Le, and
  Zhou]{wei2022chain}
J.~Wei, X.~Wang, D.~Schuurmans, M.~Bosma, E.~Chi, Q.~Le, and D.~Zhou, ``Chain
  of thought prompting elicits reasoning in large language models,''
  \emph{arXiv preprint arXiv:2201.11903}, 2022.

\bibitem[Wang et~al.(2022)Wang, Wei, Schuurmans, Le, Chi, and
  Zhou]{wang2022self}
X.~Wang, J.~Wei, D.~Schuurmans, Q.~Le, E.~Chi, and D.~Zhou, ``Self-consistency
  improves chain of thought reasoning in language models,'' \emph{arXiv
  preprint arXiv:2203.11171}, 2022.

\bibitem[Wang et~al.(2020{\natexlab{a}})Wang, Lo, Chandrasekhar, Reas, Yang,
  Eide, Funk, Kinney, Liu, Merrill, et~al.]{wang2020cord}
L.~L. Wang, K.~Lo, Y.~Chandrasekhar, R.~Reas, J.~Yang, D.~Eide, K.~Funk,
  R.~Kinney, Z.~Liu, W.~Merrill \emph{et~al.}, ``Cord-19: The covid-19 open
  research dataset,'' \emph{ArXiv}, 2020.

\bibitem[Lewis et~al.(2019)Lewis, Liu, Goyal, Ghazvininejad, Mohamed, Levy,
  Stoyanov, and Zettlemoyer]{lewis2019bart}
M.~Lewis, Y.~Liu, N.~Goyal, M.~Ghazvininejad, A.~Mohamed, O.~Levy, V.~Stoyanov,
  and L.~Zettlemoyer, ``Bart: Denoising sequence-to-sequence pre-training for
  natural language generation, translation, and comprehension,'' \emph{arXiv
  preprint arXiv:1910.13461}, 2019.

\bibitem[Lee et~al.(2020)Lee, Yoon, Kim, Kim, Kim, So, and
  Kang]{lee2020biobert}
J.~Lee, W.~Yoon, S.~Kim, D.~Kim, S.~Kim, C.~H. So, and J.~Kang, ``Biobert: a
  pre-trained biomedical language representation model for biomedical text
  mining,'' \emph{Bioinformatics}, vol.~36, no.~4, pp. 1234--1240, 2020.

\bibitem[Yang et~al.(2018{\natexlab{b}})Yang, Fang, and Lin]{yang2018anserini}
P.~Yang, H.~Fang, and J.~Lin, ``Anserini: Reproducible ranking baselines using
  lucene,'' \emph{Journal of Data and Information Quality (JDIQ)}, vol.~10,
  no.~4, pp. 1--20, 2018.

\bibitem[Hermann et~al.(2015)Hermann, Kocisky, Grefenstette, Espeholt, Kay,
  Suleyman, and Blunsom]{hermann2015teaching}
K.~M. Hermann, T.~Kocisky, E.~Grefenstette, L.~Espeholt, W.~Kay, M.~Suleyman,
  and P.~Blunsom, ``Teaching machines to read and comprehend,'' \emph{Advances
  in neural information processing systems}, vol.~28, pp. 1693--1701, 2015.

\bibitem[Narayan et~al.(2018)Narayan, Cohen, and Lapata]{narayan2018don}
S.~Narayan, S.~B. Cohen, and M.~Lapata, ``Don't give me the details, just the
  summary! topic-aware convolutional neural networks for extreme
  summarization,'' \emph{arXiv preprint arXiv:1808.08745}, 2018.

\bibitem[Savery et~al.(2020)Savery, Abacha, Gayen, and
  Demner-Fushman]{savery2020question}
M.~Savery, A.~B. Abacha, S.~Gayen, and D.~Demner-Fushman, ``Question-driven
  summarization of answers to consumer health questions,'' \emph{arXiv preprint
  arXiv:2005.09067}, 2020.

\bibitem[Lan et~al.(2019)Lan, Chen, Goodman, Gimpel, Sharma, and
  Soricut]{lan2019albert}
Z.~Lan, M.~Chen, S.~Goodman, K.~Gimpel, P.~Sharma, and R.~Soricut, ``Albert: A
  lite bert for self-supervised learning of language representations,''
  \emph{arXiv preprint arXiv:1909.11942}, 2019.

\bibitem[Dang(2005)]{dang2005overview}
H.~T. Dang, ``Overview of duc 2005,'' in \emph{Proceedings of the document
  understanding conference}, vol. 2005, 2005, pp. 1--12.

\bibitem[Hoa(2006)]{hoa2006overview}
T.~Hoa, ``Overview of duc 2006,'' in \emph{Document Understanding Conference},
  2006.

\bibitem[Xu and Lapata(2020{\natexlab{c}})]{xu2020query}
Y.~Xu and M.~Lapata, ``Query focused multi-document summarization with distant
  supervision,'' \emph{arXiv preprint arXiv:2004.03027}, 2020.

\bibitem[Paulus et~al.(2017)Paulus, Xiong, and Socher]{paulus2017deep}
R.~Paulus, C.~Xiong, and R.~Socher, ``A deep reinforced model for abstractive
  summarization,'' \emph{arXiv preprint arXiv:1705.04304}, 2017.

\bibitem[Gehrmann et~al.(2018)Gehrmann, Deng, and Rush]{gehrmann2018bottom}
S.~Gehrmann, Y.~Deng, and A.~M. Rush, ``Bottom-up abstractive summarization,''
  \emph{arXiv preprint arXiv:1808.10792}, 2018.

\bibitem[Zhang et~al.(2020)Zhang, Zhao, Saleh, and Liu]{zhang2020pegasus}
J.~Zhang, Y.~Zhao, M.~Saleh, and P.~Liu, ``Pegasus: Pre-training with extracted
  gap-sentences for abstractive summarization,'' in \emph{International
  Conference on Machine Learning}.\hskip 1em plus 0.5em minus 0.4em\relax PMLR,
  2020, pp. 11\,328--11\,339.

\bibitem[Yang et~al.(2019{\natexlab{b}})Yang, Xie, Lin, Li, Tan, Xiong, Li, and
  Lin]{yang2019end}
W.~Yang, Y.~Xie, A.~Lin, X.~Li, L.~Tan, K.~Xiong, M.~Li, and J.~Lin,
  ``End-to-end open-domain question answering with bertserini,'' \emph{arXiv
  preprint arXiv:1902.01718}, 2019.

\bibitem[Vaswani et~al.(2017{\natexlab{b}})Vaswani, Shazeer, Parmar, Uszkoreit,
  Jones, Gomez, Kaiser, and Polosukhin]{vaswani2017attention}
A.~Vaswani, N.~Shazeer, N.~Parmar, J.~Uszkoreit, L.~Jones, A.~N. Gomez,
  {\L}.~Kaiser, and I.~Polosukhin, ``Attention is all you need,'' in
  \emph{Advances in neural information processing systems}, 2017, pp.
  5998--6008.

\bibitem[Xie et~al.(2020)Xie, Zhou, Mao, and Chen]{xie2020conditional}
Y.~Xie, T.~Zhou, Y.~Mao, and W.~Chen, ``Conditional self-attention for
  query-based summarization,'' \emph{arXiv preprint arXiv:2002.07338}, 2020.

\bibitem[Laskar et~al.(2020)Laskar, Hoque, and Huang]{laskar2020query}
M.~T.~R. Laskar, E.~Hoque, and J.~Huang, ``Query focused abstractive
  summarization via incorporating query relevance and transfer learning with
  transformer models,'' in \emph{Canadian Conference on Artificial
  Intelligence}.\hskip 1em plus 0.5em minus 0.4em\relax Springer, 2020, pp.
  342--348.

\bibitem[Dong et~al.(2019)Dong, Yang, Wang, Wei, Liu, Wang, Gao, Zhou, and
  Hon]{dong2019unified}
L.~Dong, N.~Yang, W.~Wang, F.~Wei, X.~Liu, Y.~Wang, J.~Gao, M.~Zhou, and H.-W.
  Hon, ``Unified language model pre-training for natural language understanding
  and generation,'' in \emph{Advances in Neural Information Processing
  Systems}, 2019, pp. 13\,063--13\,075.

\bibitem[Raffel et~al.(2019{\natexlab{b}})Raffel, Shazeer, Roberts, Lee,
  Narang, Matena, Zhou, Li, and Liu]{raffel2019exploring}
C.~Raffel, N.~Shazeer, A.~Roberts, K.~Lee, S.~Narang, M.~Matena, Y.~Zhou,
  W.~Li, and P.~J. Liu, ``Exploring the limits of transfer learning with a
  unified text-to-text transformer,'' \emph{arXiv preprint arXiv:1910.10683},
  2019.

\bibitem[Blitzer et~al.(2007)Blitzer, Dredze, and
  Pereira]{blitzer2007biographies}
J.~Blitzer, M.~Dredze, and F.~Pereira, ``Biographies, bollywood, boom-boxes and
  blenders: Domain adaptation for sentiment classification,'' in
  \emph{Proceedings of the 45th annual meeting of the association of
  computational linguistics}, 2007, pp. 440--447.

\bibitem[Daum{\'e}~III(2009)]{daume2009frustratingly}
H.~Daum{\'e}~III, ``Frustratingly easy domain adaptation,'' \emph{arXiv
  preprint arXiv:0907.1815}, 2009.

\bibitem[Liu et~al.(2020)Liu, Xu, Yu, Dai, Ji, Cahyawijaya, Madotto, and
  Fung]{liu2020crossner}
Z.~Liu, Y.~Xu, T.~Yu, W.~Dai, Z.~Ji, S.~Cahyawijaya, A.~Madotto, and P.~Fung,
  ``Crossner: Evaluating cross-domain named entity recognition,'' \emph{arXiv
  preprint arXiv:2012.04373}, 2020.

\bibitem[Yu et~al.(2021)Yu, Liu, and Fung]{yu2021adaptsum}
T.~Yu, Z.~Liu, and P.~Fung, ``Adaptsum: Towards low-resource domain adaptation
  for abstractive summarization,'' \emph{arXiv preprint arXiv:2103.11332},
  2021.

\bibitem[Hua and Wang(2017)]{hua2017pilot}
X.~Hua and L.~Wang, ``A pilot study of domain adaptation effect for neural
  abstractive summarization,'' \emph{arXiv preprint arXiv:1707.07062}, 2017.

\bibitem[Trischler et~al.(2016)Trischler, Wang, Yuan, Harris, Sordoni, Bachman,
  and Suleman]{trischler2016newsqa}
A.~Trischler, T.~Wang, X.~Yuan, J.~Harris, A.~Sordoni, P.~Bachman, and
  K.~Suleman, ``Newsqa: A machine comprehension dataset,'' \emph{arXiv preprint
  arXiv:1611.09830}, 2016.

\bibitem[Dunn et~al.(2017)Dunn, Sagun, Higgins, Guney, Cirik, and
  Cho]{dunn2017searchqa}
M.~Dunn, L.~Sagun, M.~Higgins, V.~U. Guney, V.~Cirik, and K.~Cho, ``Searchqa: A
  new q\&a dataset augmented with context from a search engine,'' \emph{arXiv
  preprint arXiv:1704.05179}, 2017.

\bibitem[Liu and Lapata(2019)]{liu2019text}
Y.~Liu and M.~Lapata, ``Text summarization with pretrained encoders,''
  \emph{arXiv preprint arXiv:1908.08345}, 2019.

\bibitem[Lin(2004)]{lin2004rouge}
C.-Y. Lin, ``Rouge: A package for automatic evaluation of summaries,'' in
  \emph{Text summarization branches out}, 2004, pp. 74--81.

\bibitem[Mihalcea and Tarau(2004)]{mihalcea2004textrank}
R.~Mihalcea and P.~Tarau, ``Textrank: Bringing order into text,'' in
  \emph{Proceedings of the 2004 conference on empirical methods in natural
  language processing}, 2004, pp. 404--411.

\bibitem[Thorne et~al.(2018)Thorne, Vlachos, Christodoulopoulos, and
  Mittal]{thorne2018fever}
J.~Thorne, A.~Vlachos, C.~Christodoulopoulos, and A.~Mittal, ``Fever: a
  large-scale dataset for fact extraction and verification,'' in
  \emph{Proceedings of the 2018 Conference of the North American Chapter of the
  Association for Computational Linguistics: Human Language Technologies,
  Volume 1 (Long Papers)}, 2018, pp. 809--819.

\bibitem[Levy et~al.(2017)Levy, Seo, Choi, and Zettlemoyer]{levy2017zero}
O.~Levy, M.~Seo, E.~Choi, and L.~Zettlemoyer, ``Zero-shot relation extraction
  via reading comprehension,'' in \emph{Proceedings of the 21st Conference on
  Computational Natural Language Learning (CoNLL 2017)}, 2017, pp. 333--342.

\bibitem[Elsahar et~al.(2018{\natexlab{a}})Elsahar, Vougiouklis, Remaci,
  Gravier, Hare, Laforest, and Simperl]{elsahar2018t}
H.~Elsahar, P.~Vougiouklis, A.~Remaci, C.~Gravier, J.~Hare, F.~Laforest, and
  E.~Simperl, ``T-rex: A large scale alignment of natural language with
  knowledge base triples,'' in \emph{Proceedings of the Eleventh International
  Conference on Language Resources and Evaluation (LREC 2018)}, 2018.

\bibitem[Dinan et~al.(2018)Dinan, Roller, Shuster, Fan, Auli, and
  Weston]{dinan2018wizard}
E.~Dinan, S.~Roller, K.~Shuster, A.~Fan, M.~Auli, and J.~Weston, ``Wizard of
  wikipedia: Knowledge-powered conversational agents,'' in \emph{International
  Conference on Learning Representations}, 2018.

\bibitem[Joshi et~al.(2020)Joshi, Chen, Liu, Weld, Zettlemoyer, and
  Levy]{joshi2020spanbert}
M.~Joshi, D.~Chen, Y.~Liu, D.~S. Weld, L.~Zettlemoyer, and O.~Levy, ``Spanbert:
  Improving pre-training by representing and predicting spans,''
  \emph{Transactions of the Association for Computational Linguistics}, vol.~8,
  pp. 64--77, 2020.

\bibitem[Fisch et~al.(2019)Fisch, Talmor, Jia, Seo, Choi, and
  Chen]{fisch-etal-2019-mrqa}
\BIBentryALTinterwordspacing
A.~Fisch, A.~Talmor, R.~Jia, M.~Seo, E.~Choi, and D.~Chen, ``{MRQA} 2019 shared
  task: Evaluating generalization in reading comprehension,'' in
  \emph{Proceedings of the 2nd Workshop on Machine Reading for Question
  Answering}.\hskip 1em plus 0.5em minus 0.4em\relax Hong Kong, China:
  Association for Computational Linguistics, Nov. 2019, pp. 1--13. [Online].
  Available: \url{https://aclanthology.org/D19-5801}
\BIBentrySTDinterwordspacing

\bibitem[Elsahar et~al.(2018{\natexlab{b}})Elsahar, Vougiouklis, Remaci,
  Gravier, Hare, Laforest, and Simperl]{TREX}
H.~Elsahar, P.~Vougiouklis, A.~Remaci, C.~Gravier, J.~Hare, F.~Laforest, and
  E.~Simperl, ``T-rex: A large scale alignment of natural language with
  knowledge base triples,'' in \emph{Proceedings of the Eleventh International
  Conference on Language Resources and Evaluation (LREC 2018)}, 2018.

\bibitem[Petroni et~al.(2020)Petroni, Lewis, Piktus, Rockt{\"a}schel, Wu,
  Miller, and Riedel]{petronicontext}
F.~Petroni, P.~Lewis, A.~Piktus, T.~Rockt{\"a}schel, Y.~Wu, A.~H. Miller, and
  S.~Riedel, ``How context affects language models’ factual predictions,''
  \emph{AKBC 2020}, 2020.

\bibitem[Wang et~al.(2020{\natexlab{b}})Wang, Cho, and Lewis]{wang2020asking}
A.~Wang, K.~Cho, and M.~Lewis, ``Asking and answering questions to evaluate the
  factual consistency of summaries,'' \emph{Proceedings of the 58th Annual
  Meeting of the Association for Computational Linguistics}, 2020.

\bibitem[Durmus et~al.(2020)Durmus, He, and Diab]{durmus2020feqa}
E.~Durmus, H.~He, and M.~Diab, ``Feqa: A question answering evaluation
  framework for faithfulness assessment in abstractive summarization,'' in
  \emph{Proceedings of the 58th Annual Meeting of the Association for
  Computational Linguistics}, 2020, pp. 5055--5070.

\bibitem[Su et~al.(2022{\natexlab{b}})Su, Patwary, Prabhumoye, Xu, Prenger,
  Shoeybi, Fung, Anandkumar, and Catanzaro]{su2022context}
D.~Su, M.~Patwary, S.~Prabhumoye, P.~Xu, R.~Prenger, M.~Shoeybi, P.~Fung,
  A.~Anandkumar, and B.~Catanzaro, ``Context generation improves open domain
  question answering,'' \emph{arXiv preprint arXiv:2210.06349}, 2022.

\bibitem[Chowdhery et~al.(2022)Chowdhery, Narang, Devlin, Bosma, Mishra,
  Roberts, Barham, Chung, Sutton, Gehrmann, et~al.]{chowdhery2022palm}
A.~Chowdhery, S.~Narang, J.~Devlin, M.~Bosma, G.~Mishra, A.~Roberts, P.~Barham,
  H.~W. Chung, C.~Sutton, S.~Gehrmann \emph{et~al.}, ``Palm: Scaling language
  modeling with pathways,'' \emph{arXiv preprint arXiv:2204.02311}, 2022.

\bibitem[Smith et~al.(2022)Smith, Patwary, Norick, LeGresley, Rajbhandari,
  Casper, Liu, Prabhumoye, Zerveas, Korthikanti, et~al.]{smith2022using}
S.~Smith, M.~Patwary, B.~Norick, P.~LeGresley, S.~Rajbhandari, J.~Casper,
  Z.~Liu, S.~Prabhumoye, G.~Zerveas, V.~Korthikanti \emph{et~al.}, ``Using
  deepspeed and megatron to train megatron-turing nlg 530b, a large-scale
  generative language model,'' \emph{arXiv preprint arXiv:2201.11990}, 2022.

\bibitem[Liu et~al.(2021)Liu, Yuan, Fu, Jiang, Hayashi, and Neubig]{liu2021pre}
P.~Liu, W.~Yuan, J.~Fu, Z.~Jiang, H.~Hayashi, and G.~Neubig, ``Pre-train,
  prompt, and predict: A systematic survey of prompting methods in natural
  language processing,'' \emph{arXiv preprint arXiv:2107.13586}, 2021.

\bibitem[Liu et~al.(2022)Liu, Patwary, Prenger, Prabhumoye, Ping, Shoeybi, and
  Catanzaro]{liu2022multi}
Z.~Liu, M.~Patwary, R.~Prenger, S.~Prabhumoye, W.~Ping, M.~Shoeybi, and
  B.~Catanzaro, ``Multi-stage prompting for knowledgeable dialogue
  generation,'' in \emph{Findings of the Association for Computational
  Linguistics: ACL 2022}, 2022, pp. 1317--1337.

\bibitem[Shuster et~al.(2021)Shuster, Poff, Chen, Kiela, and
  Weston]{shuster2021retrieval}
K.~Shuster, S.~Poff, M.~Chen, D.~Kiela, and J.~Weston, ``Retrieval augmentation
  reduces hallucination in conversation,'' \emph{arXiv preprint
  arXiv:2104.07567}, 2021.

\bibitem[Su et~al.(2022{\natexlab{c}})Su, Li, Zhang, Shang, Jiang, Liu, and
  Fung]{su2022read}
D.~Su, X.~Li, J.~Zhang, L.~Shang, X.~Jiang, Q.~Liu, and P.~Fung, ``Read before
  generate! faithful long form question answering with machine reading,'' in
  \emph{Findings of the Association for Computational Linguistics: ACL 2022},
  2022, pp. 744--756.

\bibitem[Lewis et~al.(2020{\natexlab{d}})Lewis, Perez, Piktus, Petroni,
  Karpukhin, Goyal, K{\"u}ttler, Lewis, Yih, Rockt{\"a}schel,
  et~al.]{lewis2020retrieval}
P.~Lewis, E.~Perez, A.~Piktus, F.~Petroni, V.~Karpukhin, N.~Goyal,
  H.~K{\"u}ttler, M.~Lewis, W.-t. Yih, T.~Rockt{\"a}schel \emph{et~al.},
  ``Retrieval-augmented generation for knowledge-intensive nlp tasks,''
  \emph{arXiv preprint arXiv:2005.11401}, 2020.

\bibitem[Rae et~al.(2021)Rae, Borgeaud, Cai, Millican, Hoffmann, Song,
  Aslanides, Henderson, Ring, Young, Rutherford, Hennigan, Menick, Cassirer,
  Powell, Driessche, Hendricks, Rauh, Huang, Glaese, Welbl, Dathathri, Huang,
  Uesato, Mellor, Higgins, Creswell, McAleese, Wu, Elsen, Jayakumar,
  Buchatskaya, Budden, Sutherland, Simonyan, Paganini, Sifre, Martens, Li,
  Kuncoro, Nematzadeh, Gribovskaya, Donato, Lazaridou, Mensch, Lespiau,
  Tsimpoukelli, Grigorev, Fritz, Sottiaux, Pajarskas, Pohlen, Gong, Toyama,
  d'Autume, Li, Terzi, Mikulik, Babuschkin, Clark, Casas, Guy, Jones, Bradbury,
  Johnson, Hechtman, Weidinger, Gabriel, Isaac, Lockhart, Osindero, Rimell,
  Dyer, Vinyals, Ayoub, Stanway, Bennett, Hassabis, Kavukcuoglu, and
  Irving]{RaeGopher2022}
\BIBentryALTinterwordspacing
J.~W. Rae, S.~Borgeaud, T.~Cai, K.~Millican, J.~Hoffmann, F.~Song,
  J.~Aslanides, S.~Henderson, R.~Ring, S.~Young, E.~Rutherford, T.~Hennigan,
  J.~Menick, A.~Cassirer, R.~Powell, G.~v.~d. Driessche, L.~A. Hendricks,
  M.~Rauh, P.-S. Huang, A.~Glaese, J.~Welbl, S.~Dathathri, S.~Huang, J.~Uesato,
  J.~Mellor, I.~Higgins, A.~Creswell, N.~McAleese, A.~Wu, E.~Elsen,
  S.~Jayakumar, E.~Buchatskaya, D.~Budden, E.~Sutherland, K.~Simonyan,
  M.~Paganini, L.~Sifre, L.~Martens, X.~L. Li, A.~Kuncoro, A.~Nematzadeh,
  E.~Gribovskaya, D.~Donato, A.~Lazaridou, A.~Mensch, J.-B. Lespiau,
  M.~Tsimpoukelli, N.~Grigorev, D.~Fritz, T.~Sottiaux, M.~Pajarskas, T.~Pohlen,
  Z.~Gong, D.~Toyama, C.~d.~M. d'Autume, Y.~Li, T.~Terzi, V.~Mikulik,
  I.~Babuschkin, A.~Clark, D.~d.~L. Casas, A.~Guy, C.~Jones, J.~Bradbury,
  M.~Johnson, B.~Hechtman, L.~Weidinger, I.~Gabriel, W.~Isaac, E.~Lockhart,
  S.~Osindero, L.~Rimell, C.~Dyer, O.~Vinyals, K.~Ayoub, J.~Stanway,
  L.~Bennett, D.~Hassabis, K.~Kavukcuoglu, and G.~Irving, ``Scaling language
  models: Methods, analysis; insights from training gopher,'' 2021. [Online].
  Available: \url{https://arxiv.org/abs/2112.11446}
\BIBentrySTDinterwordspacing

\bibitem[Kwiatkowski et~al.(2019{\natexlab{b}})Kwiatkowski, Palomaki, Redfield,
  Collins, Parikh, Alberti, Epstein, Polosukhin, Kelcey, Devlin, Lee,
  Toutanova, Jones, Chang, Dai, Uszkoreit, Le, and Petrov]{naturalQ}
\BIBentryALTinterwordspacing
T.~Kwiatkowski, J.~Palomaki, O.~Redfield, M.~Collins, A.~Parikh, C.~Alberti,
  D.~Epstein, I.~Polosukhin, M.~Kelcey, J.~Devlin, K.~Lee, K.~N. Toutanova,
  L.~Jones, M.-W. Chang, A.~Dai, J.~Uszkoreit, Q.~Le, and S.~Petrov, ``Natural
  questions: a benchmark for question answering research,'' \emph{Transactions
  of the Association of Computational Linguistics}, 2019. [Online]. Available:
  \url{https://tomkwiat.users.x20web.corp.google.com/papers/natural-questions/main-1455-kwiatkowski.pdf}
\BIBentrySTDinterwordspacing

\bibitem[Dua et~al.(2019)Dua, Wang, Dasigi, Stanovsky, Singh, and
  Gardner]{Dua2019DROP}
D.~Dua, Y.~Wang, P.~Dasigi, G.~Stanovsky, S.~Singh, and M.~Gardner, ``{DROP}: A
  reading comprehension benchmark requiring discrete reasoning over
  paragraphs,'' in \emph{Proc. of NAACL}, 2019.

\bibitem[Kembhavi et~al.(2017)Kembhavi, Seo, Schwenk, Choi, Farhadi, and
  Hajishirzi]{TQA}
A.~Kembhavi, M.~Seo, D.~Schwenk, J.~Choi, A.~Farhadi, and H.~Hajishirzi, ``Are
  you smarter than a sixth grader? textbook question answering for multimodal
  machine comprehension,'' in \emph{Proceedings of the IEEE Conference on
  Computer Vision and Pattern Recognition}, 2017, pp. 4999--5007.

\bibitem[Tsatsaronis et~al.(2012)Tsatsaronis, Schroeder, Paliouras, Almirantis,
  Androutsopoulos, Gaussier, Gallinari, Artieres, Alvers, Zschunke,
  et~al.]{tsatsaronis2012bioasq}
G.~Tsatsaronis, M.~Schroeder, G.~Paliouras, Y.~Almirantis, I.~Androutsopoulos,
  E.~Gaussier, P.~Gallinari, T.~Artieres, M.~R. Alvers, M.~Zschunke
  \emph{et~al.}, ``Bioasq: A challenge on large-scale biomedical semantic
  indexing and question answering,'' in \emph{2012 AAAI Fall Symposium Series},
  2012.

\bibitem[Berant et~al.(2014)Berant, Srikumar, Chen, Vander~Linden, Harding,
  Huang, Clark, and Manning]{berant2014modeling}
J.~Berant, V.~Srikumar, P.-C. Chen, A.~Vander~Linden, B.~Harding, B.~Huang,
  P.~Clark, and C.~D. Manning, ``Modeling biological processes for reading
  comprehension,'' in \emph{Proceedings of the 2014 Conference on Empirical
  Methods in Natural Language Processing (EMNLP)}, 2014, pp. 1499--1510.

\end{thebibliography}

% \newpage
% \addcontentsline{toc}{chapter}{Publication}
% \include{publication}

% \newpage
% \addcontentsline{toc}{chapter}{Awards}
% \include{achievements}

\newpage
\addcontentsline{toc}{chapter}{Appendix}
\appendix
\chapter*{Appendix}

\section*{Chapter 3 Generating Query-relevant, Long-form Answers}

\subsection*{A Details of HLTC-MRQA Model}
\label{sec:appx-qa}
The MRQA model~\cite{su2019generalizing} is leveraged in the CAiRE-Covid system. To equip the model with better generalization ability to unseen data, the MRQA model is trained in a multi-task learning scheme on six datasets: SQuAD~\cite{rajpurkar2016squad}, NewsQA~\cite{trischler2017newsqa}, TriviaQA~\cite{Joshi_2017}, SearchQA~\cite{dunn2017searchqa}, HotpotQA~\cite{yang2018hotpotqa} and NaturalQuestions~\cite{naturalQ}. The training sets vary from each other in terms of data source, context lengths, whether multi-hop reasoning is needed and strategies for data augmentation. To evaluate the generalization ability, the authors utilized the BERT-large model~\cite{devlin2019bert}, which is trained with the same method as the MRQA model as the baseline. The models are evaluated on twelve unseen datasets, including DROP~\cite{Dua2019DROP} and TextbookQA~\cite{TQA}. From Table \ref{tab:mrqa}, the MRQA model consistently outperforms the baseline and achieves promising results on the QA samples, which are different from the training samples in terms of data resource, domain etc., including biomedical unseen datasets, such as BioASQ~\cite{tsatsaronis2012bioasq} and BioProcess~\cite{berant2014modeling}.
\begin{table}[ht!]
\centering
\begin{adjustbox}{width=0.46 \textwidth}
\begin{tabular}{l|cc|cc}
\hline
\multicolumn{1}{c|}{\multirow{2}{*}{Datasets}}  
& \multicolumn{2}{c|}{\textbf{MRQA model}} 
& \multicolumn{2}{c}{\textbf{Baseline}} \\ 
\cline{2-5} 
\multicolumn{1}{c|}{} & \textbf{EM} & \textbf{F1} & \textbf{EM} & \textbf{F1} \\ 
\hline
DROP        & 41.04 & 51.11 & 33.91 & 43.50 \\
RACE        & 37.22 & 50.46 & 28.96 & 41.42  \\
DuoRC       & 51.70 & 63.14 & 43.38 & 55.14  \\
BioASQ      & 59.62 & 74.02 & 49.74 & 66.57  \\
TQA  & 55.50 & 65.18 & 45.62 & 53.22  \\
RE          & 76.47 & 86.23 & 72.53 & 84.68  \\
BioProcess         & 56.16  & 72.91  & 46.12  & 63.63       \\
CWQ         & 54.73  & 61.39  & 51.80  & 59.05       \\
MCTest             & 64.56  & 78.72  & 59.49  & 72.20       \\
QAMR               & 56.36  & 72.47  & 48.23  & 67.39       \\
QAST               & 75.91  & 88.80  & 62.27  & 80.79       \\
TREC               & 49.85  & 63.36  & 36.34  & 53.55       \\
\hline
\end{tabular}
\end{adjustbox}
\caption{Results of the MRQA model on unseen datasets~\cite{su2019generalizing}. \textit{TQA}, \textit{RE} and \textit{CWQ} are, respectively, the abbreviations for \textit{TextbookQA}, \textit{RelationExtraction} and \textit{ComplexWebQuestions}.}
\label{tab:mrqa}
\end{table}

\section*{Chapter 4 Faithful to the Source: Generating Fact-aware Long-form Answers}

\subsection*{A Case of Zero-shot Experiments on NQ and HotpotQA}
\label{appendix:rbg_results}
In addition to automatic evaluation on the two datasets, we also show concrete examples to give a better understanding. As we can see in Table ~\ref{tab:examples}, ~\textbf{BART} tends to generate unfaithful answers as it lacks of external knowledge related to the specific question, and \textbf{DPR + BART} produces somewhat faithful but irrelevant answers, while ~\textbf{FiD} performs much better but still with partially unfaithful generations. In comparison, our ~\textbf{RBG} can generate more fluent, specific and factually accurate responses.

\begin{table*}[!th]
\centering
\small
\begin{adjustbox}{width={0.99\textwidth},totalheight={\textheight},keepaspectratio}
\begin{tabular}{p{18cm}}
\hline
\textbf{Question (from NQ):} What are the ranks in the us navy? \\
\hline
\textbf{Gold Answer:} ["E-8s senior chief petty officer", "E-9s master chief petty officer"] \\
\textbf{BART:}  (?) Navy ranks in the U.S. Navy. he Navy ranks are as follows: 1  Master Chief Petty Officer (MOS) a non-commissioned officer in the Navy. 2  Master Sergeant (MS), a Master Sergeant in the Marine Corps. 3  Sergeant Major (SSM), a Sergeant Major in the Air Force. \\
\textbf{DPR-BART:}  (?) The U.S. Navy is a branch of the United States Armed Forces. The Navy is made up of the Navy, Marine Corps, Air Force, Coast Guard, and Coast Guard.\\
\textbf{FiD:}  (*) The U.S. Navy is a branch of the United States Armed Forces. The rank system of the Navy consists of three categories: junior officers, senior officers, and flag officers. Junior officers are those officers in pay grades O-1 to O-4, \hlgreen{while senior officers are the ones with pay grades of O-5 and O-6. Senior officers have pay grades ranging from O-7 to E-9}. \hlgreen{Flag officers have a pay grade ranging from E-5 to E9}. \\
\textbf{RBG}(ours): The U.S. Navy ranks are divided into three categories: junior officers, senior officers, and flag officers. Junior officers are officers in pay grades O-1 to O-4, while senior officers are those in pay grades O-5 and O-6. Flag officers serve either as a line officer or as a staff corps officer. \\
\hline
\hline
\textbf{Question (from HotpotQA):} Which  French ace pilot and adventurer fly L'Oiseau Blanc?  \\
\hline
\textbf{Gold Answer}:["Charles Eugène"] \\
\textbf{BART:} (?) L'oiseau Blanc is a French aviator and aviator who flew the L'Oiseau blanc. He is best known for being the first aviator to fly it.. \\
\textbf{DPR-BART:} (?) L'oiseau Blanc l'oiseau blanc (French: The White Bird) was a French Levasseur Pl.8 biplane that disappeared in the early morning hours of April 1, 2014. The plane was a Levassellseur PL.8 Biplane that vanished in the morning of April 2, 2014, after being piloted by an adventurer pilot and adventurer pilot..\\
\textbf{FiD:} (*) On May 8, 1927, Charles Nungesser and Francois Coli took off from Paris on the L'Oiseau Blanc, a Levasseur PL.8 biplane \hlgreen{that had been designed by Charles Nongesser}, a French ace pilot and aviator, to fly from Paris to New York City in the Orteig Prize. \\
\textbf{RBG(ours)}:  On May 8, 1927, Charles Nungesser and Francois Coli boarded L'Oiseau blanc, a 450-hp Lorraine-powered Levasseur biplane designed to compete for the Orteig Prize. They took off from Paris on 8 May 1927 and were last seen over Ireland. Less than two weeks later, Charles Lindbergh successfully made the New York-Paris journey and claimed the prize in the Spirit of St. Louis.\\
\hline
\end{tabular}
 \end{adjustbox}
\caption{Examples of the zero-shot long-form answers on the NQ~\cite{kwiatkowski2019natural} and HotpotQA ~\cite{yang2018hotpotqa} datasets. RBG model generates more fluent, specific and factually accurate responses. ‘?’ indicates factually incorrect/irrelevant  responses; * indicates partially correct responses. We use the \hlgreen{green} to mark out the unfaithful snippets.}
\label{tab:examples}
\end{table*}

\subsection*{B Number of Retrieved Documents on Generation Quality}
\label{appendix:rbg_k_1}
\begin{table}[!ht]
\centering
\begin{tabular}{c|cc}
\hline
ndocs & ROUGE-L & F1    \\ \hline
5     & 24.63   & 27.29 \\
10    & \textbf{24.72}   & \textbf{27.52} \\
20    & 24.39   & 26.68 \\
50    & 23.43   & 25.94 \\ \hline
\end{tabular}
\caption{Generation performance versus the number of retrieved documents of our model on MS MARCO~\cite{nguyen2016ms}.}
\label{tab:results_k_1}
\end{table}

We also investigate the effects of number of retrieved documents $k$, on the answer generation quality. As we can see in Table~\ref{tab:results_k_1}, the generation quality in terms of ROUGE-L and F1, do not further improve as the number of $k$ increases, and the best performance are obtained when $k=10$ in our case.

\subsection*{C Document Retriever Model Details}
\label{appendix:rbg_retriever}
% As there is no golden retrievals provided in current LFQA datasets.
As the question/answers in LFQA may cover different domains and topics, we use a multi-task variant of DPR to guarantee the retrieval performance. The retriever is trained jointly on the union of all knowledge-intensive training data in KILT benchmark~\cite{petroni2021kilt}, including TrivaQA~\cite{Joshi_2017}, kwiatkowski2019naturaluestion~\cite{kwiatkowski2019natural}, HotpotQA~\cite{yang2018hotpotqa}, Fever~\cite{thorne2018fever}, zsRE~\cite{levy2017zero}, AY2, T-REx~\cite{elsahar2018t} and WoW~\cite{dinan2018wizard}.

\subsection*{D Human Evaluation Setup and Analysis}
\label{appendix:rbg_human_eval}

\textbf{Basic setup} As shown in Table~\ref{tab:human eval setup}, we sample 50 questions for each comparison and assign 3 annotators for each generated answer, which brings a workload of 450 judgments on model preference for each evaluation aspect. This process takes large amounts of energy and time considering the difficulty and challenges of factual-related annotation. We sample 10 questions from each of five development subsets corresponding to 5 levels of answer-passage overlap, which is a stratified sampling strategy. The answer position of each model is randomly shuffled to reduce the bias of position preference. 15 participants in our human evaluation are all researchers or students in computer science who speak and read English well.

\begin{table}[!ht]
\centering
\resizebox{0.60\textwidth}{!}
{
\begin{tabular}{c|ccc}
\hline
Comparison    & \#Questions & \#Annotators/answer  \\ \hline
RBG vs. FiD  &    50 &    3 \\
Reader analysis  &  50 &  3  \\
Pre-training analysis &  50 &  3 \\ \hline
\end{tabular}
}
\caption{Details of human evaluation for three comparisons.}
\label{tab:human eval setup}
\end{table}

\textbf{Scoring setup} We ask each annotator to select a score from \{1,2,3\} for each generated answer in terms of three aspects: \textit{fluency}, \textit{relevance} and \textit{factual correctness}. During scoring, the annotators are asked to preserve the relative better-or-not relationship between two models as much as possible. In particular, for the metric of factual correctness, the annotators check the correctness of all factual statements involved in a generated answer by referring to Wikipedia~(EN), other web pages and the golden answer. The answer with significantly fewer factual errors will get a higher score on factual correctness. We show cases in Table~\ref{tab:human eval cases} to demonstrate how the annotator evaluate three aspects in our experiment.

\textbf{Statistical analysis} We present the agreement among annotators on model preference in Table~\ref{tab:agreement} by calculating the Fleiss Kappa~\cite{fleiss_kappa} as the inter-rater consistency. The RBG vs. FiD comparison achieves better annotation agreement than other two ablation comparisons, maybe because RBG integrates both two of our contributions to improve the answer quality. In the comparison of RBG vs. FiD, annotators achieve a ``moderate agreement'' on the aspect of correctness and ``fair agreement'' on relevance~\cite{landis1977measurement}. Annotators achieve best agreements on fluency in all comparisons. It's more difficult to achieve a high degree of annotation consistency on factual correctness and relevance than fluency due to complicated facts involved in the generated answer. Therefore, we recommend taking preferred ratio as the core metric for factual-related evaluation following~\cite{krishna2021hurdles,nakano2021webgpt}. We also present score variance of four models involved in human evaluation in Table~\ref{tab:variance}. It's natural that the fluency score has the smallest variance while the difficult-to-annotate correctness has largest variance.

\begin{table}[!ht]
\centering
\resizebox{0.60\textwidth}{!}
{
\begin{tabular}{c|ccc}
\hline
Comparison    & fluency & relevance & correctness  \\ \hline
RBG vs. FiD  &    0.65 &    0.33 &    0.47 \\
Reader analysis  & 0.55 &    0.12 &    0.06  \\
Pre-training analysis&  0.62 &    0.16 &    0.08 \\ \hline
\end{tabular}
}
\caption{Agreement analysis for three comparisons in terms of three aspects. We use Fleiss Kappa~\cite{fleiss_kappa} to measure the agreement degree between annotators. The score range of [0,0.2] corresponds to slight agreement, [0.2,0.4 ] corresponds to fair agreement and [0.4,0.6] corresponds to moderate agreement~\cite{landis1977measurement}.}
\label{tab:agreement}
\end{table}

\begin{table}[!ht]
\centering
\resizebox{0.60\textwidth}{!}
{
\begin{tabular}{c|ccc}
\hline
Comparison    & fluency & relevance & correctness  \\ \hline
RBG  &    0.14 &    0.33 &    0.46 \\
RBG w/o reader & 0.22 &    0.34 &    0.48 \\
RBG w/o pre-training &  0.16 &    0.26 &    0.45 \\ 
RBG w/o reader+pre-training &  0.28 &    0.42 &    0.64 \\ \hline
\end{tabular}
}
\caption{Score variance of four models involved in human evaluation.}
\label{tab:variance}
\end{table}

\begin{table*}[htbp]
\small
\begin{adjustbox}{width={0.99\textwidth},totalheight={\textheight},keepaspectratio}
\begin{tabular}{|p{110pt}|p{260pt}lll|}
\hline
 Question & Generated answer & f & r & c \\ \hline
\multirow{2}{*}{effects of hyperthyroidism} &  Model1: Hyperthyroidism is a condition in which the thyroid gland produces too much thyroid hormone. The thyroid gland is a butterfly-shaped gland located in the neck. It is the only gland in the body that is responsible for the production of thyroid hormone, which helps regulate the body's metabolism. & 3 & 1 & 3 \\ \cline{2-5} 
 & Model2: Hyperthyroidism is a condition in which the thyroid gland produces too much thyroid hormone. This can cause the body to produce more thyroid hormone than it needs. This causes the thyroid to overproduce thyroid hormone, which can lead to a number of symptoms, including: 1  Weight gain. 2  Nausea. 3  Fatigue. 4  Vomiting. & 3 & 3 & 3 \\ \hline
\multirow{2}{*}{where is cabo de hornos} &  Model1: Cabo de Hornos is the southern tip of South America's Cape Horn Archipelago. It is the \hlgreen{largest island in the Pacific Ocean} and is the \hlgreen{only island in South America with a population of more than 2,000,000 inhabitants}. The island is also the \hlgreen{world's second-largest island} in terms of land area, with a total area of \hlgreen{1,826 square miles (4,926 square kilometers)} . & 3 & 2 & 1 \\ \cline{2-5} 
 &  Model2: Cabo de Hornos, Chile. Cabo de Hanoque, also known as Cape Horn, is the southern tip of South America. It is located in the south of Tierra del Fuego archipelago, in the AntÃ¡rtica Province, Magallanes Region. The municipality of Cabo De Hornos is the capital of the province of Puerto Williams. & 3 & 3 & 3 \\ \hline
\end{tabular}
\end{adjustbox}
\caption{Cherry-picked scoring cases by annotators in our human evaluation. f, r, and c respect fluency, relevance and correctness respectively. We use \hlgreen{green} to mark out factual incorrect snippets the annotator finds. The first case shows how annotators distinguish different levels of relevance and the second case shows that of factual correctness.}
\label{tab:human eval cases}
\end{table*}

\clearpage

\section*{Chapter 5 Generating Succinct Answers from Long-form Answers}
\subsection*{A Standard Few-shot Prompting of GPT-3}
\label{sec:appendix-gpt3}
~\citet{brown2020language} adopted the standard few-shot prompting on GPT-3, and evaluated on the three open-domain QA datasets NQ~\cite{kwiatkowski2019natural}, WQ~\cite{berant2013semantic} and TQA~\cite{joshi2017triviaqa}, for \textit{closed-book} QA task. In order to compare with their reported results, we re-implement their method using the same few-shot configuration as described in the paper and query the OpenAI API.

\textbf{Experimental Setups} 

As OpenAI hasn’t officially release information about their API model sizes, we deduce the sizes of OpenAI API models based on their performances from EleutherAI's blog\footnote{https://blog.eleuther.ai/gpt3-model-sizes/}. Specifically, we query Ada and Babbage models' API, trying to reproduce the reported results for GPT-3 Medium (350M) and GPT-3 XL (1.3B) models, respectively.

We use two prompt formats to query the OpenAI API. The first prompt format is the one described in the paper~\cite{brown2020language} (referred as \textit{GPT-3 format}): randomly draw 64 question-answer pairs from the corresponding training set, and use 'Q: ' and 'A: ' respectively as prefix before each question and answer, to build the conditioning prompts. We also use the prompt format from EleutherAI's language model evaluation harness github\footnote{https://github.com/EleutherAI/lm-evaluation-harness} (referred as \textit{EleutherAI}). Furthermore, we experiment using the same prompting format as we used in our standard prompting baseline (LM-530B) in Section~\ref{sec:baselines} (referred as \textit{Our format}), and prompting the LM of size 357M and 1.3B to compare.

\textbf{Results} 

We show the results of prompting GPT-3 under zero-shot, one-shot and few-shot settings in Table~\ref{tab:appendix-gpt-3-zeroshot}, Table~\ref{tab:appendix-gpt3-oneshot} and Table~\ref{tab:appendix-gpt3-few-shot} respectively. As we can see, no matter what prompting formats we use, the results reported in the GPT-3 paper~\cite{brown2020language} are almost always higher than our reproduced ones on all three datasets, over the two different LM sizes. The gaps become even larger at few-shot setting. Thus we conjuncture that we are not able to reproduce the results reported by ~\citet{brown2020language} using GPT-3 (175B) on the three QA datasets. So we did not include their reported results to compare with our CGAP method in Table~\ref{tab:main_results}.

Furthermore, we notice that the results based on our baseline's prompting configuration are always on par with the results from querying OpenAI API. Thus we believe that the \textbf{LM-530B} is a reliable and fair standard few-shot prompting baseline to compare with.

\subsection*{B LFQA Examples}

\label{sec:appendix_example}
We show three examples from NQ, TQA and WQ test set in Table~\ref{tab:example-NQ}, Table~\ref{tab:example-TQA} and Table~\ref{tab:example-WQ} respectively. In each table, we show the predicted answers from (1) standard prompting, (2) two-stage prompting using top-1 retrieved context $c^{\text{top-1}}_r$, (3) CGAP w/o marginalization, and (4) CGAP. All those predicted answers are based on LMs of size 530B.

We also show an example illustrate CGAP with 8 generated context and their corresponding predicted answer in Table~\ref{tab:appendix-gcap}. As we can see, the contexts that contains lot of factually inaccurate or irrelevant content (e.g. generated context 1, 2, 4, 5, 8), thus the corresponding answer is wrong/inaccurate. However, the context generation LM also generates contexts that are more relevant and factual (e.g. generated context 3, 6, 7), and they help the answer prediction LM generate a correct answer. Therefore, CGAP can predict the final answer correctly based on marginalization over generated contexts.

\begin{table*}[!ht]
\centering
\begin{adjustbox}{width=0.9\textwidth}{}
{
\begin{tabular}{ccllccc}
\hline
\multicolumn{2}{c}{\multirow{2}{*}{\textbf{Model Sizes}}} & \multirow{2}{*}{\textbf{Model sources}} & \multicolumn{1}{c}{\multirow{2}{*}{\textbf{Prompting format}}} & \multicolumn{3}{c}{\textbf{zero-shot}} \\ \cline{5-7} 
\multicolumn{2}{c}{} &  & \multicolumn{1}{c}{} & \textbf{NaturalQuestion} & \textbf{TriviaQA} & \multicolumn{1}{l}{\textbf{WebQuestion}} \\ \hline
\multicolumn{2}{c}{} & GPT-3 Medium & GPT-3 paper~\cite{brown2020language} & \textbf{1.75} & \textbf{7.61} & \textbf{3.20} \\
\multicolumn{2}{c}{\multirow{4}{*}{350M}} & \multirow{2}{*}{OpenAI API (Ada)} & GPT-3  format & 1.36 & 5.45 & 1.92 \\
\multicolumn{2}{c}{} &  & EleutherAI & 1.39 & 5.54 & 2.46 \\
\multicolumn{2}{c}{} & LM-357M & Our format & 1.41 & 5.04 & 2.12 \\ \hline
\multicolumn{2}{c}{} & GPT-3 XL & GPT-3 paper ~\cite{brown2020language} & \textbf{4.40} & \textbf{19.70} & 4.63 \\
\multicolumn{2}{c}{\multirow{4}{*}{1.3B}} & \multirow{2}{*}{OpenAI API (Babbage)} & GPT-3  format & 2.27 & 9.84 & 2.12 \\
\multicolumn{2}{c}{} &  & EleutherAI & 2.47 & 12.77 & 5.22 \\
\multicolumn{2}{c}{} & LM-1.3B & Our format & 3.88 & 14.13 & \textbf{5.61} \\ \hline
\end{tabular}
}
\end{adjustbox}
\caption{Standard zero-shot prompting of GPT-3 for open-domain QA. }
\label{tab:appendix-gpt-3-zeroshot}
\end{table*}

\begin{table*}[htbp]
\centering
\begin{adjustbox}{width=0.9\textwidth}{}
{
\begin{tabular}{ccllccc}
\hline
\multicolumn{2}{c}{\multirow{2}{*}{\textbf{Model Sizes}}} & \multirow{2}{*}{\textbf{Model sources}} & \multicolumn{1}{c}{\multirow{2}{*}{\textbf{Prompting format}}} & \multicolumn{3}{c}{\textbf{one-shot(k=1)}} \\ \cline{5-7} 
\multicolumn{2}{c}{} &  & \multicolumn{1}{c}{} & \textbf{NaturalQuestion} & \textbf{TriviaQA} & \multicolumn{1}{l}{\textbf{WebQuestion}} \\ \hline
\multicolumn{2}{c}{} & GPT-3 Medium & GPT-3 paper~\cite{brown2020language} & \textbf{3.07} & \textbf{12.90} & \textbf{6.20} \\
\multicolumn{2}{c}{\multirow{4}{*}{350M}} & \multirow{2}{*}{OpenAI API (Ada)} & GPT-3  format & 1.83 & 10.26 & 5.07 \\
\multicolumn{2}{c}{} &  & EleutherAI & 1.77 & 10.02 & 5.61 \\
\multicolumn{2}{c}{} & LM-357M & Our format & 2.24 & 9.75 & 5.12 \\ \hline
\multicolumn{2}{c}{} & GPT-3 XL & GPT-3 paper~\cite{brown2020language} & \textbf{5.43} & \textbf{26.50} & 9.15 \\
\multicolumn{2}{c}{\multirow{4}{*}{1.3B}} & \multirow{2}{*}{OpenAI API (Babbage)} & GPT-3  format & 3.55 & 20.56 & 8.27 \\
\multicolumn{2}{c}{} &  & EleutherAI & 3.55 & 21.45 & \textbf{9.45} \\
\multicolumn{2}{c}{} & LM-1.3B & Our format & 4.71 & 21.21 & 8.76 \\ \hline
\end{tabular}
}
\end{adjustbox}
\caption{Standard one-shot prompting of GPT-3 for open-domain QA. }
\label{tab:appendix-gpt3-oneshot}
\end{table*}

\begin{table*}[!ht]
\centering
\begin{adjustbox}{width=0.9\textwidth}{}
{
\begin{tabular}{ccllccc}
\hline
\multicolumn{2}{c}{\multirow{2}{*}{\textbf{Model Sizes}}} & \multirow{2}{*}{\textbf{Model sources}} & \multicolumn{1}{c}{\multirow{2}{*}{\textbf{Prompting format}}} & \multicolumn{3}{c}{\textbf{few-shot(k=64)}} \\ \cline{5-7} 
\multicolumn{2}{c}{} &  & \multicolumn{1}{c}{} & \textbf{NaturalQuestion} & \textbf{TriviaQA} & \multicolumn{1}{l}{\textbf{WebQuestion}} \\ \hline
\multicolumn{2}{c}{} & GPT-3 Medium & GPT-3 paper~\cite{brown2020language} & \textbf{4.46} & \textbf{16.30} & \textbf{12.60} \\
\multicolumn{2}{c}{\multirow{4}{*}{350M}} & \multirow{2}{*}{OpenAI API (Ada)} & GPT-3  format & 3.43 & 12.46 & 10.73 \\
\multicolumn{2}{c}{} &  & EleutherAI & 3.71 & 12.46 & 10.29 \\
\multicolumn{2}{c}{} & LM-357M & Our format & 3.85 & 11.66 & 10.97 \\ \hline
\multicolumn{2}{c}{} & GPT-3 XL & GPT-3 paper~\cite{brown2020language} & \textbf{9.72} & \textbf{32.10} & \textbf{19.60} \\
\multicolumn{2}{c}{\multirow{4}{*}{1.3B}} & \multirow{2}{*}{OpenAI API (Babbage)} & GPT-3  format & 8.28 & 24.70 & 18.95 \\
\multicolumn{2}{c}{} &  & EleutherAI & 7.81 & 24.93 & 18.16 \\
\multicolumn{2}{c}{} & LM-1.3B & Our format & 7.87 & 24.88 & 17.52 \\ \hline
\end{tabular}
}
\end{adjustbox}
\caption{Standard few-shot (k=64) prompting of GPT-3 for open-domain QA.}
\label{tab:appendix-gpt3-few-shot}
\end{table*}

\clearpage
\begin{table*}[!ht]
\centering
\begin{adjustbox}{width=0.95\textwidth}
{
\begin{tabular}{ll}
\hline
\multicolumn{2}{l}{\textbf{Question}: When is the next deadpool movie being released?} \\ \hline
Golden Answer: & {[}May 18, 2018{]} \\
Predicted Answer (standard prompting): & \textcolor{mypink3}{Prime availability TBD} \\
Predicted Answer ($c^{\text{top-1}}_r$): & \textcolor{mygreen1}{May 18, 2018} \\
Predicted Answer (CGAP w/o marginalization): & \textcolor{mygreen1}{May 18, 2018} / \textcolor{mypink3}{date21-May-2018} / \textcolor{mypink3}{May 29, 2019} /\textcolor{mypink3}{16th May 2018} \\
Predicted Answer (CGAP): &\textcolor{mygreen1}{May 18, 2018} \\ \hline
\end{tabular}
}
\end{adjustbox}
\caption{Example from NQ~\cite{kwiatkowski2019natural} test set. Red and green colors denote \textcolor{mypink3}{in-correct} and \textcolor{mygreen1}{correct} answer, respectively.}
\label{tab:example-NQ}
\end{table*}

\begin{table*}[!ht]
\centering
\begin{adjustbox}{width=\textwidth}
{
\begin{tabular}{ll}
\hline
\multicolumn{2}{l}{\textbf{Question}: Which sitcom star appeared on the big screening 'The Object of My Affection'?} \\ \hline
Golden Answer: & [Jennifer Anniston, Jen Aniston,  ...] \\
Predicted Answer (standard prompting): & \textcolor{mypink3}{Ross Hatley} \\
Predicted Answer ($c^{\text{top-1}}_r$): & \textcolor{mypink3}{Laurie Metcalfe} \\
Predicted Answer (CGAP w/o marginalization): & \textcolor{mygreen1}{Jennifer Aniston} / \textcolor{mypink3}{ Paul Rudd} /\textcolor{mypink3}{ Christine Baranski} / \textcolor{mypink3}{Lisa Kudrow} \\
Predicted Answer (CGAP): & \textcolor{mygreen1}{Jennifer Aniston} \\ \hline
\end{tabular}
}
\end{adjustbox}
\caption{Example from TriviaQA~\cite{joshi2017triviaqa} test set.}
\label{tab:example-TQA}
\end{table*}

\begin{table*}[!ht]
\centering
\begin{adjustbox}{width=\textwidth}
{
\begin{tabular}{ll}
\hline
\multicolumn{2}{l}{\textbf{Question}: Where George Lopez was born?} \\ \hline
Golden Answer: & {[}Mission Hills{]} \\
Predicted Answer (standard prompting): & \textcolor{mypink3}{San Fernando, California} \\
Predicted Answer ($c^{\text{top-1}}_r$): & \textcolor{mypink3}{Washington Heights} \\
Predicted Answer (CGAP w/o marginalization): & \textcolor{mygreen1}{Mission Hills} / \textcolor{mypink3}{Los Angeles}/ \textcolor{mypink3}{San Fernando} /\textcolor{mypink3}{ Castle Hill} /\textcolor{mypink3}{ canada }\\
Predicted Answer (CGAP): & \textcolor{mygreen1}{Mission Hills} \\ \hline
\end{tabular}
}
\end{adjustbox}
\caption{Example from WebQuestions~\cite{berant2013semantic} test set.}
\label{tab:example-WQ}
\end{table*}

\clearpage

\begin{table*}[htb]
\small
\centering
\begin{adjustbox}{width={.95\textwidth},totalheight={1.\textheight},keepaspectratio}
\begin{tabular}{p{20cm}}
\hline
\textbf{Question}: Where George Lopez was born?       \textbf{Golden Answer:} [Mission Hills]\\
\hline
\textbf{Generated Context 1}: George Lopez They are American citizens, born in Los Angeles, to Gregorio Lopez (October 26, 1934 – January 25, 2011) and Maria Torres (October 12, 1938 – October 25, 2014), Mexican immigrants who were born in Reynosa, Tamaulipas. Lopez's paternal grandparents were Mexican, while his maternal grandparents were of Mexican and Spanish descent. Lopez's father was a worker for the Los Angeles Metropolitan Transit Authority. He married Maria in Los Angeles when he was 16 years old, and she was 15. Maria filed for divorce from Gregorio in the mid-1970s. \\ 
\textbf{Predicted Answer:} Los Angeles \\
\textbf{Generated Context 2}: George Lopez born April 23, 1961. The family eventually moved from Marín to Los Angeles in the mid-1960s. After Lopez graduated from San Fernando High School in 1979, he attended the California State University, Los Angeles. It was there where Lopez discovered his comedic talent after a video of him doing a stand-up routine made it into the hands of "Tonight Show" booker, Fred De Cordova, in 1985. \\
\textbf{Predicted Answer:} canada \\
\textbf{Generated Context 3}: George Lopez at 20th Century Fox Television until 2010, when he began appearing in his self-produced sitcom, George Lopez. Lopez was born on April 23, 1961, in Mission Hills, Los Angeles, California. He was raised by his maternal grandmother, Benita Gutierrez, a factory worker, and his maternal grandfather, Refugio Gutierrez, who worked as a janitor and metal fabricator at an ice skating rink. \\
\textbf{Predicted Answer:} Mission Hills \\
\textbf{Generated Context 4}: George Lopez in San Fernando, California. He is of Mestizo Mexican descent. His father, Saul Lopez, was from Sonora, and his mother, Elvira, was a native of Arizona. They were migrant workers who picked crops and moved often, sometimes taking their seven children with them. While growing up, Lopez did not speak English until he was seven. Lopez attended Our Lady of Lourdes, a Roman Catholic elementary school in Pico Rivera. \\
\textbf{Predicted Answer:} San Fernando \\
\textbf{Generated Context 5}: George Lopez A theater called "The Tower", which now houses a branch of Bank of America, stands on the former site of his childhood home. Lopez was born at Daniel Freeman Memorial Hospital in Inglewood, California and was raised by his maternal grandmother, Benita Gutierrez, a factory worker who took him to the movies every week. He grew up in the Echo Park neighborhood of Los Angeles, California and attended many different schools in the Los Angeles Unified School District.\\
\textbf{Predicted Answer:} Los Angeles \\
\textbf{Generated Context 6}: George Lopez López was born in Mission Hills, Los Angeles, California. He is the son of Frieda and Anatasio López, Mexican immigrants from Tepic, Nayarit, Mexico. Lopez's father was a farm worker, and his mother worked at a canning factory. Lopez has stated that he had a difficult childhood; his parents divorced when he was five, and his mother struggled to raise him and his two sisters. \\
\textbf{Predicted Answer:}  Mission Hills \\
\textbf{Generated Context 7}: George Edward Lopez. Lopez was born in Mission Hills, Los Angeles. His parents, Frieda and Anatasio Lopez, were both born and raised in Texas. His mother was of Mexican descent and his father, also called ""Charlie"", was of Sicilian descent. Lopez grew up with two older sisters. He attended San Fernando High School in the Los Angeles neighborhood of San Fernando.\\
\textbf{Predicted Answer:} Mission Hills \\
\textbf{Generated Context 8}: George Lopez had in his mouth. He was raised by his mother, Frieda, a telephone company employee, and his grandmother, Benita Gutierrez, a factory worker. Lopez graduated from the Roman Catholic Saint Monica's High School in 1974, where he met his future wife, Ann Serrano. He moved to California and graduated from the California State University, Los Angeles, in 1986 with a degree in communications.\\
\textbf{Predicted Answer:} Castle Hill \\ 
\hline
\textbf{Predicted Answer (CGAP):}  Mission Hills \\

\hline
\end{tabular}
\end{adjustbox}
\vspace{-10pt}
\caption{Example illustrates CGAP, in which we marginalize over 8 generated contexts.}
\label{tab:appendix-gcap}
\end{table*}

\end{document}